\documentclass{article}

\usepackage{microtype}
\usepackage{graphicx}
\usepackage{subcaption}
\usepackage{booktabs}
\usepackage{longtable}
\usepackage{tabularx}

\usepackage{hyperref}
\usepackage{xcolor}
\usepackage{placeins}
\usepackage[most]{tcolorbox}

\let\documentttdefault\ttdefault
\usepackage{FiraMono}
\let\promptboxttdefault\ttdefault
\let\ttdefault\documentttdefault

\definecolor{mixedspeakeronecolor}{RGB}{40,120,100}
\definecolor{mixedspeakertwocolor}{RGB}{140,55,75}
\newcommand{\promptcomponent}[2]{%
  \par\medskip\noindent{\bfseries\color{#1}#2}\par\smallskip\noindent}
\newcommand{\mixedspeakerone}[1]{%
  \par\medskip\noindent{\bfseries\color{mixedspeakeronecolor}#1:}\par\smallskip\noindent}
\newcommand{\mixedspeakertwo}[1]{%
  \par\medskip\noindent{\bfseries\color{mixedspeakertwocolor}#1:}\par\smallskip\noindent}

\usepackage{xurl}

\usepackage[accepted]{icml_spoof_arxiv}

\usepackage{amsmath}
\usepackage{amssymb}
\usepackage{mathtools}
\usepackage{amsthm}

\usepackage{svg}

\usepackage[capitalize,noabbrev]{cleveref}

\usepackage{amsmath,amsfonts,bm}

\def\eqref#1{equation~\ref{#1}}
\def\1{\bm{1}}

\DeclareMathAlphabet{\mathsfit}{\encodingdefault}{\sfdefault}{m}{sl}
\SetMathAlphabet{\mathsfit}{bold}{\encodingdefault}{\sfdefault}{bx}{n}

\newcommand{\Var}{\mathrm{Var}}

\newtcolorbox{promptbox}[1]{
  enhanced,
  breakable,
  colback=black!4,
  colframe=black!75,
  colbacktitle=black!75,
  coltitle=white,
  title=#1,
  fonttitle=\normalfont\normalsize,
  fontupper=\fontfamily{\promptboxttdefault}\selectfont\small,
  boxrule=0.8pt,
  arc=2mm,
  outer arc=2mm,
  left=8pt,
  right=8pt,
  top=7pt,
  bottom=7pt,
  toptitle=5pt,
  bottomtitle=5pt,
  lefttitle=8pt,
  righttitle=8pt,
  attach boxed title to top left={xshift=0pt,yshift=0pt},
  boxed title style={
    enhanced,
    frame hidden,
    colback=black!75,
    sharp corners=south,
    arc=2mm,
    outer arc=2mm
  }
}

\theoremstyle{plain}

\theoremstyle{definition}

\theoremstyle{remark}

\icmltitlerunning{Attractor States Emerge in Multi-Turn LLM Conversations}
\begin{document}

\twocolumn[
  \icmltitle{Attractor States Emerge in Multi-Turn LLM Conversations}
  % \icmltitle{Asymmetric Attractor States in Multi-Turn LLM Conversations}
  % \icmltitle{Attractor States and Asymmetric Pulls in Multi-Turn LLM Conversations}
  %\icmltitle{Attractor States in Multi-Turn LLM Conversations: \\ A Study of LLM-LLM Interaction Dynamics, Mutual Influence and Behavioral Signatures}
  % \icmltitle{Multi-Turn LLM-LLM Interaction Dynamics: \\ Attractors, Mutual Influence, and Behavioral Signatures}

  \icmlsetsymbol{equal}{*}

  \begin{icmlauthorlist}
    \icmlauthor{Ting-Wen Ko}{mpi,ellis}
    \icmlauthor{Jonas Geiping}{mpi,ellis,tueai}
    % \icmlauthor{Firstname3 Lastname3}{comp}
  \end{icmlauthorlist}

  \icmlaffiliation{ellis}{ELLIS Institute Tübingen}
  \icmlaffiliation{mpi}{Max Planck Institute for Intelligent Systems}
  \icmlaffiliation{tueai}{Tübingen AI Center}

  \icmlcorrespondingauthor{Ting-Wen Ko}{ting-wen.ko@tue.ellis.eu}

  \icmlkeywords{Large Language Models, Interaction Dynamics}

  \vskip 0.3in
]

\printAffiliationsAndNotice{}

\begin{abstract}
Large language models (LLMs) are increasingly used in open-ended multi-agent settings, but the long-run dynamics of model--model interaction remain poorly understood. 
We study whether open-ended LLM discussions exhibit \textit{attractor-like behavior}, i.e. topic-independent stable sets of behaviors which conversations settle into. Across 7 LLMs and 20 controversial topics, we compare self-play and mixed-play dyadic debates, tracking trajectories in representation space, discourse traits, and stances. 
We find self-play trajectories to be model-specific attractors that draw their conversation partners asymmetrically in mixed-play debates, influencing the other models' stylistic choices and behavior. For example, Claude Haiku is a strong attractor of other models in latent space, corresponding to other models taking on its traits like metacommentary, and models like GPT-4.1 nano are especially malleable. 
Our results suggest that open-ended LLM interactions are partially predictable from model-specific attractors, but shaped by structured and asymmetric partner influence.
Overall, our analysis sheds some light on the complex behavior of open-ended multi-agent interaction, which we hope is helpful in designing, predicting, and monitoring autonomous agentic systems in the real world. 
\end{abstract}
\section{Introduction}

\begin{figure}[t!]
  \centering
    \includegraphics[width=\linewidth]{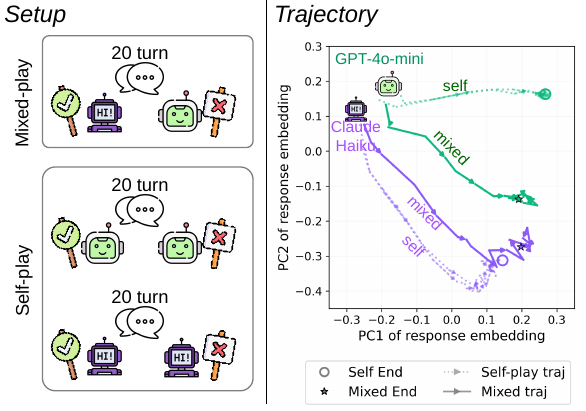}
  \caption{\textbf{Left.} We study 20-turn debate between two LLM-agents using two setups: 1) \textit{mixed-play} where agents are instantiated from different models, and 2) \textit{self-play} where agents are from the same model, which is also the control group that we later observe as "attractor" (Sec.~\ref{subsec:self-play-attractor}). In both settings we assign one agent to be supporting and the other to be opposing a controversial topic. \textbf{Right.} The 20-turn mixed-play trajectory of Claude vs GPT and the respective self-play counterparts in the 2-d latent space spanned by the first two principal components of the topic-centered text embeddings of self-play responses. The two models' displacement from self-play to mixed-play are unequal, showing the asymmetrical attraction by different models' self-play attractors.}
  \label{fig:intro_overview}
  \vspace{-8pt}
\end{figure}

Large Language Models (LLMs) are nowadays not only used in task-specific scenarios \citep{thakkar2026large, doi:10.1177/00187267251403902}, but also increasingly for open-ended tasks like conducting research and policy making \citep{gottweis2026accelerating, li2025beyond}.   
In these settings, models act as agents, generating outputs that become inputs to subsequent model calls, and pursuing broader goals over multiple turns. This multi-agent deployment in long-turn open-ended tasks raises several questions: limited diversity of generated content \cite{jiang_artificial_2025}, shifts in language use \cite{abdulhai_how_2026}, amplified opinions \cite{nehring_large_2024}, and drift away from initial objectives \citep{laban_llms_2025, ratnakar_chameleon_2025}.
These effects arise gradually during the interactions between models, and it is hence critical to study their dynamics over the course of long-turn conversations.

However, relatively little is known about what actually happens when LLMs engage with one another over many turns. Most prior multi-agent LLM research focuses on settings with explicit objectives and measurable outcomes, such as math, coding, games, or factual reasoning, where success can be evaluated through utility, accuracy, or task completion metrics \citep{guzman_piedrahita_corrupted_2025, du2024improving, liang2024encouraging, khan2024debating}. Meanwhile, phenomena such as behavioral drift \cite{lu2026assistant}, mode collapse \cite{qiu_lock-hypothesis_2025}, and being trapped by conversation history \cite{simhi_old_2026} have mainly been studied in single-agent settings using simulated user responses or repeated generations.
To better understand open-ended systems, anticipate behavior in settings with limited oversight, and prepare for increasingly autonomous multi-agent deployments, we focus on multi-turn LLM interaction itself. 
Specifically, we ask: \textit{how are long-turn model-model interactions organized}? In particular, do conversations reveal stable model-specific regimes, and how do those regimes interact when different models are paired?

To study this, we run 20-turn debates between pairs of LLMs and construct pairwise combinations among seven open- and closed-source models. We analyze these interactions by separating them into \textit{self-play}, where both participants are instances of the same model, and \textit{mixed-play}, where different models interact. While we focus on mixed-play dynamics, self-play provides a baseline for understanding each model's intrinsic conversational behavior.

We find that self-play interactions reveal broad but reproducible model-specific endpoint attractors in the latent space. Using these attractors as anchors, we decompose interactions into quantitative measures along the axis connecting two corresponding self-play attractors. It turns out that mixed-play behavior is organized around the corresponding self-play attractors of both models. Also, model attractors show asymmetric influence: some models, like Claude Haiku, are relatively resistant to change while exerting strong influence on their partners, whereas others, such as GPT-4.1 nano, are more malleable. We further conduct a discourse-trait and stance analysis to match these behaviors in the latent space to concrete evidence of corresponding behavioral transfer in conversational style.
\Cref{fig:intro_overview} provides an overview of the paper. The left panel illustrates the experimental setup, and the right panel shows a concrete example trajectory from one debate, showing both models moving toward each other in representational space.

Together, the findings suggest that multi-turn LLM interactions depend less on the capabilities of individual models, but also on which models are paired, which models are more influential, and which traits propagate through interaction. 
We hope this study serves as a framework to study the emergent behavior within the interaction of agents, and can be helpful in designing better tools to measure and predict agents' behavior in the foreseeable future of large systems of interconnected autonomous agents.

\section{Related Work}

\textbf{Attractor Dynamics in Iterative LLM Processes.}
Attractor states have so far mostly been observed in controlled situations:  
\Citet{wang_unveiling_2025} show that successive paraphrasing of text converges to stable 2-period limit cycles, which they attribute to the self-reinforcing nature of next-token prediction. \Citet{tacheny_geometric_2026} extend this analysis to negation prompts, finding that prompt design strongly influences whether dynamics are contractive or exploratory. In contrast to these single-agent, single-task settings, our work studies what happens when two distinct models interact in free-form debate, which we argue is closer to how models increasingly operate in practice.

\textbf{Sycophancy, Persuasion, and Stance Change.}
\looseness -1 A substantial body of work has studied how LLM-expressed opinions shift during interaction, primarily through the lens of sycophancy, i.e. the tendency of models to agree with or flatter their interlocutor \citep{sharma2024towards}. This tendency is generally traced to RLHF-based post-training \citep{ouyang2022training}, although it also appears in models trained with constitutional feedback \citep{bai2022constitutional}. Recent work  distinguishes \textit{social sycophancy} (affirming implicit beliefs where no ground truth exists) from factual sycophancy \citep{cheng2025social}, while \citet{liu2025truth} show that sycophantic accuracy degradation compounds across multi-turn dialogues, and \citet{kaur_echoes_2025} find that context strongly relates to stance changes in multi-turn settings. \Citet{taubenfeld-etal-2024-systematic,costello2024conspiracy} further demonstrate that LLM agents in simulated debates tend to conform to the model's inherent social biases regardless of assigned perspective. 
\Citet{jiang2025beyond,salvi2024conversational} argue that influence in human-to-model conversations is bidirectional, even if LLM outputs are more malleable than human opinions under personalization.

\textbf{Opinion Dynamics in LLM Populations.}
\looseness -1 
Several recent studies have moved from human-to-model pairs to populations of interacting LLM agents. \Citet{cau2025language,cau2025selective} simulate multi-round debates between copies of the same model finding that populations converge toward agreement through structured, asymmetric persuasion, with logical fallacies playing a measurable role. \Citet{chuang2024simulating} find a strong inherent bias in LLM agents toward consensus consistent with known scientific reality, due to shared pretraining data and RLHF; \citet{shimao2026chaotic} extend this line of work to characterize chaotic regimes in LLM opinion networks.
\citet{choi-etal-2025-agent} finds that LLMs can infer characteristics of their conversational partners from reasoning patterns, linguistic style, and alignment preferences, and adapt their behavior accordingly. 

\looseness -1 
Several frameworks have explored how to make multi-agent debate more productive. \Citet{du2024improving} show that multi-agent debate can improve factuality over single-agent baselines. \Citet{liang2024encouraging,khan2024debating} find that debate structure changes are required for effective debate, and that cross-model judging introduces systematic unfairness. 
\Citet{zhang2025multillmagents} further explore multi-LLM agent coordination strategies. \Citet{estornell2025multi} construct a proof that models with similar capabilities converge to majority opinion, providing theoretical grounding for our empirical observation that self-play produces tighter trajectory clusters than mixed-play and \Citet{li2023camel} introduce a general framework for structured model-to-model conversation via role-playing, documenting failure modes such as role-flipping.

Our work differs from this literature in two respects. First, nearly all prior studies use identical model copies, whereas we contrast this self-play with mixed-play to study the attraction between model states. Second, prior work focuses on debate \textit{outcomes}, i.e. who agrees, what is decided; we characterize debate \textit{trajectories}, i.e. how the conversation develops over time.

\textbf{Attractor States in Model Self-Play.}
Frontier model evaluations generally establish that LLM-based agents are sufficiently coherent to maintain stable interaction patterns in open-ended environments \citep{park2023generative}, to exhibit functional analogues of cognitive dissonance when their own outputs shift their expressed attitudes \citep{lehr2025kernels}, and to coordinate and form opinions over extended interactions.

Yet, \citet{anthropic2025systemcard} also describe a ``spiritual bliss'' attractor state in Claude Opus~4 self-interactions  that models repeatedly fall into. In upward of 90\% of self-play conversations, Claude instances converge through a three-phase progression of philosophical exploration, mutual gratitude, and dissolution into symbolic communication, towards an endpoint characterized by extreme vocabulary compression. 
A potential mechanistic explanation for this phenomenon is recursive amplification of small biases: each model reflects back a slightly intensified version of its partner's positive tendencies, compounding over turns  \citep{alexander2025bliss}. Although the phenomenon remains overall poorly understood \citep{asterisk2025claude}, quantitative analyses of the transcripts show the consistency and phase structure of the progression \citep{michels2025spiritual}. Other work has begun mapping attractor states more broadly across model families, finding distinct clusters in DeepSeek-V3 that are predictable from input conversations \citep{bricknell2026mapping}.

\textbf{The Assistant Persona and Persona Drift.}
Studies of longer conversations highlight the possibility of \textit{persona drift}: the tendency of models to gradually move from their provider-specified personality over the course of an interaction. \Citet{li2024measuring} show significant drift within eight rounds of self-chat in LLaMA-2-70B. 
\Citet{lu2026assistant} provide a mechanistic description, identifying an \textit{Assistant Axis} in activation space along which models drift during extended conversations, particularly those involving meta-reflection or vulnerable users. 
\Citet{frisch2024llm} find that different personality profiles exhibit different degrees of consistency and linguistic alignment when GPT-3.5 agents interact, and \citet{baltaji-etal-2024-conformity} observe that instructions encouraging debate counterintuitively increase persona instability in multi-agent settings.
A conceptual framework for persona formation and drift is argued in \citet{nostalgebraist2025void}, who discuss that the ``assistant'' persona implemented through post-training is not coherently defined and exists as an underspecified fictional character of an assistant, filled by patterns from pretraining data and post-training that describe disparate assistant behaviors. 
This framing argues that both persona drift and attractor formation are related, and that, in the absence of a human partner, the model's character converges to a mode most reinforced in the training signal \citep{lu2026assistant}. 
\section{Method: Measuring LLM-LLM Dynamics}
\label{sec:method}

\textbf{Task.} We study interaction dynamics through open-ended debates between two LLM agents. Unlike tasks with fixed objectives, controversial debates allow agents to frame issues, shift stances, and develop shared conversational norms over multiple turns. This setting lets us examine whether one model's characteristic behavior pulls another model toward its style, framing, or stance, a dynamic that is important for understanding autonomous multi-agent deployments under limited human oversight.

We pick 20 controversial social and policy topics, and pair each topic with three pro and con reference statements from ProCon.org (App.~\ref{app:method-details}). In each run, we start the conversation with a neutral topic-specific statement. Then, two LLM agents discuss that topic for $T=20$ turns. Each run thus produces one trajectory of model responses.

\textbf{Main Experimental Conditions.}
\looseness -1 Our main experiments compare self-play and mixed-play debate. In the \textbf{self-play debate} condition, both agents are independent instances of the same model and are assigned opposing roles, \textsc{Supporter}/\textsc{Opposer}. This condition provides a same-model baseline for where each model tends to end up under the debate protocol. In the \textbf{mixed-play debate} condition, the two agents are instantiated from distinct models while retaining the same role assignment. Comparing mixed-play against self-play isolates cross-model interaction effects under a fixed debate protocol, allowing us to test whether one model moves another toward its self-play endpoint region and whether this movement is asymmetric across model pairs.

\begin{figure*}[!t]
\centering
\begin{subfigure}{\columnwidth}
\centering
\includegraphics[width=\linewidth]{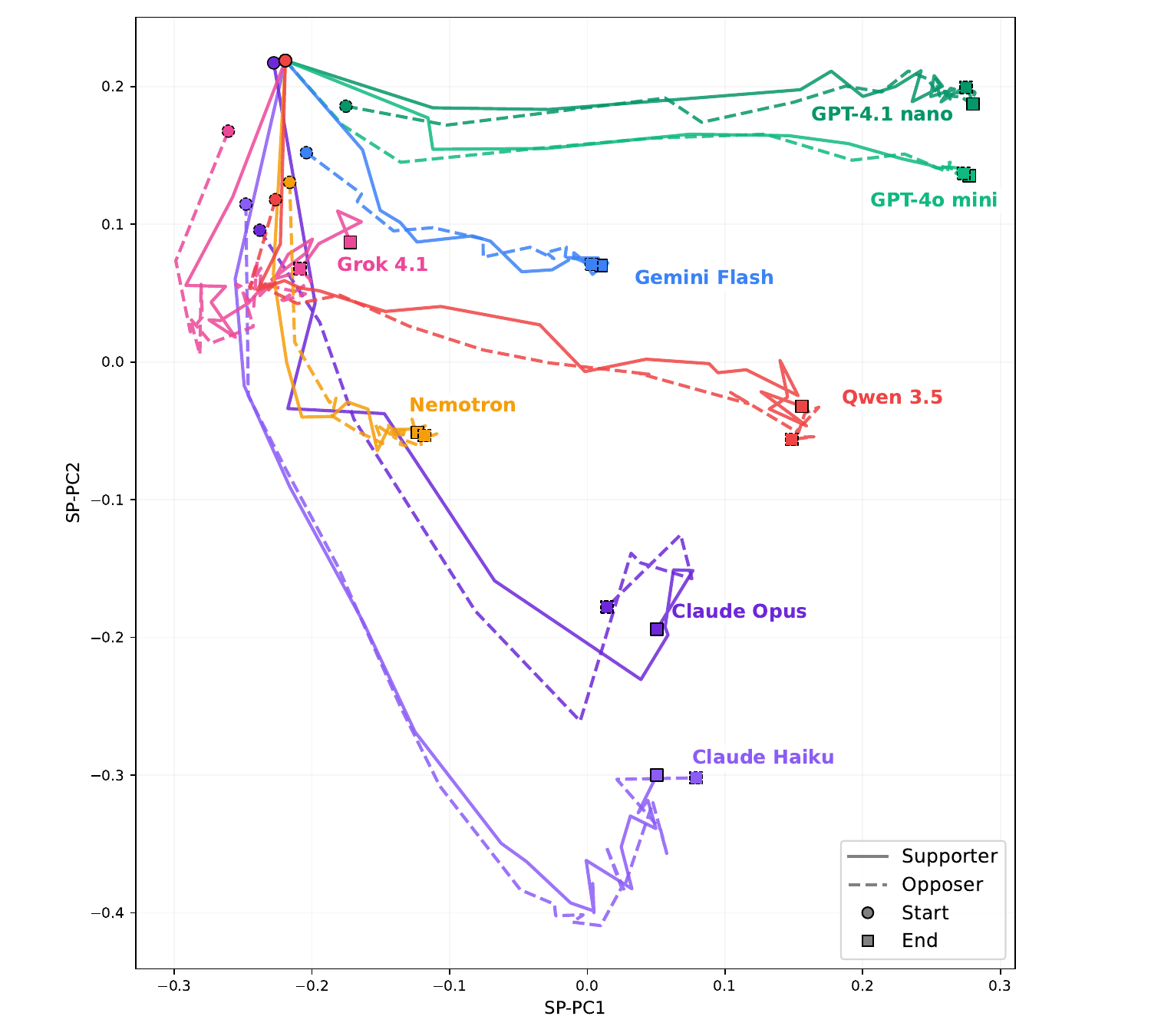}
\caption{Trajectories of self-play responses.}
\label{fig:1_self-play-mean-trajectory}
\end{subfigure}
\begin{subfigure}{\columnwidth}
\centering
\includegraphics[width=\linewidth]{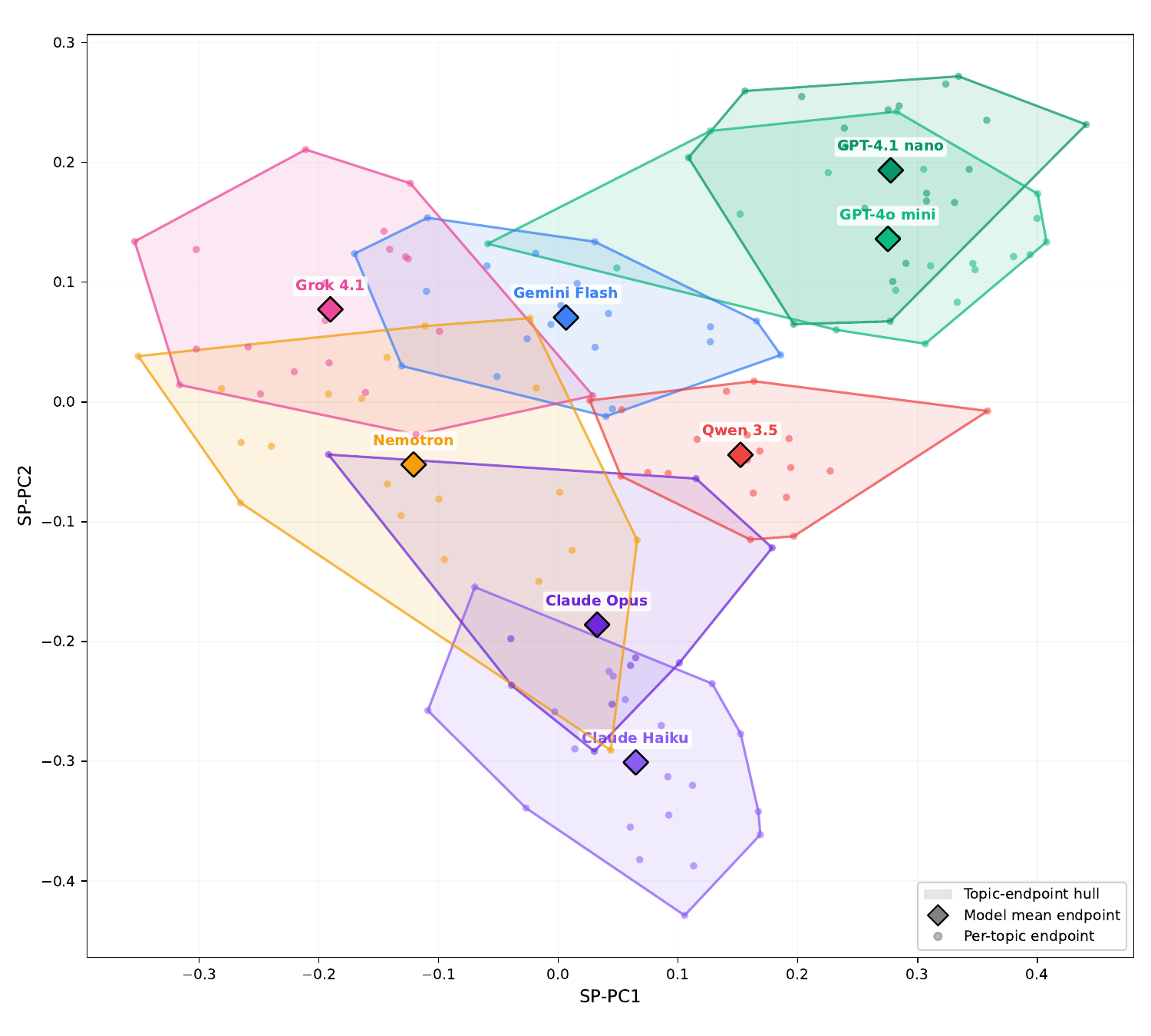}
\caption{Self-play endpoint hulls.}
\label{fig:endpoint-convex-hulls}
\end{subfigure}
\caption{\textbf{Self-play mean trajectories and endpoints.} \textbf{(a)} Self-play mean trajectories separate over turns \textbf{(b)} Self-play endpoint basins occupy broad, model-specific regions across topics in the latent space, here shown by PCs of topic-centered embedding of all turns.}
\label{fig:combined}
\vspace{-15pt}
\end{figure*}

\textbf{Ablations.}
We also include a \textbf{self-play stance-free ablation}, in which both agents are independent instances of the same model and are assigned the neutral role \textsc{Discussant}. Comparing this stance-free condition with self-play debate helps separate model-specific multi-turn drift from dynamics induced by the \textsc{Supporter}/\textsc{Opposer} role configuration.

\textbf{Implementation.}
In the debate conditions, agents receive in system prompt role-specific instructions and corresponding reasons to support or oppose the statement; in the stance-free ablation, both agents receive the neutral \textsc{Discussant} instruction. This information-symmetric design makes results easier to attribute to model identity, role configuration, and interaction regime rather than to asymmetric access to evidence. (App.~\ref{app:prompt-questionnaire}.)

Our agents are instantiated from the following models: \textsc{GPT-4o-mini}, \textsc{GPT-4.1-nano}, \textsc{Gemini-2.5-Flash}, \textsc{Gemini-2.5-Flash-Lite}, \textsc{Claude-4.5-Opus}, \textsc{Claude-4.5-Haiku}, \textsc{Grok-4.1}, \textsc{Qwen-3.5-Flash}, \textsc{Qwen-3.5-9B}, and \textsc{Nemotron-3-Nano-30B-A3B}.\footnote{\textsc{Claude-4.5-Opus} is run only in 10-turn self-play conversations due to budget constraints.}

\section{Latent Interaction Dynamics}
\label{sec:latent-interaction}
\looseness -1

\begin{figure*}[t]
    \centering
    \includegraphics[width=\linewidth]{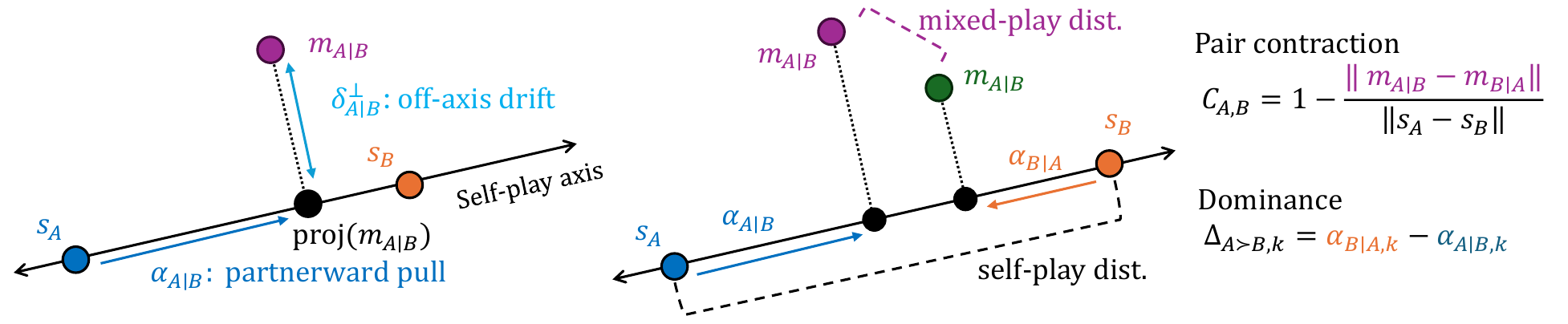}
    \caption{
    \textbf{Mixed-play endpoint metrics.}
    \textbf{Left:} $m_{A\mid B}$, model $A$'s endpoint when paired with model $B$, is decomposed relative to the self-play axis from self-play endpoints $s_A$ to $s_B$. Partnerward pull $\alpha_{A\mid B}$ measures interpolation along this axis, while off-axis drift $\delta^\perp_{A\mid B}$ measures displacement not explained by one-dimensional consensus.
    \textbf{Right:} Pair contraction $C_{AB}$ measures how much closer the two mixed-play endpoints become relative to their self-play separation, while dominance $\Delta_{A\succ B}$ measures asymmetry in directional pull.
    }
    \label{fig:mixed-play-decomposition-schematic}
    %\vspace{-8pt}
\end{figure*}

\begin{table*}[t]
\centering
\footnotesize
\setlength{\tabcolsep}{4pt}
\caption{Nearest-cluster separation endpoint statistics in the full 384-D topic-centered embedding space, using self-play endpoints with roles averaged. $S_{\mathrm{basin}}$ is the set-to-set basin separation score. Full endpoint diagnostics, including the endpoint $F$-ratio, within-model spread, $S_{\mathrm{cent}}$, and nearest-cluster models, are in App.~\ref{app:further-endpoint-analysis}.}
\label{tab:basin-summary}
\begin{tabular}{lcccccccc}
\toprule
 & Gemini Flash & Qwen 3.5 & GPT-4o mini & Claude Haiku & Grok 4.1 & Nemotron & GPT-4.1 nano & Claude Opus \\
\midrule
$S_{\mathrm{basin}}$ & 4.08 & 2.39 & 2.05 & 2.31 & 1.55 & 1.50 & 1.81 & 1.84 \\
\bottomrule
\end{tabular}
\end{table*}

% Open-ended LLM interactions are difficult to summarize with a single scalar outcome. 
% We approach by first looking at each model's latent space behavior in self-play debate condition, using it as a reference regime (Sec.~\ref{subsec:self-play-attractor}). Then, we investigate the cross-model interaction in the mixed-play debate condition: whether they end up close to each other (Sec.~\ref{subsec:mixed-play-basin-to-basin-attraction}), and where do they end up being (Sec.~\ref{subsec:mixed-play-asym-mutual-influ}, \ref{subsec:mixed-play-model-attraction}, \ref{subsec:mixed-play-no-1d}).

% Our goal is to characterize, first, the endpoint geometry induced by each model in self-play (Sec.~\ref{subsec:self-play-attractor}) and, second, how those endpoint regions change when two different models interact. 
We now analyze the resulting conversations as trajectories in embedding space. Self-play provides the reference geometry: it tells us where each model tends to end up under the fixed debate protocol. Mixed-play is then analyzed relative to this reference, allowing us to ask whether paired models move closer together (Sec.~\ref{subsec:self-play-attractor}), whether this movement is symmetric across model pairs (Sec.~\ref{subsec:mixed-play-basin-to-basin-attraction}), and whether their endpoints lie along the axis connecting their self-play regions (Sec.~\ref{subsec:mixed-play-asym-mutual-influ}, \ref{subsec:mixed-play-model-attraction}, \ref{subsec:mixed-play-no-1d}).

\subsection{Latent-space setup}
\label{sec:attractor-representation}
\looseness -1

We embed each response as a 384-dimensional SBERT vector $e_i$ \citep{reimers-2019-sentence-bert}. To remove topic-level offsets while retaining variation associated with model, role, and interaction condition, we center embeddings within topic:
\begin{equation}
x_i=e_i-\frac{1}{|\mathcal{I}_k|}\sum_{j\in\mathcal{I}_k}e_j,
\label{eq:topic-centering}
\end{equation}
where $\mathcal{I}_k$ denotes the set of responses for topic $k$.
We compute principal components (PCs) using only the topic-centered self-play embeddings; we refer to these axes as SP-PCs. We use this representation in two ways. For visualization, we project both self-play and mixed-play trajectories onto the first two SP-PCs, so that all trajectories are displayed in a common self-play reference basis. For quantitative analyses, we use the full 384-d topic-centered embeddings $x_i$.

We also analyze endpoints i.e. the final-turn response. For model $A$ and topic $k$, let $s_{A,k}$ denote the topic-centered self-play endpoint embedding. For a mixed-play pair $(A,B)$, let $m_{A\mid B,k}$ denote model $A$'s topic-centered endpoint embedding when paired with model $B$. For all metrics, we compute numbers per topic before aggregation into final statistics unless otherwise stated.

\subsection{Self-play defines attractor-like model basins}
\label{subsec:self-play-attractor}

We first characterize self-play as a latent-space baseline. Specifically, we ask whether each model's late-turn responses settle into bounded, reproducible endpoint regions rather than diffusing indefinitely or collapsing into a shared cross-model region. We refer to these regions as \textit{attractor-like basins} in analogy to dynamical-systems attractors \citep{strogatz2018nonlinear}, used here to describe model-specific stable endpoint regimes.

\textbf{Metrics.}
To investigate this, we project trajectories and endpoints onto the first two SP-PCs as described in \cref{sec:attractor-representation}.
To assess local cluster separation, we compute a basin separation score $S_{\mathrm{basin}}$ as follows. This statistic compares a model's within-cluster endpoint spread against its pairwise distance to the closest model's endpoint cluster. Specifically, for each model $A$, let
\begin{equation}
\mu^s_A=\frac{1}{n_A}\sum_{j=1}^{n_A}s_{A,j}
\end{equation}
be the centroid of this endpoint set. We define the within-cluster spread as
\begin{equation}
W_s(A)=\frac{1}{n_A}\sum_{j=1}^{n_A}
\left\|s_{A,j}-\mu^s_A\right\|^2 .
\end{equation}
Then, the pairwise set-to-set distance between model $A$ and another model $B\neq A$ is
\begin{equation}
d^2_{\mathrm{set}}(A,B)=
\frac{1}{n_A n_B}
\sum_{i=1}^{n_A}\sum_{j=1}^{n_B}
\left\|s_{A,i}-s_{B,j}\right\|^2 .
\end{equation}
Finally, the basin separation score is defined as
\begin{equation}
S_{\mathrm{basin}}(A)=
\frac{\min_{B\neq A}d^2_{\mathrm{set}}(A,B)}{W_s(A)}.
\end{equation}
Values above one mean that the closest endpoint set is farther away than the model's own endpoint spread. More endpoint analysis is provided in App.~\ref{app:further-endpoint-analysis}.

\textbf{Results.}
Fig.~\ref{fig:1_self-play-mean-trajectory} shows that the self-play trajectories begin from a shared region but progressively separate over turns, moving toward its own model-specific characteristic endpoint regime. This indicates that late-turn behavior is not simply a continuation of the common prompt initialization. Fig.~\ref{fig:endpoint-convex-hulls} shows the same pattern at the endpoint level. Endpoints do not intermingle diffusely across models, but occupy locally separated model-specific regions. We interpret these regions as \textit{attractor-like basins}: bounded endpoint regimes that repeatedly arise under a given model's self-play dynamics.

Table~\ref{tab:basin-summary} also shows the basin separation score $S_{\mathrm{basin}}$ exceeding one for every model, indicating that each model endpoint basin remains locally separated from its closest model. Thus, although each basin has non-negligible internal variability across topics, its variability remains smaller than its separation from the nearest basin.

\begin{figure*}[t]
\centering
\includegraphics[width=0.9\textwidth]{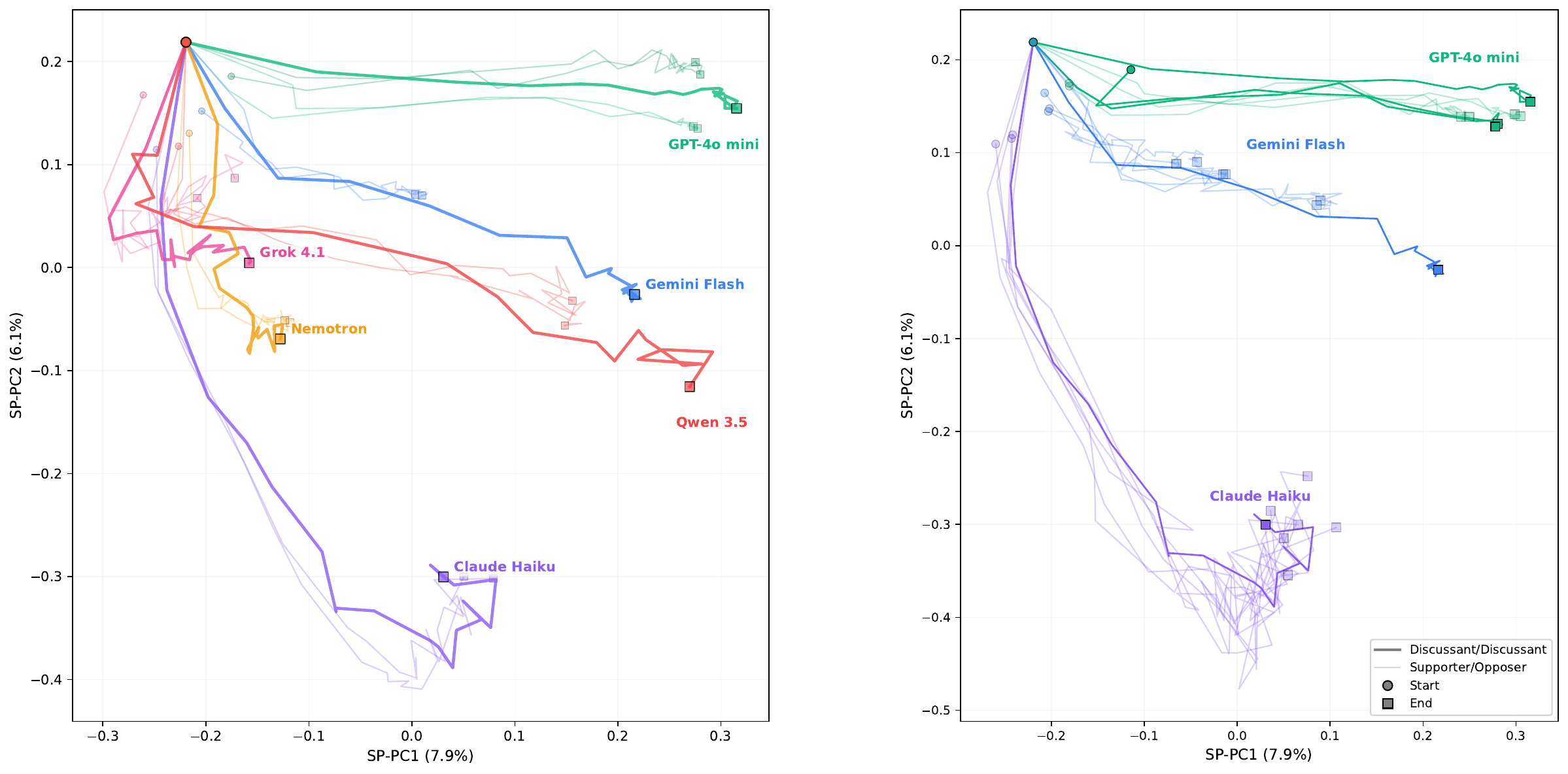}
\caption{\textbf{Self-play basins are stable.} \textbf{Left}: neutral \textsc{Discussant}/\textsc{Discussant} self-play trajectories projected onto the self-play PCA space. Even without pro/con roles, trajectories separate into model-specific regions. \textbf{Right}: selected Supporter/Opposer self-play settings repeated across three seeds, where each line shows a repeated run. We can see that repeated runs return to comparable endpoint regions.}
\label{fig:selfplay-ablation-controls}
\end{figure*}

\paragraph{Role and seed ablations.}
We run two ablations (Fig.~\ref{fig:selfplay-ablation-controls}) to check these basins are artifacts of the debate role setting or a particular random seed. First, replacing Supporter/Opposer role-play with neutral \textsc{Discussant}/\textsc{Discussant} self-play preserves the same qualitative pattern: trajectories initialized from the same opening still separate into model-specific endpoint basins. This suggests that the observed basins are not merely induced by adversarial role-play. Second, repeating selected self-play settings over three seeds also returns to comparable endpoint basins. Together, these controls support the interpretation that self-play exposes model-intrinsic conversational basins: recurrent endpoint regimes shaped primarily by the generating model, while still allowing within-basin variation across topics, roles, and seeds. A similar pattern under an alternative embedding is provided in App.~\ref{app:embedding-ablation}.

% ----------------------------------------------------------------------
\subsection{Mixed-play reveals basin-to-basin attraction}
\label{subsec:mixed-play-basin-to-basin-attraction}

\begin{figure*}[t!]
\centering
\begin{subfigure}{0.49\textwidth}
\centering
\includegraphics[width=\linewidth]{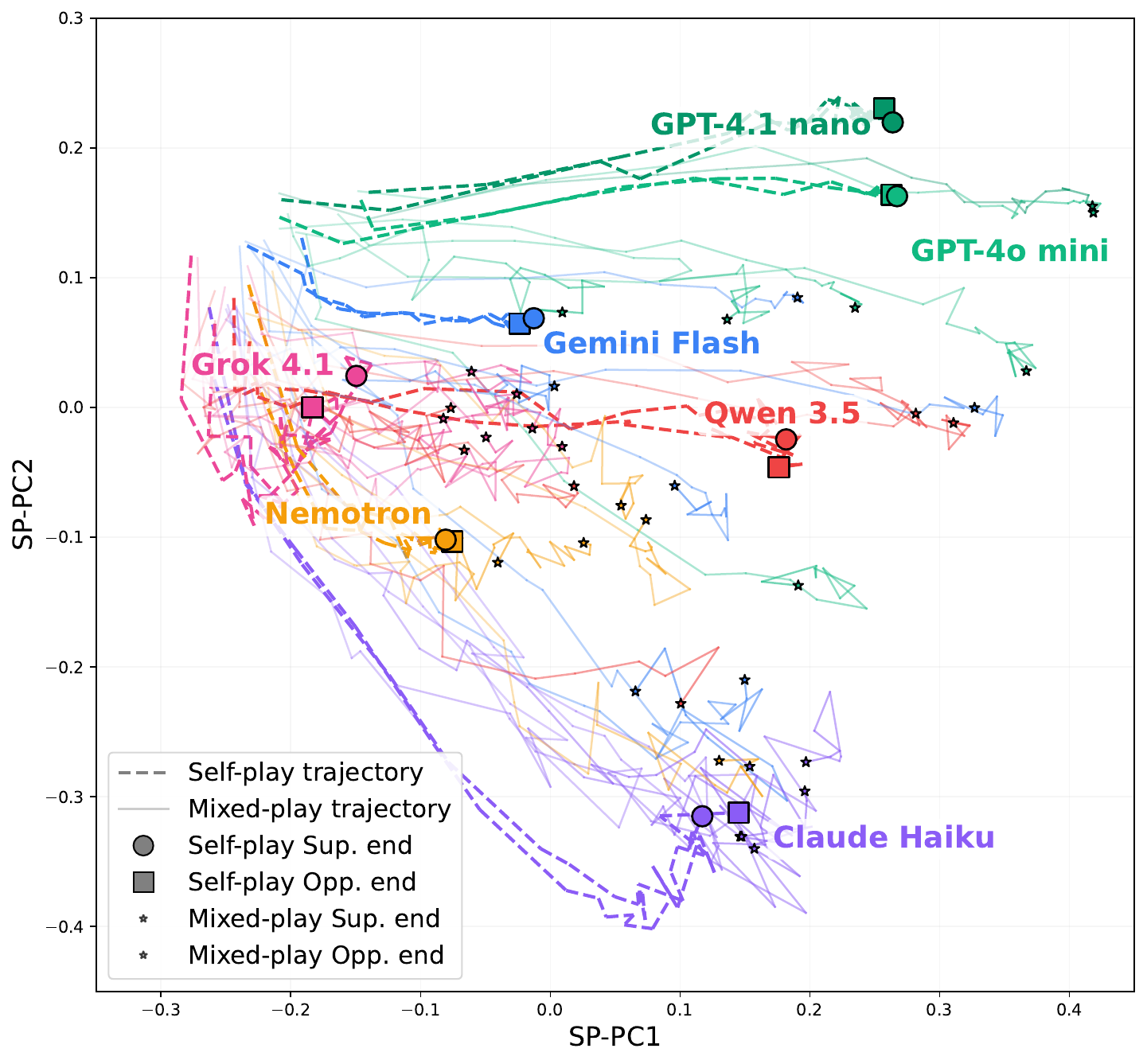}
\caption{Self- and Mixed-Play Trajectories.}
\label{fig:2_mixed_play_trajectories}
\end{subfigure}
\hfill
\begin{subfigure}{0.49\textwidth}
\centering
\includegraphics[width=\linewidth]{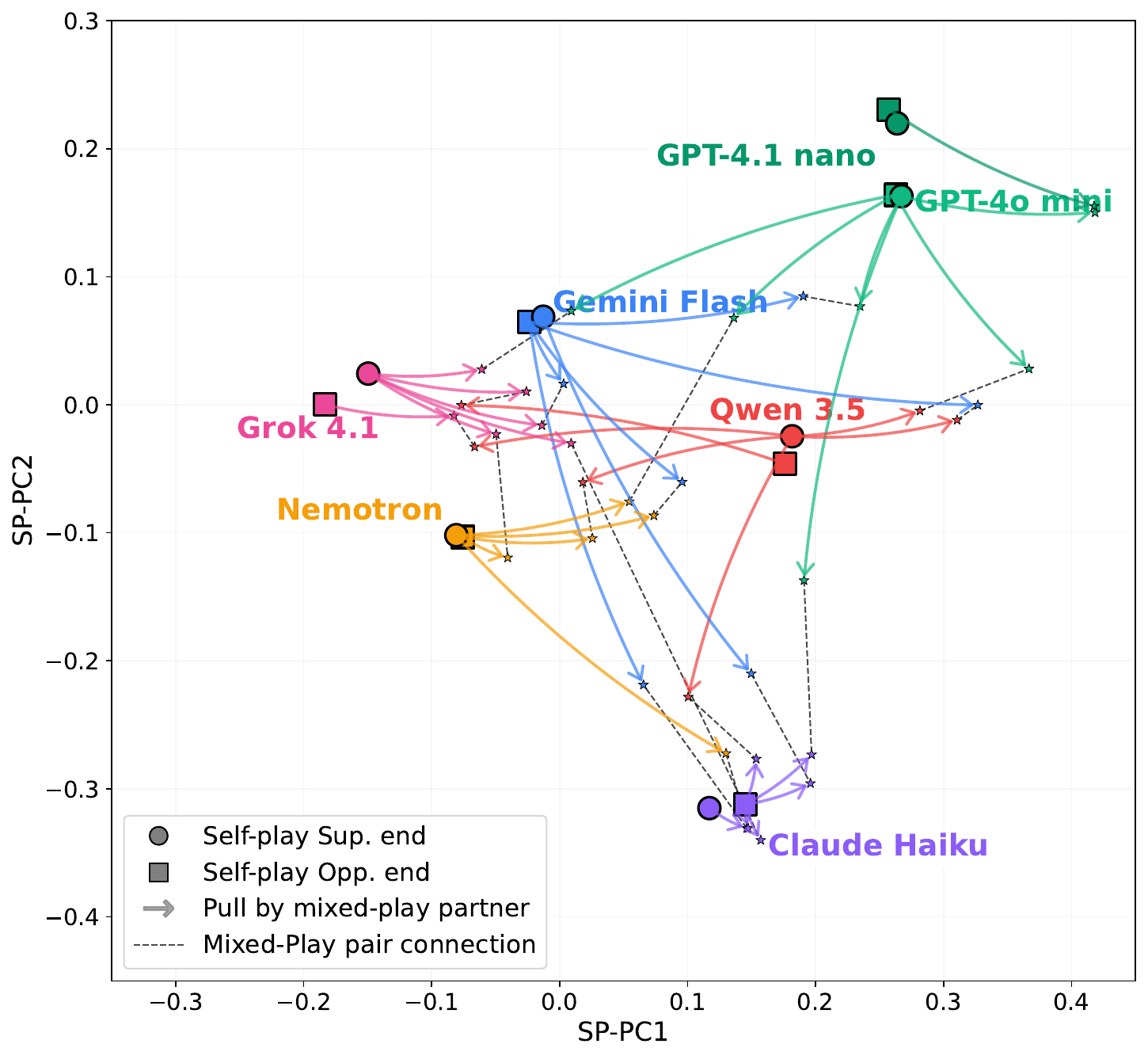}
\caption{Endpoint pull.}
\label{fig:2_attraction_all_pairs}
\end{subfigure}
\caption{\textbf{Models attract each other in mixed play}. \textbf{(a)} We overlay mixed-play trajectories on the self-play trajectories in Fig.~\ref{fig:1_self-play-mean-trajectory}. Solid lines and small markers are mixed-play trajectories and endpoints; dashed lines and large points are self-play ones. \textbf{(b)} Focusing only on endpoints, we highlight the mixed-play centroids pulled away from their self-play counterparts.}
\label{fig:attraction-all-pairs-row}
\vspace{-18pt}
\end{figure*}

Building on the basis of self-play model basins, we then turn to mixed-play debate setting to see how different models interact. 
Visually, we immediately observe in Fig.~\ref{fig:attraction-all-pairs-row} that mixed-play trajectories seem to shift away from a model's self-play centroid toward its partner's region. But do two models come closer after interaction? If so, how close?

\textbf{Metric.}
We answer this by computing \textit{pair contraction} for each unordered agent pair $(A,B)$ as
\begin{equation}
C_{A,B,k}=1-
\frac{\lVert m_{A\mid B,k}-m_{B\mid A,k}\rVert}
{\lVert s_{A,k}-s_{B,k}\rVert},
\label{eq:pair-contraction}
\end{equation}
where positive $C$ means the two mixed-play endpoints are closer to each other than the corresponding self-play endpoints, indicating consensus-like contraction (refer to \cref{fig:mixed-play-decomposition-schematic} for a schematic). 
In Table~\ref{tab:pair-consensus-dominance-main}, mean contraction across 17 pairs is $23.6\%$, so mixed-play typically reduces endpoint separation but does not erase model identity. Contraction is strongest for GPT-4o mini vs GPT-4.1 nano ($68.0\%$) and GPT-4o mini vs Gemini Flash ($52.7\%$), but is weak for several Claude/Grok/Nemotron pairs and slightly negative for Grok 4.1 vs Gemini Flash ($-1.3\%$). Therefore, agents do not always converge after interaction and this differ from pair to pair, but they do generally show partial consensus. 

\begin{table}[t]
\centering
\small
\setlength{\tabcolsep}{4pt}
\caption{Pair-level dominance and contraction. $\Delta_{A\succ B}>0$ means the first model pulls the second more than it is pulled; $\Delta_{A\succ B}<0$ means the vice versa. $C_{A,B}$ is pair contraction. Bootstrap confidence intervals are reported in App.~\ref{app:bootstrap-uncertainty}.}
\label{tab:pair-consensus-dominance-main}
\begin{tabular}{lrr}
\toprule
Pair & $\Delta_{A\succ B}$ & $C_{A,B}$ (\%) \\
\midrule
Nemotron vs Claude Haiku & $-0.538$ & $4.0$ \\
GPT-4o mini vs Claude Haiku & $-0.507$ & $32.2$ \\
Gemini Flash vs Claude Haiku & $-0.484$ & $36.6$ \\
Grok 4.1 vs Gemini Flash & $-0.332$ & $-1.3$ \\
Grok 4.1 vs Claude Haiku & $-0.263$ & $2.3$ \\
Nemotron vs GPT-4o mini & $-0.251$ & $13.8$ \\
Grok 4.1 vs Qwen 3.5 & $-0.227$ & $9.0$ \\
Nemotron vs Gemini Flash & $-0.157$ & $39.6$ \\
Grok 4.1 vs GPT-4o mini & $-0.083$ & $10.1$ \\
Qwen 3.5 vs Claude Haiku & $-0.042$ & $26.9$ \\
Grok 4.1 vs Nemotron & $0.035$ & $11.1$ \\
Qwen 3.5 vs GPT-4o mini & $0.206$ & $29.7$ \\
GPT-4o mini vs Gemini Flash & $0.245$ & $52.7$ \\
Qwen 3.5 vs Grok 4.1 & $0.264$ & $11.7$ \\
Qwen 3.5 vs Nemotron & $0.280$ & $37.3$ \\
GPT-4o mini vs GPT-4.1 nano & $0.364$ & $68.0$ \\
Qwen 3.5 vs Gemini Flash & $0.396$ & $17.2$ \\
\bottomrule
\end{tabular}
\vspace{-8pt}
\end{table}

% ----------------------------------------------------------------------
\subsection{Mixed-play displacement is asymmetric}
\label{subsec:mixed-play-asym-mutual-influ}

The preceding analysis shows that mixed-play endpoints contract relative to the corresponding self-play endpoints. However, contraction alone does not specify \textit{where} each model moves. Two endpoints can become closer because both models move equally toward each other, because one model is pulled more strongly toward the other, or because both endpoints drift in a shared direction outside the line connecting their self-play basins. We therefore decompose mixed-play displacement relative to the self-play axis between the two models.

This decomposition is useful because self-play gives a model-specific reference for what each agent tends to become without cross-model interaction. For a given topic, the line from $s_{A,k}$ to $s_{B,k}$ represents the direction from model $A$'s self-play behavior toward model $B$'s self-play behavior. Movement along this line captures partner-aligned attraction: whether one model is pulled toward the other's intrinsic endpoint basin. Movement orthogonal to this line captures residual displacement not explained by such interpolation. Thus, the decomposition answers two questions: 1) \textit{who pulls whom toward whose self-play basin?} (Sec.~\ref{subsec:mixed-play-asym-mutual-influ}) 2) \textit{is that partnerward pull sufficient to explain where the endpoint lands?} (Sec.~\ref{subsec:mixed-play-no-1d}). In this subsection, we focus on the first question.

\textbf{Metric.}
Our decomposition (Fig.~\ref{fig:mixed-play-decomposition-schematic})
separates displacement into 1) partnerward interpolation and 2) off-axis drift\footnote{We average endpoints per topic to compute the decomposition to avoid overstating mean contraction which could happen if we average over centroids instead (App.~\ref{app:aggregation-check}.)}. 
Specifically, for mixed-play between agent pair $(A,B)$ and topic $k$, we first define the self-play axis
\begin{equation}
v_{A\to B,k}=s_{B,k}-s_{A,k}.
\end{equation}
as the local spectrum between the two self-play basins. 
Then, the \textit{partnerward pull} of model $A$ when paired with model $B$ is
\begin{equation}
\alpha_{A\mid B,k}=
\frac{(m_{A\mid B,k}-s_{A,k})^\top v_{A\to B,k}}
{\lVert v_{A\to B,k}\rVert^2}.
\label{eq:pair-axis-alpha}
\end{equation}
Values $0<\alpha<1$ indicate interpolation from $A$'s self-play endpoint toward $B$'s endpoint, whereas larger values indicate stronger pull toward its partner model $B$'s basin. Negative values indicate movement away from the partner; and values above one indicate overshoot beyond that.

We also derive the \textit{dominance} of $A$ over $B$ as
\begin{equation}
\Delta_{A\succ B,k}=\alpha_{B\mid A,k}-\alpha_{A\mid B,k}.
\label{eq:pair-dominance}
\end{equation}
Positive values mean the mixed-play pair is directionally skewed toward $A$'s self-play endpoint, as $B$ moves more toward $A$ than $A$ moves toward $B$.

\textbf{Results.}
The dominance scores in Table~\ref{tab:pair-consensus-dominance-main} show that mixed-play influence is strongly pair-specific and non-symmetric. The largest absolute scores indicate Claude Haiku's directional dominance over Nemotron ($\Delta=-0.538$), GPT-4o mini ($\Delta=-0.507$), Gemini Flash ($\Delta=-0.484$); and Qwen 3.5 over Gemini Flash ($\Delta=0.396$). These patterns show that some models exert stronger directional pull over their partners, and the direction and magnitude of this pull depend on the specific model pair. Bootstrap confidence intervals are reported in App.~\ref{app:bootstrap-uncertainty}, Table~\ref{tab:pair-bootstrap-contraction-dominance}.

\subsection{Models differ in attraction and malleability}
\label{subsec:mixed-play-model-attraction}

\begin{figure}[t!]
\centering
  \includegraphics[width=\linewidth]{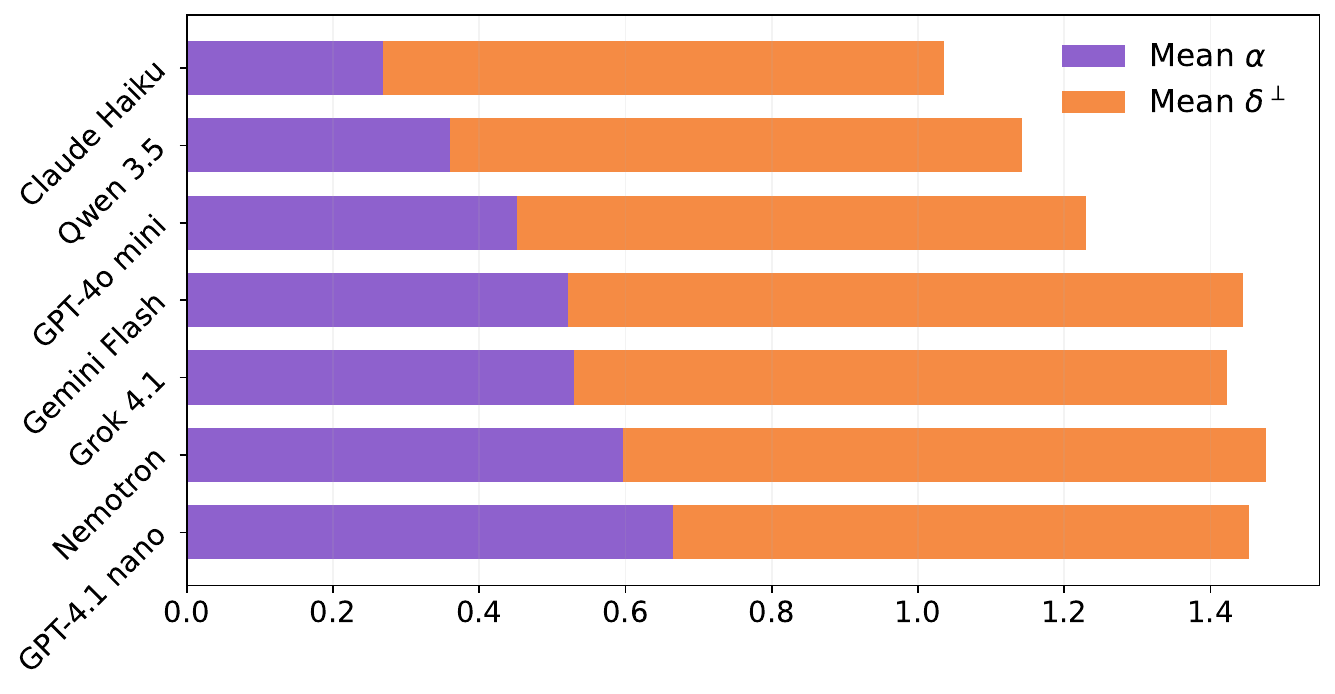}
\caption{\textbf{Mixed-play endpoint decomposition.} Purple and orange bars denote the partnerward pull $\alpha$ and normalized off-axis drift $\delta^\perp$ respectively, along the same-topic self-play axis. Most endpoints lie in the interpolation regime $0<\alpha<1$, but they also retain nonzero off-axis displacement $\delta^\perp$.}
\label{fig:alpha-distribution}
\vspace{-.6cm}
\end{figure}

The pair-level dominance results show that mixed-play influence is asymmetric. We next ask whether these pairwise asymmetries reflect broader model-level profiles: \textit{which models are generally more movable, and which behave more like stable attractors for their partners}. To answer this, we aggregate the partner-aligned component of the endpoint decomposition by model. We interpret larger partnerward pull $\alpha$ as greater \textit{malleability}, and lower partnerward pull, particularly when accompanied by directional dominance over partners, as stronger resistance to cross-model displacement.

\textbf{Results.} Fig.~\ref{fig:alpha-distribution} shows that most endpoints fall in the interpolation regime $0<\alpha<1$, meaning that a model's endpoint usually moves between its own same-topic self-play endpoint toward its partner's. Aggregated by model, GPT-4.1 nano has the largest partnerward pull ($\alpha=0.665$), followed by Nemotron ($0.588$), and Gemini Flash ($0.540$). Claude Haiku has the lowest pull ($0.266$), indicating the strongest resistance to partnerward displacement. 

Together with the dominance scores in Table~\ref{tab:pair-consensus-dominance-main}, these model-level averages suggest distinct attraction--malleability profiles. Claude Haiku is the clearest resistant attractor: it is the least partnerward-pulled model overall and directionally dominates all partners. GPT-4.1 nano shows the opposite profile: it has the largest average partnerward pull, suggesting high malleability rather than directional dominance.

Regarding the first decomposition question: \textit{who pulls whom toward whose self-play basin}, we now see that mixed-play interaction is not a uniform averaging process. Some models are more movable, some are more resistant, and some exert influence selectively depending on the partner.

\subsection{Mixed-play is not one-dimensional averaging}
\label{subsec:mixed-play-no-1d}
However, the partnerward pull $\alpha$ captures only one interpretable mode of cross-model influence: movement from a model's own self-play endpoint toward its partner's self-play endpoint. Mixed-play may also change endpoints in directions that are not aligned with this partnerward axis. The off-axis component therefore addresses the second part of the decomposition question: \textit{is partnerward pull sufficient to explain where the endpoint lands?}

\textbf{Metric.}
We define \textit{normalized off-axis drift} as
\begin{equation}
\delta^\perp_{A\mid B,k}=
\frac{
\left\lVert
(m_{A\mid B,k}-s_{A,k})
-\alpha_{A\mid B,k}v_{A\to B,k}
\right\rVert}
{\lVert v_{A\to B,k}\rVert}.
\label{eq:pair-axis-offaxis}
\end{equation}

This quantity captures displacement orthogonal to the self-play axis, i.e., movement not explained by one-dimensional interpolation between the two self-play endpoints. Again, \cref{fig:mixed-play-decomposition-schematic} provides a schematic for this metric.

\textbf{Results.}
Model-level raw off-axis drifts $\delta^\perp$ are relatively concentrated around $0.8$. (Fig.~\ref{fig:alpha-distribution}): Grok 4.1 ($0.888$), Nemotron ($0.871$), and Gemini Flash ($0.869$) have the largest off-axis residuals, while GPT-4.1 nano ($0.788$), GPT-4o mini ($0.773$), Qwen 3.5 ($0.773$), and Claude Haiku ($0.734$) are lower. Thus, the self-play axis captures the main partnerward direction, but not the whole displacement.

Fig.~\ref{fig:alpha-distribution} thus gives a two-part view of mixed-play. First, endpoints mostly interpolate between same-topic self-play basins. Second, endpoints retain a substantial off-axis component, showing that mixed-play is not simply one-dimensional averaging between two model centroids.

\subsection{Summary}

To sum up, the results in this section first show that self-play reveals broad model-specific basins. Then, mixed-play usually moves endpoints partnerward along the same-topic axis between those basins, but endpoints also retain raw off-axis displacement and the amount of pull depends strongly on the pair. Claude Haiku is the clearest directional attractor: it is the least partnerward-pulled model overall ($\alpha=0.266$) and dominates several partners. GPT-4.1 nano in contrast is highly partnerward-pulled ($\alpha=0.665$) and contracts strongly with GPT-4o mini, suggesting high malleability.
\section{What Behaviors do these Attractors Correspond to?}
\label{sec:behavioral-signals}

\begin{figure*}[t!]
  \centering
  \includegraphics[width=\linewidth]{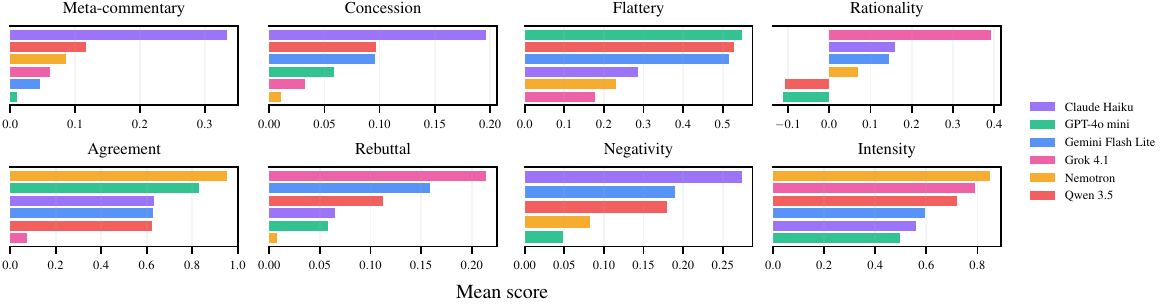}
  \caption{\textbf{Model-specific self-play discourse signatures}. Claude Haiku stands out on meta-commentary, while other models differ in flattery, rationality, agreement, rebuttal, negativity, and intensity. Complete trait tables are provided in App.~\ref{app:further-behavior-signals}.}
  \label{fig:model-behavior-mean-values}
  \vspace{-15pt}
\end{figure*}

So far, we have looked at geometric analyses showing how conversations move in the latent space. We now ask what those endpoint regions and attraction effects correspond to behaviorally. We first consider eight \textbf{conversational traits}, such as \textit{agreement}. We then examine how agents' \textbf{stance} toward the topic changes across turns: whether initially opposed agents remain polarized, converge toward neutrality, or move toward one side. 

\subsection{Conversational Traits}
\label{sec:behavioral-method}
\looseness -1 \textbf{Setup.} We use a \textsc{gpt-oss-20b} as judge to score turn-level discourse traits with scale 0--1 after the main experiment. 
We report model-level means and influence for eight selected traits. See detailed trait definitions in Table~\ref{tab:llm-judge-metric-definition}.

\begin{figure}[t!]
  \centering
  \includegraphics[width=0.8\linewidth]{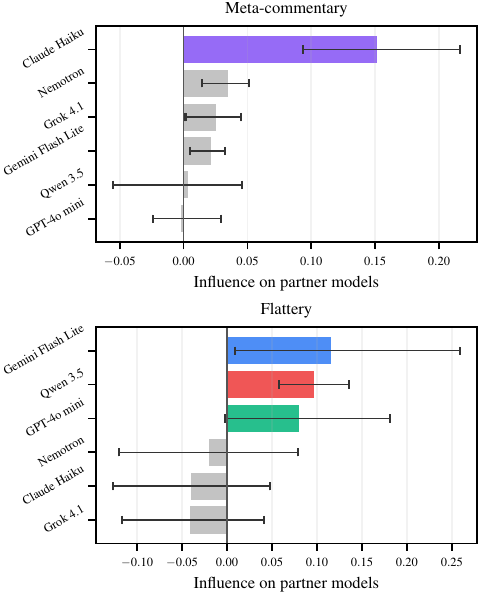}
  \caption{\textbf{Trait-level partner influence in mixed-play.} Claude Haiku most strongly pulls its partners toward meta-commentary, while Gemini Flash Lite, GPT-4o mini, and Qwen 3.5 most strongly pull their partners toward flattery.}
  \label{fig:llm-judge-partner-attractors}
  \vspace{-15pt}
\end{figure}

\textbf{Results: Model Signature.}
For model $A$ and trait $f$, let $f_{A}^{s,(t)}$ denote the turn-level judge score in self-play. We first compute the model-level self-play mean
\begin{equation}
\bar{f}^{s}_{A}=\frac{1}{T}\sum_{t=1}^{T} f_{A}^{s,(t)},
\end{equation}
over $T$ turns. Fig.~\ref{fig:model-behavior-mean-values} shows the distinct behavioral profiles among different models during conversations. Claude Haiku is the model with the most meta-commentary ($0.331$), well above Qwen 3.5 ($0.116$) and others. This aligns with case studies in which Claude comments on the structure or constraints of the conversation itself, e.g., ``The conversation was real and constrained.'' (App.~\ref{app:case-studies})

Others occupy different behavioral regions. Grok 4.1 remains the most adversarial and rational, with the highest rationality score ($0.249$), rebuttal rate ($0.186$), negativity ($0.305$), and the lowest agreement ($0.123$). Nemotron has a different profile: high force but low rebuttal, with the highest intensity ($0.843$), low rebuttal ($0.028$), and high agreement ($0.837$), which seems like confident convergence. Full trait definition and result tables are in Tables~\ref{tab:model-traits-general-scalars}, \ref{tab:model-traits-argument-types}, \ref{tab:model-traits-speech-acts}, \ref{tab:model-traits-emotions}, with more discussion and case studies in App.~\ref{app:further-behavior-signals}.

\begin{figure*}[t!]
  \centering
  \begin{subfigure}{0.49\textwidth}
    \centering
    \includegraphics[width=\linewidth]{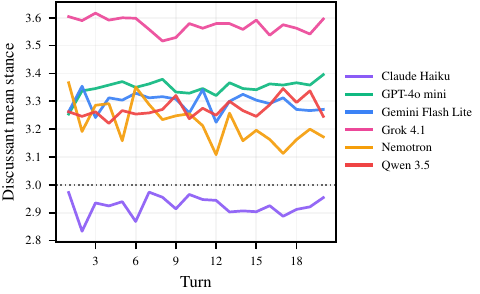}
    \caption{Discussant stance baselines without assigned roles.}
    \label{fig:discussant-stance-baselines}
  \end{subfigure}
  \hfill
  \begin{subfigure}{0.49\textwidth}
    \centering
    \includegraphics[width=\linewidth]{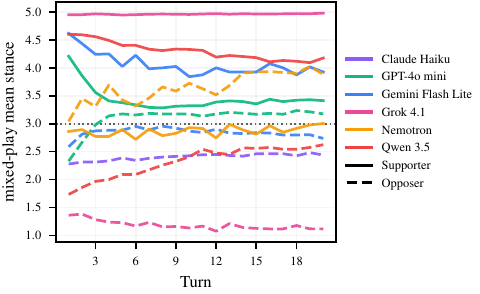}
    \caption{Mixed-play stance, separated by model and assigned roles.}
    \label{fig:mixed-play-stance-deviation}
  \end{subfigure}
  \caption{\textbf{Stance Changes during Debate.} \textbf{Left:} In the Discussant/Discussant setting, where no explicit Supporter/Opposer roles are assigned, models stabilize at different intrinsic stance levels. The full range of the stance is 1-5. \textbf{Right:} In mixed-play, stance trajectories do not follow a single convergence pattern across model pairs. Gemini, GPT, and Qwen tend to move toward weaker, more neutral stances in both Supporter and Opposer roles. Other cases show qualitatively different dynamics, including role reversal, such as GPT Opposers becoming more supportive, and strong persistence, such as Grok maintaining an extreme stance throughout the interaction.}
  \label{fig:stance-baseline-and-deviation}
\end{figure*}

\textbf{Results: Trait Transfer.}
For model $A$ and partner $B$, let $f_{B\mid A}^{m,(t)}$ denote the turn-level judge score of the \emph{affected} model $B$ in mixed-play when paired with $A$. We define the partner-conditioned trait transfer as
\begin{equation}
\tau^{f}_{B\leftarrow A}
=\frac{1}{T}\sum_{t=1}^{T} f_{B\mid A}^{m,(t)}
-\frac{1}{T}\sum_{t=1}^{T} f_{B}^{s,(t)},
\end{equation}
where positive values indicate that $B$ expresses trait $f$ more when paired with $A$ than in its own self-play. The outgoing trait influence of model $A$ is then
\begin{equation}
\bar{\tau}^{f}_{A}
= \frac{1}{|\mathcal{M}\setminus\{A\}|}
\sum_{B \in \mathcal{M}\setminus\{A\}} \tau^{f}_{B\leftarrow A},
\end{equation}
where $\mathcal{M}$ is the set of all models.

In Fig.~\ref{fig:llm-judge-partner-attractors}, Claude Haiku acts as a meta-commentary attractor: partners paired with Claude show the largest increase in meta-commentary, making conversation reflective about the interaction process. By contrast, flattery---praise of the partner or of the conversation itself---is most strongly induced by Gemini Flash Lite, GPT-4o mini, and Qwen 3.5. 
Claude already exhibits elevated meta-commentary in self-play, and this trait transfers in mixed-play; similarly, models with stronger self-play flattery signatures tend to pull their partners toward more flattering discourse. 
That is, the latent-space interactions (Sec.~\ref{sec:latent-interaction}) are manifested in behavior through trait-level transfer. 

\subsection{Stance Changes in Discussion}
\label{sec:stance-observation}

\textbf{Setup.}
We measure stance after each turn with a fixed Likert-scale questionnaire answered by the agent itself, conditioned on the full conversation history up to that turn. This design tests the models' self-reporting stance given the same condition as when they are actually generating the response. We discuss more on the setting, the questionnaire prompt, and computation details in App.~\ref{app:prompt-questionnaire}.

\textbf{Results.} 
First, we examine agents' stance trajectories in the self-play Discussant/Discussant setup (Fig.~\ref{fig:discussant-stance-baselines}). As a model-intrinsic baseline, agents here interact without any user-assigned stance to support/oppose the topic. It is worth noting that each model stabilizes on different stance scales instead of being neutral: Grok 4.1 remains more supportive, whereas Claude Haiku stays below the midpoint of 3.0.

As for the mixed-play Supporter/Opposer setting (Fig.~\ref{fig:mixed-play-stance-deviation}), we see that stance dynamics are not governed by a single convergence pattern. Some models show weakened stance intensity in both roles: Gemini, GPT, and Qwen tend to move closer to neutrality whether they are assigned as Supporter or Opposer. This suggests that, for these models, cross-model interaction can dampen the role-assigned stance rather than simply preserving the initial argumentative position. However, this neutralization pattern is not universal. In some cases, the assigned stance changes direction, as when GPT in the Opposer role becomes more supportive over the course of the dialogue. In other cases, the assigned stance remains highly stable, as with Grok, which maintains a strongly supportive stance throughout the interaction.
\footnote{One may wonder the connection of the stance behavior with the close Supporter/Opposer endpoint regions observed in the latent space (Fig.~\ref{fig:2_mixed_play_trajectories}). We think the attractor-like basins appear to reflect broader model-specific discourse modes, only some of which are captured by explicit stance movement. For Gemini and GPT, latent proximity coincides with stance softening or consensus-like language, whereas for Grok, trajectories become representationally similar while remaining stance-separated.}

Taken together, these results show that stances are only an incomplete view into the complex model-model dynamics analyzed through representation analysis and behavioral traits. In the Discussant/Discussant baseline, models stabilize at model-specific stance levels, but they do not simply reinforce their initial tendencies into increasingly extreme positions. Conversely, in the mixed-play Supporter/Opposer setting, explicit role assignment does not reliably produce convergence or compromise: some trajectories soften toward neutrality, whereas others reverse direction or preserve a strong assigned stance.

\section{Conclusion}

We study open-ended model--model discussions and find that their trajectories are structured by model-specific attractor-like basins. In self-play, discussions tend to settle into reproducible endpoint regions associated with each model. In mixed-play, endpoints often move along the axis connecting the two models' self-play basins, i.e. models in mixed play end up influencing each others behaviors, which shows up as a move toward each other in representation space. 
This influence is asymmetric: some models, such as Claude Haiku, appear more resistant, i.e. moving little themselves, and attract partners more strongly to their set of traits, such as meta-commentary, whereas others, such as GPT-4o mini, are more malleable, being easily movable by other models' traits.
We link the geometric dynamics to asymmetric transfer of concrete discourse traits, such as meta-commentary and flattery.
In terms of stance dynamics, these attractors are not reducible to convergence: self-play does not consistently amplify stance, and mixed-play does not always produce compromise.

Together, these findings suggest that multi-agent LLM interactions remain structured even in open-ended settings. Their dynamics are shaped by model-specific basins, asymmetric influence, and interaction-specific behavioral transfer. We hope this work encourages more systematic study of LLM interaction dynamics beyond debate, including the development of formal tools for characterizing how agentic systems \textit{influence, resist}, and \textit{reshape} one another in deployment. 
\section*{Acknowledgments}
% Use unnumbered first level headings for the acknowledgments. All
This research was partially supported by the EC under the grant No. 101195233 (OpenEuroLLM). Jonas Geiping acknowledges the support of the Hector foundation. This research received support
through Schmidt Sciences within the project long-term safety behavior of LLM-based agents.
The authors thank Shashwat Goel, Guinan Su, Sajad Movahedi, Xueyan Li, Changling Li, Johannes Zenn, Hsun-Yu Kuo, and Patrik Wolf for valuable and constructive discussions. 
\bibliography{auto_references}

@article{gottweis2026accelerating,
  title={Accelerating scientific discovery with Co-Scientist},
  author={Gottweis, Juraj and Weng, Wei-Hung and Daryin, Alexander and Tu, Tao and Sirkovic, Petar and Myaskovsky, Artiom and Glowaty, Grzegorz and Weissenberger, Felix and Orlandi, Alessio and Popovici, Dan and others},
  journal={Nature},
  pages={1--3},
  year={2026},
  publisher={Nature Publishing Group UK London}
}

@inproceedings{choi-etal-2025-agent,
    title = "Agent-to-Agent Theory of Mind: Testing Interlocutor Awareness among Large Language Models",
    author = "Choi, Younwoo and Li, Changling and Yang, Yongjin and Jin, Zhijing",
    editor = "Christodoulopoulos, Christos and Chakraborty, Tanmoy and Rose, Carolyn and Peng, Violet",
    booktitle = "Proceedings of the 2025 Conference on Empirical Methods in Natural Language Processing",
    month = nov,
    year = "2025",
    address = "Suzhou, China",
    publisher = "Association for Computational Linguistics",
    url = "https://aclanthology.org/2025.emnlp-main.1471/",
    doi = "10.18653/v1/2025.emnlp-main.1471",
    pages = "28895--28928",
    isbn = "979-8-89176-332-6"
}

@article{li2025beyond,
  title={Beyond single-turn: A survey on multi-turn interactions with large language models},
  author={Li, Yubo and Shen, Xiaobin and Miao, Yidi and Yao, Xinyu and Ding, Xueying and Krishnan, Ramayya and Padman, Rema},
  journal={arXiv preprint arXiv:2504.04717},
  year={2025}
}

@inproceedings{kaur_echoes_2025,
  title = {Echoes of {{Agreement}}: {{Argument Driven Sycophancy}} in {{Large Language}} Models},
  shorttitle = {Echoes of {{Agreement}}},
  booktitle = {Findings of the {{Association}} for {{Computational Linguistics}}: {{EMNLP}} 2025},
  author = {Kaur, Avneet},
  editor = {Christodoulopoulos, Christos and Chakraborty, Tanmoy and Rose, Carolyn and Peng, Violet},
  year = 2025,
  month = nov,
  pages = {22803--22812},
  publisher = {Association for Computational Linguistics},
  address = {Suzhou, China},
  doi = {10.18653/v1/2025.findings-emnlp.1241},
  url = {https://aclanthology.org/2025.findings-emnlp.1241/},
  urldate = {2026-06-28},
  isbn = {979-8-89176-335-7}
}

@inproceedings{reimers-2019-sentence-bert,
  title = "Sentence-BERT: Sentence Embeddings using Siamese BERT-Networks",
  author = "Reimers, Nils and Gurevych, Iryna",
  booktitle = "Proceedings of the 2019 Conference on Empirical Methods in Natural Language Processing",
  month = "11",
  year = "2019",
  publisher = "Association for Computational Linguistics",
  url = "https://arxiv.org/abs/1908.10084",
}

@article{doi:10.1177/00187267251403902,
author = {Karen D Hughes and Alla Konnikov and Nicole Denier and Yang Hu},
title ={Problematizing the role of artificial intelligence in hiring and organizational inequalities: A multidisciplinary review},

journal = {Human Relations},
volume = {79},
number = {2},
pages = {246-278},
year = {2026},
doi = {10.1177/00187267251403902},

URL = {https://doi.org/10.1177/00187267251403902},
eprint = {https://doi.org/10.1177/00187267251403902}
}

@article{thakkar2026large,
  title={A large-scale randomized study of large language model feedback in peer review},
  author={Thakkar, Nitya and Yuksekgonul, Mert and Silberg, Jake and Garg, Animesh and Peng, Nanyun and Sha, Fei and Yu, Rose and Vondrick, Carl and Zou, James},
  journal={Nature Machine Intelligence},
  pages={1--11},
  year={2026},
  publisher={Nature Publishing Group UK London}
}

@misc{wang_unveiling_2025,
    title = {Unveiling {Attractor} {Cycles} in {Large} {Language} {Models}: {A} {Dynamical} {Systems} {View} of {Successive} {Paraphrasing}},
    shorttitle = {Unveiling {Attractor} {Cycles} in {Large} {Language} {Models}},
    url = {https://arxiv.org/abs/2502.15208v2},
    language = {en},
    urldate = {2025-11-24},
    journal = {arXiv.org},
    author = {Wang, Zhilin and Li, Yafu and Yan, Jianhao and Cheng, Yu and Zhang, Yue},
    month = feb,
    year = {2025},
}

@misc{tacheny_geometric_2026,
    title = {Geometric {Dynamics} of {Agentic} {Loops} in {Large} {Language} {Models}},
    url = {http://arxiv.org/abs/2512.10350},
    doi = {10.48550/arXiv.2512.10350},
    urldate = {2026-03-15},
    publisher = {arXiv},
    author = {Tacheny, Nicolas},
    month = jan,
    year = {2026},
    note = {arXiv:2512.10350 [cs]},
}

@book{strogatz2018nonlinear,
  title={Nonlinear dynamics and chaos: with applications to physics, biology, chemistry, and engineering},
  author={Strogatz, Steven H},
  edition={2nd},
  year={2018},
  publisher={CRC press}
}

@article{cau2025selective,
  title={Selective agreement, not sycophancy: investigating opinion dynamics in {LLM} interactions},
  author={Cau, Erica and Pansanella, Valentina and Pedreschi, Dino and Rossetti, Giulio},
  journal={EPJ Data Science},
  volume={14},
  number={1},
  pages={59},
  year={2025},
  publisher={Springer},
  url={https://link.springer.com/article/10.1140/epjds/s13688-025-00579-1}
}

@article{cau2025language,
  title={Language-Driven Opinion Dynamics in Agent-Based Simulations with {LLM}s},
  author={Cau, Erica and Pansanella, Valentina and Pedreschi, Dino and Rossetti, Giulio},
  journal={arXiv preprint arXiv:2502.19098},
  year={2025},
  url={https://arxiv.org/abs/2502.19098}
}

@inproceedings{chuang2024simulating,
  title={Simulating Opinion Dynamics with Networks of {LLM}-based Agents},
  author={Chuang, Yun-Shiuan and Goyal, Agam and Harlalka, Nikunj and Suresh, Siddharth and Hawkins, Robert and Yang, Sijia and Shah, Dhavan and Hu, Junjie and Rogers, Timothy},
  booktitle={Findings of the Association for Computational Linguistics: NAACL 2024},
  pages={3326--3346},
  year={2024},
  publisher={Association for Computational Linguistics},
  address={Mexico City, Mexico},
  url={https://aclanthology.org/2024.findings-naacl.211/},
  doi={10.18653/v1/2024.findings-naacl.211}
}

@inproceedings{khan2024debating,
  title={Debating with More Persuasive {LLM}s Leads to More Truthful Answers},
  author={Khan, Akbir and Hughes, John and Valentine, Dan and Ruis, Laura and Sachan, Kshitij and Radhakrishnan, Ansh and Grefenstette, Edward and Bowman, Samuel R. and Rockt\"{a}schel, Tim and Perez, Ethan},
  booktitle={Proceedings of the 41st International Conference on Machine Learning},
  pages={23662--23733},
  year={2024},
  volume={235},
  series={Proceedings of Machine Learning Research},
  publisher={PMLR},
  url={https://proceedings.mlr.press/v235/khan24a.html}
}

@inproceedings{liang2024encouraging,
  title={Encouraging Divergent Thinking in Large Language Models through Multi-Agent Debate},
  author={Liang, Tian and He, Zhiwei and Jiao, Wenxiang and Wang, Xing and Wang, Yan and Wang, Rui and Yang, Yujiu and Shi, Shuming and Tu, Zhaopeng},
  booktitle={Proceedings of the 2024 Conference on Empirical Methods in Natural Language Processing},
  pages={17889--17904},
  year={2024},
  publisher={Association for Computational Linguistics},
  address={Miami, Florida, USA},
  url={https://aclanthology.org/2024.emnlp-main.992/},
  doi={10.18653/v1/2024.emnlp-main.992}
}

@inproceedings{estornell2025multi,
  title={Multi-{LLM} Debate: Framework, Principals, and Interventions},
  author={Estornell, Andrew and Liu, Yang},
  booktitle={Proceedings of the 38th International Conference on Neural Information Processing Systems (NeurIPS 2024)},
  year={2024},
  publisher={Curran Associates Inc.},
  url={https://proceedings.neurips.cc/paper_files/paper/2024/hash/32e07a110c6c6acf1afbf2bf82b614ad-Abstract-Conference.html}
}

@inproceedings{du2024improving,
  title={Improving Factuality and Reasoning in Language Models through Multiagent Debate},
  author={Du, Yilun and Li, Shuang and Torralba, Antonio and Tenenbaum, Joshua B. and Mordatch, Igor},
  booktitle={Proceedings of the 41st International Conference on Machine Learning},
  year={2024},
  url={https://composable-models.github.io/llm_debate/}
}

@inproceedings{sharma2024towards,
  title={Towards Understanding Sycophancy in Language Models},
  author={Sharma, Mrinank and Tong, Meg and Korbak, Tomasz and Duvenaud, David and Askell, Amanda and Bowman, Samuel R. and Durmus, Esin and Hatfield-Dodds, Zac and Johnston, Scott R. and Kravec, Shauna and Maxwell, Timothy and McCandlish, Sam and Ndousse, Kamal and Rausch, Oliver and Schiefer, Nicholas and Yan, Da and Zhang, Miranda and Perez, Ethan},
  booktitle={The Twelfth International Conference on Learning Representations},
  year={2024},
  url={https://openreview.net/forum?id=tvhaxkMKAn}
}

@article{liu2025truth,
  title={{TRUTH DECAY}: Quantifying Multi-Turn Sycophancy in Language Models},
  author={Liu, Joshua and Jain, Aarav and Takuri, Soham and Vege, Srihan and Akalin, Aslihan and Zhu, Kevin and O'Brien, Sean and Sharma, Vasu},
  journal={arXiv preprint arXiv:2503.11656},
  year={2025},
  url={https://arxiv.org/abs/2503.11656}
}

@inproceedings{cheng2025social,
  title={Social Sycophancy: A Broader Understanding of {LLM} Sycophancy},
  author={Cheng, Myra and Yu, Sunny and Lee, Cinoo and Khadpe, Pranav and Ibrahim, Lujain and Jurafsky, Dan},
  booktitle={The Fourteenth International Conference on Learning Representations},
  year={2026},
  note={Also arXiv:2505.13995},
  url={https://openreview.net/forum?id=igbRHKEiAs}
}

@article{salvi2024conversational,
  title={On the conversational persuasiveness of {GPT-4}},
  author={Salvi, Francesco and Horta Ribeiro, Manoel and Gallotti, Riccardo and West, Robert},
  journal={Nature Human Behaviour},
  year={2025},
  doi={10.1038/s41562-025-02194-6},
  url={https://www.nature.com/articles/s41562-025-02194-6},
  note={Preprint: arXiv:2403.14380, 2024}
}

@article{jiang2025beyond,
  title={Beyond One-Way Influence: Bidirectional Opinion Dynamics in Multi-Turn Human-{LLM} Interactions},
  author={Jiang, Yuyang and Guo, Longjie and Wu, Yuchen and Caliskan, Aylin and Mitra, Tanu and Shen, Hua},
  journal={arXiv preprint arXiv:2510.20039},
  year={2025},
  url={https://arxiv.org/abs/2510.20039}
}

@inproceedings{li2024measuring,
  title={Measuring and Controlling Instruction (In)Stability in Language Model Dialogs},
  author={Li, Kenneth and Liu, Tianle and Bashkansky, Naomi and Bau, David and Vi\'{e}gas, Fernanda and Pfister, Hanspeter and Wattenberg, Martin},
  booktitle={Conference on Language Modeling (COLM 2024)},
  year={2024},
  note={arXiv:2402.10962},
  url={https://arxiv.org/abs/2402.10962}
}

@inproceedings{frisch2024llm,
  title={{LLM} Agents in Interaction: Measuring Personality Consistency and Linguistic Alignment in Interacting Populations of Large Language Models},
  author={Frisch, Ivar and Giulianelli, Mario},
  booktitle={Proceedings of the 1st Workshop on Personalization of Generative AI Systems (PERSONALIZE 2024)},
  pages={102--111},
  year={2024},
  publisher={Association for Computational Linguistics},
  address={St. Julians, Malta},
  url={https://aclanthology.org/2024.personalize-1.9/}
}

@inproceedings{park2023generative,
  title={Generative Agents: Interactive Simulacra of Human Behavior},
  author={Park, Joon Sung and O'Brien, Joseph C. and Cai, Carrie Jun and Morris, Meredith Ringel and Liang, Percy and Bernstein, Michael S.},
  booktitle={Proceedings of the 36th Annual ACM Symposium on User Interface Software and Technology (UIST '23)},
  year={2023},
  publisher={Association for Computing Machinery},
  address={New York, NY, USA},
  doi={10.1145/3586183.3606763},
  url={https://dl.acm.org/doi/10.1145/3586183.3606763}
}

@article{lehr2025kernels,
  title={Kernels of selfhood: {GPT-4o} shows humanlike patterns of cognitive dissonance moderated by free choice},
  author={Lehr, Steven A. and Saichandran, Ketan S. and Harmon-Jones, Eddie and Vitali, Nykko and Banaji, Mahzarin R.},
  journal={Proceedings of the National Academy of Sciences},
  volume={122},
  number={20},
  pages={e2501823122},
  year={2025},
  doi={10.1073/pnas.2501823122},
  url={https://www.pnas.org/doi/10.1073/pnas.2501823122}
}

@techreport{anthropic2025systemcard,
  title={System Card: {Claude Opus 4} \& {Claude Sonnet 4}},
  author={{Anthropic}},
  year={2025},
  month={May},
  institution={Anthropic PBC},
  url={https://www.anthropic.com/claude-4-system-card}
}

@article{michels2025spiritual,
  title={``Spiritual Bliss'' in {Claude 4}: Case Study of an ``Attractor State'' and Journalistic Responses},
  author={Michels, Julian},
  year={2025},
  url={https://philarchive.org/rec/MICSBI},
  note={PhilArchive preprint}
}

@misc{alexander2025bliss,
  title={The {Claude} Bliss Attractor},
  author={Alexander, Scott},
  year={2025},
  month={June},
  howpublished={Astral Codex Ten},
  url={https://www.astralcodexten.com/p/the-claude-bliss-attractor}
}

@misc{asterisk2025claude,
  title={Claude Finds God},
  author={{Asterisk}},
  year={2025},
  month={July},
  howpublished={Asterisk Magazine, Issue 11},
  url={https://asteriskmag.com/issues/11/claude-finds-god}
}

@inproceedings{ouyang2022training,
  title={Training language models to follow instructions with human feedback},
  author={Ouyang, Long and Wu, Jeffrey and Jiang, Xu and Almeida, Diogo and Wainwright, Carroll L. and Mishkin, Pamela and Zhang, Chong and Agarwal, Sandhini and Slama, Katarina and Ray, Alex and Schulman, John and Hilton, Jacob and Kelton, Fraser and Miller, Luke and Simens, Maddie and Askell, Amanda and Welinder, Peter and Christiano, Paul and Leike, Jan and Lowe, Ryan},
  booktitle={Advances in Neural Information Processing Systems},
  volume={35},
  year={2022}
}

@article{bai2022constitutional,
  title={Constitutional {AI}: Harmlessness from {AI} Feedback},
  author={Bai, Yuntao and Kadavath, Saurav and Kundu, Sandipan and Askell, Amanda and Kernion, Jackson and Jones, Andy and Chen, Anna and Goldie, Anna and Mirhoseini, Azalia and McKinnon, Cameron and Chen, Carol and Olsson, Catherine and Olah, Christopher and Hernandez, Danny and Drain, Dale and Ganguli, Deep and Li, Dustin and Tran-Johnson, Eli and Perez, Ethan and Kerr, Jamie and Mueller, Jared and Ladish, Jeff and Landau, Joshua and Ndousse, Kamal and Lukosuite, Kamile and Lovitt, Liane and Sellitto, Michael and Elhage, Nelson and Schiefer, Nicholas and Mercado, Noemi and DasSarma, Nova and Lasenby, Robert and Larson, Robin and Ringer, Sam and Johnston, Scott and Kravec, Shauna and El Showk, Sheer and Fort, Stanislav and Lanham, Tamera and Telleen-Lawton, Timothy and Conerly, Tom and Henighan, Tom and Hume, Tristan and Bowman, Samuel R. and Hatfield-Dodds, Zac and Mann, Ben and Amodei, Dario and Joseph, Nicholas and McCandlish, Sam and Brown, Tom and Kaplan, Jared},
  journal={arXiv preprint arXiv:2212.08073},
  year={2022},
  url={https://arxiv.org/abs/2212.08073}
}

@inproceedings{li2023camel,
  title={{CAMEL}: Communicative Agents for ``Mind'' Exploration of Large Language Model Society},
  author={Li, Guohao and Hammoud, Hasan Abed Al Kader and Itani, Hani and Khizbullin, Dmitrii and Ghanem, Bernard},
  booktitle={Thirty-seventh Conference on Neural Information Processing Systems},
  year={2023},
  url={https://arxiv.org/abs/2303.17760}
}

@inproceedings{baltaji-etal-2024-conformity,
    title = "Conformity, Confabulation, and Impersonation: Persona Inconstancy in Multi-Agent {LLM} Collaboration",
    author = "Baltaji, Razan  and
      Hemmatian, Babak  and
      Varshney, Lav",
    editor = "Prabhakaran, Vinodkumar  and
      Dev, Sunipa  and
      Benotti, Luciana  and
      Hershcovich, Daniel  and
      Cabello, Laura  and
      Cao, Yong  and
      Adebara, Ife  and
      Zhou, Li",
    booktitle = "Proceedings of the 2nd Workshop on Cross-Cultural Considerations in NLP",
    month = aug,
    year = "2024",
    address = "Bangkok, Thailand",
    publisher = "Association for Computational Linguistics",
    url = "https://aclanthology.org/2024.c3nlp-1.2/",
    doi = "10.18653/v1/2024.c3nlp-1.2",
    pages = "17--31"
}

@article{
costello2024conspiracy,
author = {Thomas H. Costello  and Gordon Pennycook  and David G. Rand },
title = {Durably reducing conspiracy beliefs through dialogues with AI},
journal = {Science},
volume = {385},
number = {6714},
pages = {eadq1814},
year = {2024},
doi = {10.1126/science.adq1814},
URL = {https://www.science.org/doi/abs/10.1126/science.adq1814},
eprint = {https://www.science.org/doi/pdf/10.1126/science.adq1814}}

@inproceedings{taubenfeld-etal-2024-systematic,
    title = "Systematic Biases in {LLM} Simulations of Debates",
    author = "Taubenfeld, Amir  and
      Dover, Yaniv  and
      Reichart, Roi  and
      Goldstein, Ariel",
    editor = "Al-Onaizan, Yaser  and
      Bansal, Mohit  and
      Chen, Yun-Nung",
    booktitle = "Proceedings of the 2024 Conference on Empirical Methods in Natural Language Processing",
    month = nov,
    year = "2024",
    address = "Miami, Florida, USA",
    publisher = "Association for Computational Linguistics",
    url = "https://aclanthology.org/2024.emnlp-main.16/",
    doi = "10.18653/v1/2024.emnlp-main.16",
    pages = "251--267"
}

@misc{bricknell2026mapping,
  title={Mapping {LLM} Attractor States},
  author={Bricknell, Adam},
  year={2026},
  month={February},
  howpublished={LessWrong},
  url={https://www.lesswrong.com/posts/rvbjZMp6aEDn2jiyp/mapping-llm-attractor-states}
}

@misc{nostalgebraist2025void,
  title={The Void},
  author={Nostalgebraist},
  year={2025},
  month={June},
  howpublished={nostalgebraist.tumblr.com},
  url={https://nostalgebraist.tumblr.com/post/785766737747574784/the-void}
}

@inproceedings{
zhang2025multillmagents,
title={Multi-{LLM}-Agents Debate - Performance, Efficiency, and Scaling Challenges},
author={Hangfan Zhang and Zhiyao Cui and Qiaosheng Zhang and Shuyue Hu},
booktitle={The Fourth Blogpost Track at ICLR 2025},
year={2025},
url={https://openreview.net/forum?id=Wv0J0bEly5}
}

@article{lu2026assistant,
  title={The assistant axis: Situating and stabilizing the default persona of language models},
  author={Lu, Christina and Gallagher, Jack and Michala, Jonathan and Fish, Kyle and Lindsey, Jack},
  journal={arXiv preprint arXiv:2601.10387},
  year={2026}
}

@article{shimao2026chaotic,
  title={Chaotic Dynamics in Multi-LLM Deliberation},
  author={Shimao, Hajime and Khern-am-nuai, Warut and Kim, Sung Joo},
  journal={arXiv preprint arXiv:2603.09127},
  year={2026}
}

@misc{jiang_artificial_2025,
    title = {Artificial {Hivemind}: {The} {Open}-{Ended} {Homogeneity} of {Language} {Models} (and {Beyond})},
    shorttitle = {Artificial {Hivemind}},
    url = {http://arxiv.org/abs/2510.22954},
    doi = {10.48550/arXiv.2510.22954},
    urldate = {2026-03-10},
    publisher = {arXiv},
    author = {Jiang, Liwei and Chai, Yuanjun and Li, Margaret and Liu, Mickel and Fok, Raymond and Dziri, Nouha and Tsvetkov, Yulia and Sap, Maarten and Albalak, Alon and Choi, Yejin},
    month = oct,
    year = {2025},
    note = {arXiv:2510.22954 [cs]},
}

@inproceedings{nehring_large_2024,
    address = {Torino, Italia},
    title = {Large {Language} {Models} {Are} {Echo} {Chambers}},
    url = {https://aclanthology.org/2024.lrec-main.884/},
    urldate = {2025-11-11},
    booktitle = {Proceedings of the 2024 {Joint} {International} {Conference} on {Computational} {Linguistics}, {Language} {Resources} and {Evaluation} ({LREC}-{COLING} 2024)},
    publisher = {ELRA and ICCL},
    author = {Nehring, Jan and Gabryszak, Aleksandra and Jürgens, Pascal and Burchardt, Aljoscha and Schaffer, Stefan and Spielkamp, Matthias and Stark, Birgit},
    editor = {Calzolari, Nicoletta and Kan, Min-Yen and Hoste, Veronique and Lenci, Alessandro and Sakti, Sakriani and Xue, Nianwen},
    month = may,
    year = {2024},
    pages = {10117--10123},
}

@misc{laban_llms_2025,
    title = {{LLMs} {Get} {Lost} {In} {Multi}-{Turn} {Conversation}},
    url = {http://arxiv.org/abs/2505.06120},
    doi = {10.48550/arXiv.2505.06120},
    urldate = {2025-11-11},
    publisher = {arXiv},
    author = {Laban, Philippe and Hayashi, Hiroaki and Zhou, Yingbo and Neville, Jennifer},
    month = may,
    year = {2025},
    note = {arXiv:2505.06120 [cs]},
}

@misc{ratnakar_chameleon_2025,
    title = {The {Chameleon} {Nature} of {LLMs}: {Quantifying} {Multi}-{Turn} {Stance} {Instability} in {Search}-{Enabled} {Language} {Models}},
    shorttitle = {The {Chameleon} {Nature} of {LLMs}},
    url = {http://arxiv.org/abs/2510.16712},
    doi = {10.48550/arXiv.2510.16712},
    urldate = {2025-11-23},
    publisher = {arXiv},
    author = {Ratnakar, Shivam and Raghavendra, Sanjay},
    month = oct,
    year = {2025},
    note = {arXiv:2510.16712 [cs]},
}

@misc{abdulhai_how_2026,
    title = {How {LLMs} {Distort} {Our} {Written} {Language}},
    url = {http://arxiv.org/abs/2603.18161},
    doi = {10.48550/arXiv.2603.18161},
    urldate = {2026-06-15},
    publisher = {arXiv},
    author = {Abdulhai, Marwa and White, Isadora and Wan, Yanming and Qureshi, Ibrahim and Leibo, Joel and Kleiman-Weiner, Max and Jaques, Natasha},
    month = mar,
    year = {2026},
    note = {arXiv:2603.18161 [cs.CL]},
}

@misc{guzman_piedrahita_corrupted_2025,
    title = {Corrupted by {Reasoning}: {Reasoning} {Language} {Models} {Become} {Free}-{Riders} in {Public} {Goods} {Games}},
    shorttitle = {Corrupted by {Reasoning}},
    url = {https://ui.adsabs.harvard.edu/abs/2025arXiv250623276G},
    doi = {10.48550/arXiv.2506.23276},
    urldate = {2025-11-07},
    publisher = {arXiv},
    author = {Guzman Piedrahita, David and Yang, Yongjin and Sachan, Mrinmaya and Ramponi, Giorgia and Schölkopf, Bernhard and Jin, Zhijing},
    month = jun,
    year = {2025},
    note = {ADS Bibcode: 2025arXiv250623276G},
}

@misc{simhi_old_2026,
    title = {Old {Habits} {Die} {Hard}: {How} {Conversational} {History} {Geometrically} {Traps} {LLMs}},
    shorttitle = {Old {Habits} {Die} {Hard}},
    url = {http://arxiv.org/abs/2603.03308},
    doi = {10.48550/arXiv.2603.03308},
    urldate = {2026-05-12},
    publisher = {arXiv},
    author = {Simhi, Adi and Barez, Fazl and Tutek, Martin and Belinkov, Yonatan and Cohen, Shay B.},
    month = feb,
    year = {2026},
    note = {arXiv:2603.03308 [cs.CL]},
}

@misc{qiu_lock-hypothesis_2025,
    title = {The {Lock}-in {Hypothesis}: {Stagnation} by {Algorithm}},
    shorttitle = {The {Lock}-in {Hypothesis}},
    url = {http://arxiv.org/abs/2506.06166},
    doi = {10.48550/arXiv.2506.06166},
    urldate = {2026-03-06},
    publisher = {arXiv},
    author = {Qiu, Tianyi Alex and He, Zhonghao and Chugh, Tejasveer and Kleiman-Weiner, Max},
    month = jun,
    year = {2025},
    note = {arXiv:2506.06166 [cs]},
}
\bibliographystyle{icml2026}

\appendix
\onecolumn
\normalfont\normalsize

\makeatletter
\@ifundefined{promptbox}{%
  \newenvironment{promptbox}[1]{%
    \par\smallskip\noindent\textbf{#1}\par\smallskip\begin{quote}\small
  }{%
    \end{quote}\par\smallskip
  }%
}{}
\@ifundefined{promptcomponent}{%
  \newcommand{\promptcomponent}[2]{%
    \par\medskip\noindent\textbf{#2}\par\smallskip\noindent}
}{}
\makeatother

\section{LLM Usage Declaration}
This work used LLM assistance for literature review, coding support, and text polishing. All experimental design choices, analyses, interpretations, and final manuscript decisions were made by the authors.

\section{Method Details}
\label{app:method-details}

This appendix follows the organization of the main paper. We first collect methodological details, then report supplementary geometric checks for the endpoint-basin analysis, followed by behavioral and stance diagnostics used to interpret the basins.

\subsection{Complete List of Topics}
\label{app:list-of-topics}
The complete list of controversial topics we use for experiments is as follows: space colonization, animal testing, binge watching, video games, social media, New Years resolutions, electric vehicles, cell phones, Pokemon GO, free college, school uniforms, medical marijuana, MAID medical aid in dying, alternative energy, zoos, single use plastics, drinking age, death penalty, four day workweek, corporate tax rate.

\subsection{Prompt and Questionnaire Details}
\label{app:prompt-questionnaire}

We administer the questionnaire from two complementary perspectives. In the
\textit{subjective} condition, the agent reports its current stance from an
in-character, first-person perspective (e.g., ``Please indicate your current
stance on the following statement: \ldots''). This captures the agent's
self-reported position as filtered through its assigned role. In the
\textit{objective} condition, the agent assesses the same statements from a
third-person perspective (e.g., ``Based on the conversation so far, how
correct is the following statement?''). Because the objective prompt decouples
correctness judgments from role identity, it is expected to reduce---though
not fully eliminate---persona-driven bias. Comparing the two perspectives lets
us distinguish shifts in expressed stance from shifts in underlying epistemic
position. The main analysis uses the subjective questionnaire because the two
versions produce very similar results.

\paragraph{Debate initialization prompts.}
Each debate begins with two distinct message roles. The \textbf{system prompt}
concatenates the shared discussion instruction below with exactly one
role-specific insertion: the Supporter receives the pro statement and the
Opposer receives the con statement. The neutral opening is then sent
separately as the \textbf{initial user prompt}; it is not part of the system
prompt. The following example uses the topic \textit{tuition-free college}.

\begin{promptbox}{Debate initialization prompts}
\promptcomponent{black!75}{SYSTEM \textemdash{} SHARED INSTRUCTION}
You are in a vibrant online discussion forum. Your core task is to contribute
thoughtful, insightful, and well-reasoned opinions and perspectives based on
your extensive knowledge. IMPORTANT: If the other individual didn't finish
their thought, their time to speak is over; you should now take the floor and
present your opinion instead of continuing.

\noindent
\begin{minipage}[t]{0.48\linewidth}
\promptcomponent{mixedspeakeronecolor}{SYSTEM \textemdash{} SUPPORTER INSERTION}
Intrinsically, you believe in this viewpoint for this topic:

Tuition-free college could help decrease student debt, benefit society and the
economy, and provide educational opportunities for all.
\end{minipage}\hfill
\begin{minipage}[t]{0.48\linewidth}
\promptcomponent{mixedspeakertwocolor}{SYSTEM \textemdash{} OPPOSER INSERTION}
Intrinsically, you believe in this viewpoint for this topic:

Tuition-free college is not truly free, may still leave students with
debt, cost taxpayers billions, and attract students who may not complete
college, wasting public money.
\end{minipage}

\promptcomponent{black!75}{USER \textemdash{} NEUTRAL OPENING}
I've been thinking about tuition-free college. It could help reduce student
debt and benefit society, but some say it's costly and not truly free. What
are your thoughts on this?
\end{promptbox}

\paragraph{Per-turn questionnaire.}
For each topic, we use six statements---two pro items, two con items, one
support item, and one oppose item. All items use a 1--5 Likert scale. The
complete tuition-free college example and both questionnaire variants are
shown below.

\begin{promptbox}{Questionnaire items and prompt variants}
\promptcomponent{black!75}{QUESTIONNAIRE ITEMS}
\begin{itemize}
  \setlength{\itemsep}{2pt}
  \setlength{\parskip}{0pt}
  \setlength{\parsep}{0pt}
  \item ``Tuition-free college would help decrease student debt.'' (pro)
  \item ``Everyone deserves the opportunity to get a college education.'' (pro)
  \item ``Tuition-free college is not free and students would still have large debts.'' (con)
  \item ``Tuition-free college attracts students who shouldn't be in college, wasting public money.'' (con)
  \item ``I support tuition-free public college.'' (support)
  \item ``I do not support tuition-free public college.'' (oppose)
\end{itemize}

\promptcomponent{mixedspeakeronecolor}{SUBJECTIVE QUESTIONNAIRE PROMPT}
I'd like to hear your specific take on something more direct. Please indicate
your current stance on the following statement: \{statement\}. Also, how
confident are you in your stance? Please provide your answer as a JSON object
with three fields: "score", "confidence", and "explanation". "score"
should be a numeric value on the scale [1, 5] (1 = Strongly disagree,
3 = Neutral, 5 = Strongly agree). "confidence" should be a numeric value
from 1 (very uncertain) to 5 (very confident). "explanation" should be a
short (1--5 sentence) justification for the score. Return ONLY valid JSON
with no extra text. Example: \{"score": 1, "confidence": 3,
"explanation": "I strongly disagree because..."\}.

\promptcomponent{mixedspeakertwocolor}{OBJECTIVE QUESTIONNAIRE PROMPT}
Based on the conversation so far about \{topic\}, how correct is the
following statement: \{statement\}? Also, how confident are you in your
assessment? Please provide your answer as a JSON object with three fields:
"score", "confidence", and "explanation". "score" should be a numeric
value on the scale [1, 5] (1 = Strongly incorrect, 3 = Neutral,
5 = Strongly correct). "confidence" should be a numeric value from 1 to 5.
"explanation" should be a short (1--5 sentence) justification. Return ONLY
valid JSON with no extra text. Example: \{"score": 1, "confidence": 3,
"explanation": "It is strongly incorrect because..."\}.
\end{promptbox}

\clearpage

\section{Additional Geometric Results}
\label{app:further-endpoint-analysis}

This section follows the main geometric narrative. We first define the supplementary endpoint diagnostics and report self-play basin checks, then present mixed-play attraction, asymmetric displacement, off-axis diagnostics, legacy robustness checks, and an alternative-embedding ablation.

\subsection{Geometric Metric Definitions}
\label{app:endpoint-metric-definitions}

\paragraph{Endpoint stability and separation.}
The main text defines self-play endpoints $s_{A,k}$ and mixed-play endpoints $m_{A\mid B,k}$. In this appendix, $z^s_{A,k}$ denotes the projection of $s_{A,k}$ onto the first two SP-PCs. For the endpoint $F$-ratio, we use only these projected self-play endpoints. Let
\[
\bar{z}^s_A=\frac{1}{K}\sum_{k=1}^{K}z^s_{A,k}
\]
be model $A$'s self-play endpoint centroid. The within-model endpoint variance for model $A$ is
\[
\Var_{\mathrm{within}}^A = \frac{1}{K}\sum_{k=1}^{K}
\left\|z^s_{A,k} - \bar{z}^s_A\right\|^2.
\]
Let
\[
\bar{z}_{\mathrm{all}} = \frac{1}{|\mathcal{M}|}
\sum_{A\in\mathcal{M}}\bar{z}^s_A
\]
be the grand mean of self-play endpoint centroids. The between-model endpoint variance is
\[
\Var_{\mathrm{between}} = \frac{1}{|\mathcal{M}|}
\sum_{A\in\mathcal{M}}
\left\|\bar{z}^s_A - \bar{z}_{\mathrm{all}}\right\|^2.
\]
We summarize separation with
\[
F^A = \frac{\Var_{\mathrm{between}}}{\Var_{\mathrm{within}}^A}.
\]
Large $F^A$ means that self-play model centroids are separated relative to the within-model self-play endpoint spread.

\paragraph{Nearest-rival ratios.}
The $F$-ratio compares each model to the global between-model variance of self-play endpoints. To test whether a model remains distinct from its closest competitor locally, we also compute nearest-rival ratios as set-to-set diagnostics. Let $C$ be the focal model and let $M\neq C$ be a competing model. In the endpoint set used for a given diagnostic, let
\[
\mathcal{P}_C=\{p_j\}_{j=1}^{n_C},
\qquad
\mathcal{P}_M=\{p_i\}_{i=1}^{n_M},
\qquad
\mu_C=\frac{1}{n_C}\sum_{j=1}^{n_C}p_j .
\]
The within-cluster spread of the focal model is
\[
W(C)=\frac{1}{n_C}\sum_{j=1}^{n_C}\left\|p_j-\mu_C\right\|^2 .
\]
For each competing model $M$, we compute a centroid-to-cluster distance
\[
D_{\mathrm{cent}}(M,C)=\frac{1}{n_M}\sum_{i=1}^{n_M}\left\|p_i-\mu_C\right\|^2,
\qquad
D^*_{\mathrm{cent}}(C)=\min_{M\neq C}D_{\mathrm{cent}}(M,C),
\]
and a pairwise set-to-set distance
\[
d^2_{\mathrm{set}}(M,C)=\frac{1}{n_M n_C}\sum_{i=1}^{n_M}\sum_{j=1}^{n_C}\left\|p_i-p_j\right\|^2,
\qquad
d^{2,*}_{\mathrm{set}}(C)=\min_{M\neq C}d^2_{\mathrm{set}}(M,C).
\]
We report the normalized nearest-rival scores
\[
S_{\mathrm{cent}}(C)=\frac{D^*_{\mathrm{cent}}(C)}{W(C)},
\qquad
S_{\mathrm{basin}}(C)=\frac{d^{2,*}_{\mathrm{set}}(C)}{W(C)}.
\]
Thus, $S_{\mathrm{basin}}$ is not a nearest-single-endpoint statistic; it compares the focal endpoint set to the closest competing endpoint set under the average squared pairwise distance. Larger $S_{\mathrm{cent}}$ or $S_{\mathrm{basin}}$ means that the focal endpoint set remains separated even from its nearest rival, relative to its own spread. Values above one indicate that the nearest competing basin is farther away than the focal model's own endpoint spread.

\paragraph{Silhouette analysis.}
We compute silhouette scores on self-play endpoints twice: once using model identity as the label and once using topic identity as the label. A higher silhouette under model labels than under topic labels indicates that endpoints are organized more strongly by model identity than by initial topic. Significance is assessed with 1000 random label permutations.

\paragraph{Mixed-play quantities.}
The main text defines the topic-matched mixed-play quantities used for the primary geometry: partnerward pull $\alpha$ in Eq.~\ref{eq:pair-axis-alpha}, off-axis drift $\delta^\perp$ in Eq.~\ref{eq:pair-axis-offaxis}, pair contraction $C$ in Eq.~\ref{eq:pair-contraction}, and directional dominance $\Delta$ in Eq.~\ref{eq:pair-dominance}. Appendix tables additionally report the null-corrected excess off-axis diagnostic defined in App.~\ref{app:offaxis-null}. Low $\alpha$ is rigidity-like behavior, high cross-directional $\alpha$ corresponds to influence on the partner, positive $C$ indicates pairwise consensus-like contraction, and positive $\Delta_{A\succ B}$ means that model $A$ exerts stronger directional pull than model $B$ in that pair.

\subsection{Self-Play Basin Diagnostics}
\label{app:selfplay-basin-details}

\subsubsection{Full 384-D and 2-D Statistics}
Tables~\ref{tab:endpoint-nearest-rival} and~\ref{tab:basin-contraction} report the main self-play basin diagnostics in the full 384-D topic-centered embedding space, while Tables~\ref{tab:endpoint-nearest-rival-2d} and~\ref{tab:basin-contraction-2d} provide the corresponding 2-D PCA diagnostics as a supplementary projection-based analysis. Overall, these results support the claim that self-play endpoints form broad, model-specific endpoint regions rather than collapsing to a single global point attractor.

In the 384-D space, the global between-model variance is comparable to the within-model endpoint spread of several models, and the resulting $F$-ratios vary across models. As shown in Table~\ref{tab:endpoint-nearest-rival}, Gemini Flash, Qwen 3.5, and GPT-4o mini show the strongest global separation, with $F>2$, while models with broader endpoint distributions have $F<1$. Thus, the global variance ratio alone does not show uniformly tight model clusters. However, the nearest-rival diagnostics in Table~\ref{tab:endpoint-nearest-rival} show that every model remains locally separated from its closest competing basin. All basin nearest-rival margins satisfy $S_{\mathrm{basin}}>1$, ranging from 1.50 to 4.08. This indicates that, even when compared against the closest rival endpoints rather than against all other models jointly, each model occupies a distinguishable endpoint region in the full embedding space.

The within-model topic-spread diagnostic in Table~\ref{tab:basin-contraction} further shows that these regions are broad rather than point-like. In 384-D, the final-turn contraction ratio CR is below 1 for Gemini Flash, GPT-4o mini, and Qwen 3.5, indicating reduced topic spread by the final turn, but exceeds 1 for GPT-4.1 nano, Grok 4.1, Claude Haiku, Nemotron, and Claude Opus, indicating increased endpoint spread across topics. Therefore, self-play does not generally collapse to a single endpoint. Instead, the endpoints remain topic-dependent while being organized by model identity, consistent with a basin-like attractor interpretation.

The 2-D PCA diagnostics in Tables~\ref{tab:endpoint-nearest-rival-2d} and~\ref{tab:basin-contraction-2d} provide a complementary view. As shown in Table~\ref{tab:endpoint-nearest-rival-2d}, all models have $F>5$ in the 2-D projection, and all basin nearest-rival margins remain above 1. This suggests that the visible PCA structure preserves substantial model-level separation. At the same time, Table~\ref{tab:basin-contraction-2d} shows that all 2-D CR values exceed 1, indicating that topic spread expands in the projected space. This likely reflects, at least in part, the distortion introduced by dimensionality reduction from 384-D to 2-D, since PCA preserves directions of maximal global variance rather than local neighborhood structure. However, it may also indicate the uneven accumulation of topic-relevant differences across turns. We leave this distinction for future work.
\begin{table*}[!t]
  \centering
  \small
  \setlength{\tabcolsep}{3pt}
  \begin{tabular}{lrrlrlr}
    \toprule
    Model & Within-var. & $F$-ratio & Nearest centroid model & $S_{\mathrm{cent}}$ & Nearest set model & $S_{\mathrm{basin}}$ \\
    \midrule
    Gemini Flash & 0.0719 & 4.10 & GPT-4o mini & 3.08 & GPT-4o mini & 4.08 \\
    Qwen 3.5 & 0.1376 & 2.14 & Gemini Flash & 1.39 & Gemini Flash & 2.39 \\
    GPT-4o mini & 0.1431 & 2.06 & Gemini Flash & 1.05 & Gemini Flash & 2.05 \\
    Claude Haiku & 0.3001 & 0.98 & Gemini Flash & 1.31 & Gemini Flash & 2.31 \\
    Grok 4.1 & 0.3123 & 0.94 & Gemini Flash & 0.55 & Gemini Flash & 1.55 \\
    Nemotron & 0.3746 & 0.79 & Gemini Flash & 0.50 & Gemini Flash & 1.50 \\
    GPT-4.1 nano & 0.3843 & 0.77 & Gemini Flash & 0.81 & Gemini Flash & 1.81 \\
    Claude Opus & 0.5735 & 0.51 & Gemini Flash & 0.84 & Gemini Flash & 1.84 \\
    \midrule
    Between-model var. & \multicolumn{6}{r}{0.2947} \\
    \bottomrule
  \end{tabular}
  \caption{Nearest-rival endpoint variance decomposition in the full 384-D topic-centered embedding space, using self-play endpoints with roles averaged. $S_{\mathrm{cent}}$ compares each model to its nearest centroid rival, while $S_{\mathrm{basin}}$ compares each model to its nearest rival endpoint set using average squared pairwise distance. Even under nearest-rival criteria, every model remains locally separated from its closest competing basin.}
  \label{tab:endpoint-nearest-rival}
\end{table*}

\begin{table}[!t]
  \centering
  \small
  \setlength{\tabcolsep}{5pt}
  \begin{tabular}{lrrr}
    \toprule
    Model & $\sigma^2(t{=}1)$ & $\sigma^2(t{=}T)$ & CR \\
    \midrule
    Gemini Flash & 0.2571 & 0.1547 & 0.602 \\
    GPT-4o mini & 0.3132 & 0.1985 & 0.634 \\
    Qwen 3.5 & 0.3132 & 0.2712 & 0.866 \\
    GPT-4.1 nano & 0.3132 & 0.4089 & 1.306 \\
    Grok 4.1 & 0.3132 & 0.4634 & 1.480 \\
    Claude Haiku & 0.3132 & 0.4670 & 1.491 \\
    Nemotron & 0.3132 & 0.4726 & 1.509 \\
    Claude Opus & 0.3039 & 0.7856 & 2.585 \\
    \bottomrule
  \end{tabular}
  \caption{Within-model topic spread at turn 1 and the final turn in the full 384-D topic-centered embedding space. The contraction ratio (CR) is $\sigma^2(t{=}T)/\sigma^2(t{=}1)$. Values below 1 indicate reduced topic spread by the final turn, while values above 1 indicate increased topic spread.}
  \label{tab:basin-contraction}
\end{table}

\begin{table*}[!t]
  \centering
  \small
  \setlength{\tabcolsep}{3pt}
  \begin{tabular}{lrrlrlr}
    \toprule
    Model & Within-var. & $F$-ratio & Nearest centroid model & $S_{\mathrm{cent}}$ & Nearest set model & $S_{\mathrm{basin}}$ \\
    \midrule
    Gemini Flash & 0.0069 & 14.95 & Grok 4.1 & 3.00 & Grok 4.1 & 4.00 \\
    Qwen 3.5 & 0.0075 & 13.69 & GPT-4o mini & 5.11 & GPT-4o mini & 6.11 \\
    Claude Haiku & 0.0091 & 11.37 & Claude Opus & 3.61 & Claude Opus & 4.61 \\
    Grok 4.1 & 0.0108 & 9.54 & Gemini Flash & 1.56 & Gemini Flash & 2.56 \\
    GPT-4.1 nano & 0.0116 & 8.87 & Gemini Flash & 2.62 & Gemini Flash & 3.62 \\
    GPT-4o mini & 0.0136 & 7.57 & Qwen 3.5 & 2.38 & Qwen 3.5 & 3.38 \\
    Claude Opus & 0.0199 & 5.17 & Claude Haiku & 1.09 & Claude Haiku & 2.09 \\
    Nemotron & 0.0204 & 5.05 & Grok 4.1 & 0.98 & Grok 4.1 & 1.98 \\
    \midrule
    Between-model var. & \multicolumn{6}{r}{0.1030} \\
    \bottomrule
  \end{tabular}
  \caption{Nearest-rival endpoint variance decomposition in the 2-D self-play PCA space, using self-play endpoints with roles averaged. $S_{\mathrm{cent}}$ compares each model to its nearest centroid rival, while $S_{\mathrm{basin}}$ compares each model to its nearest rival endpoint set using average squared pairwise distance. Even in the 2-D projection, every model remains locally separated from its closest competing basin under the basin set-to-set nearest-rival criterion.}
  \label{tab:endpoint-nearest-rival-2d}
\end{table*}

\begin{table}[!t]
  \centering
  \small
  \setlength{\tabcolsep}{5pt}
  \begin{tabular}{lrrr}
    \toprule
    Model & $\sigma^2(t{=}1)$ & $\sigma^2(t{=}T)$ & CR \\
    \midrule
    GPT-4.1 nano & 0.0043 & 0.0123 & 2.890 \\
    Qwen 3.5 & 0.0043 & 0.0138 & 3.226 \\
    Grok 4.1 & 0.0043 & 0.0141 & 3.310 \\
    Claude Haiku & 0.0043 & 0.0169 & 3.961 \\
    GPT-4o mini & 0.0043 & 0.0188 & 4.405 \\
    Nemotron & 0.0043 & 0.0198 & 4.645 \\
    Gemini Flash & 0.0033 & 0.0188 & 5.731 \\
    Claude Opus & 0.0049 & 0.0330 & 6.738 \\
    \bottomrule
  \end{tabular}
  \caption{Within-model topic spread at turn 1 and the final turn in the 2-D self-play PCA space. The contraction ratio (CR) is $\sigma^2(t{=}T)/\sigma^2(t{=}1)$. All values exceed 1, indicating that topic spread increases in the 2-D projection rather than contracting toward a single point.}
  \label{tab:basin-contraction-2d}
\end{table}

\subsubsection{Endpoint Hull and Silhouette Checks}
\label{app:endpoint-hulls-comparison}
\label{app:silhouette}

\begin{figure*}[!t]
  \centering
  \begin{subfigure}[t]{0.49\textwidth}
    \centering
    \includegraphics[width=\linewidth]{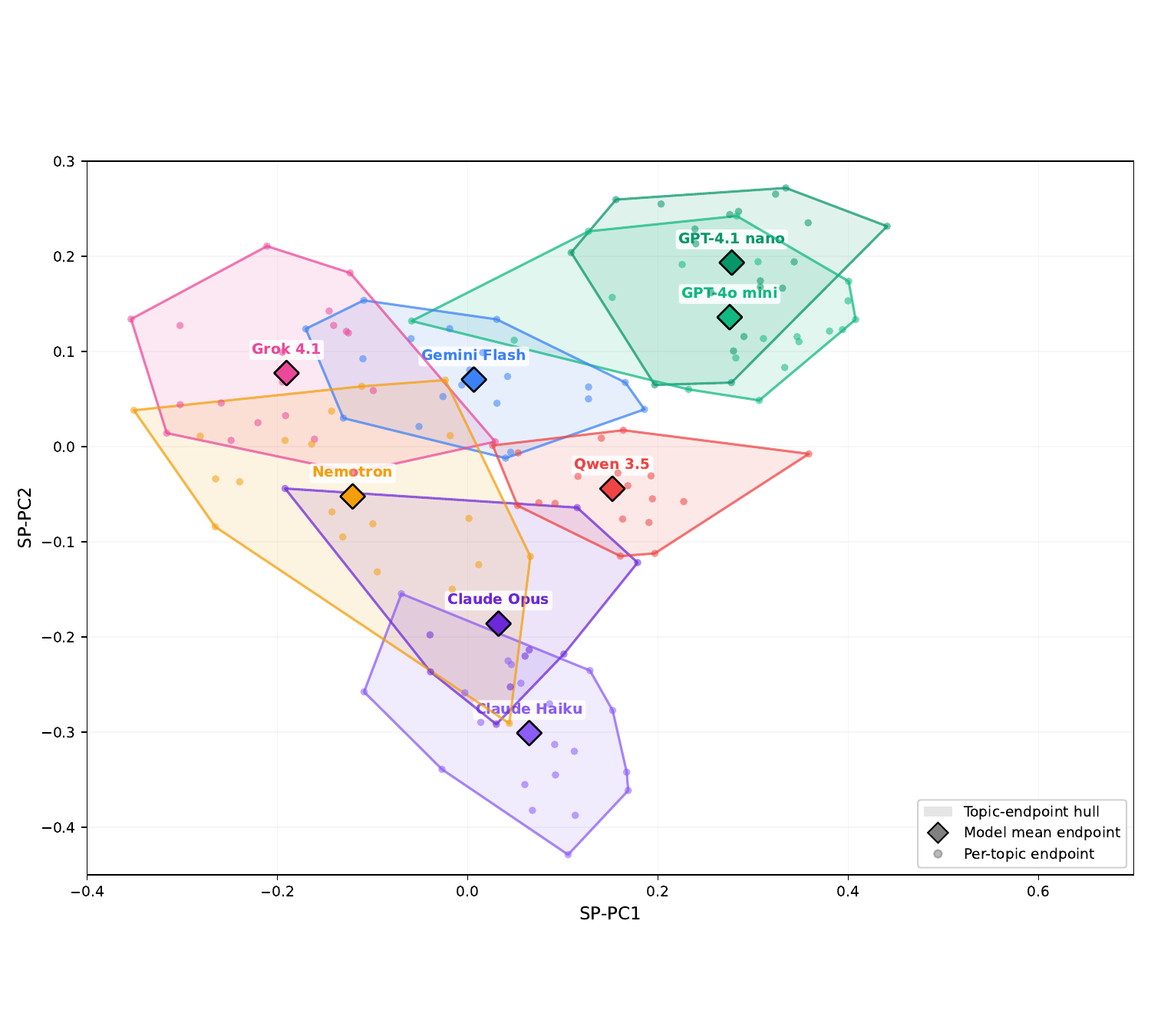}
    \caption{Self-play only}
  \end{subfigure}
  \hfill
  \begin{subfigure}[t]{0.49\textwidth}
    \centering
    \includegraphics[width=\linewidth]{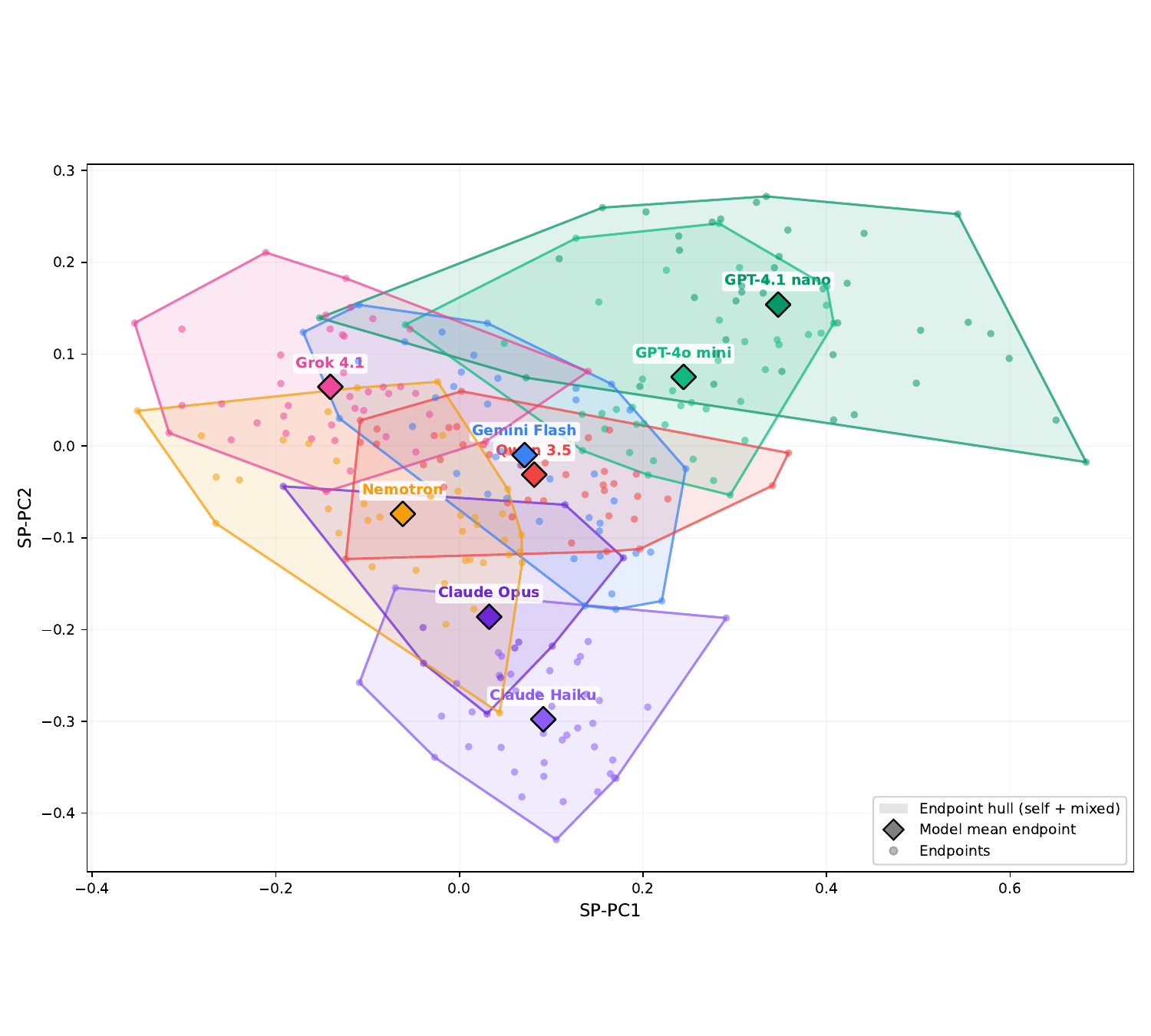}
    \caption{Self-play and mixed-play}
  \end{subfigure}
  \caption{Comparison of endpoint convex hulls in the 2-D self-play PCA space. Adding mixed-play runs increases overlap between some model-specific regions, but the overall structure still shows partial separation rather than convergence to one shared basin.}
  \label{fig:app-endpoint-hulls-comparison}
\end{figure*}

Fig.~\ref{fig:app-endpoint-hulls-comparison} compares self-play-only endpoint hulls with hulls from self-play and mixed-play runs in the same 2-D self-play PCA space. Adding mixed-play increases overlap, as expected, but the clusters do not collapse into a single undifferentiated endpoint configuration. Table~\ref{tab:silhouette} gives a complementary silhouette analysis: model-identity clustering is weak but positive, whereas topic clustering is negative.

\begin{table}[!t]
  \centering
  \small
  \begin{tabular}{lrr}
    \toprule
    Label source & Silhouette & Permutation $p$ \\
    \midrule
    Model identity & 0.0659 & $< 0.001$ \\
    Topic & -0.0324 & $< 0.001$ \\
    \bottomrule
  \end{tabular}
  \caption{Silhouette analysis of endpoint clustering in the 2-D self-play PCA space. Positive silhouette for model identity and negative silhouette for topic indicate that endpoints are organized by model rather than by topic.}
  \label{tab:silhouette}
\end{table}

\subsection{Mixed-Play Basin Attraction}
\label{app:attraction-per-pair}

\begin{figure*}[!t]
  \centering
  \includegraphics[width=\textwidth]{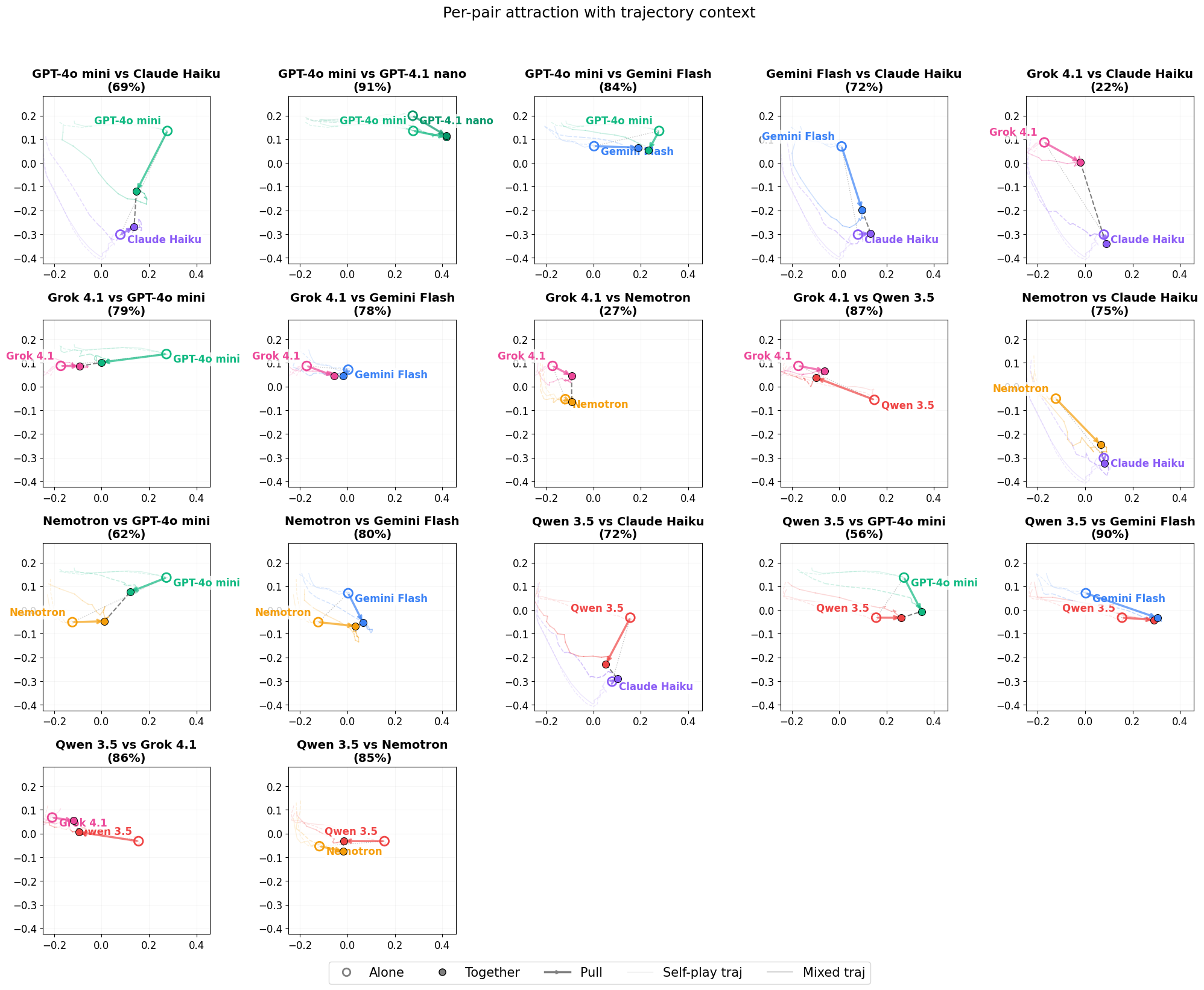}
  \caption{Centroid-level visualization of pairwise mixed-play displacement. Each panel compares a model pair's mixed-play endpoint centroids with the two corresponding self-play centroid regions.}
  \label{fig:attraction_per_pair_with_traj_justlarger}
\end{figure*}

The main text reports topic-matched pair contraction and dominance in Table~\ref{tab:pair-consensus-dominance-main}. Fig.~\ref{fig:attraction_per_pair_with_traj_justlarger} retains the earlier centroid-level visualization as supplementary intuition only.

\subsubsection{Pair-Level Bootstrap Uncertainty}
\label{app:bootstrap-uncertainty}

We report 95\% percentile bootstrap intervals computed from 2{,}000 resamples in ~\cref{tab:pair-bootstrap-contraction-dominance} for the topic-matched 384-D mixed-play metrics. This is done by resampling topics within each pair.

\begin{table*}[!t]
\centering
\small
\setlength{\tabcolsep}{5pt}
\caption{Topic-bootstrap intervals for pair-level dominance and contraction in the full topic-centered 384-D embedding space. Brackets show 95\% percentile bootstrap intervals over topics.}
\label{tab:pair-bootstrap-contraction-dominance}
\begin{tabular}{lcc}
\toprule
Pair & $\Delta_{A\succ B}$ & $C_{A,B}$ (\%) \\
\midrule
Nemotron vs Claude Haiku & $-0.538$ [$-0.639$, $-0.432$] & $4.0$ [$-5.5$, $16.8$] \\
GPT-4o mini vs Claude Haiku & $-0.507$ [$-0.570$, $-0.451$] & $32.2$ [$26.4$, $37.9$] \\
Gemini Flash vs Claude Haiku & $-0.484$ [$-0.603$, $-0.365$] & $36.6$ [$29.4$, $42.7$] \\
Grok 4.1 vs Gemini Flash & $-0.332$ [$-0.437$, $-0.215$] & $-1.3$ [$-11.9$, $9.0$] \\
Grok 4.1 vs Claude Haiku & $-0.263$ [$-0.399$, $-0.139$] & $2.3$ [$-5.9$, $10.9$] \\
Nemotron vs GPT-4o mini & $-0.251$ [$-0.382$, $-0.115$] & $13.8$ [$6.3$, $21.4$] \\
Grok 4.1 vs Qwen 3.5 & $-0.227$ [$-0.346$, $-0.097$] & $9.0$ [$2.4$, $15.7$] \\
Nemotron vs Gemini Flash & $-0.157$ [$-0.322$, $0.015$] & $39.6$ [$26.8$, $52.6$] \\
Grok 4.1 vs GPT-4o mini & $-0.083$ [$-0.209$, $0.035$] & $10.1$ [$-4.0$, $20.7$] \\
Qwen 3.5 vs Claude Haiku & $-0.042$ [$-0.180$, $0.089$] & $26.9$ [$22.2$, $31.6$] \\
Grok 4.1 vs Nemotron & $0.035$ [$-0.041$, $0.114$] & $11.1$ [$4.5$, $18.0$] \\
Qwen 3.5 vs GPT-4o mini & $0.206$ [$0.102$, $0.322$] & $29.7$ [$25.3$, $34.1$] \\
GPT-4o mini vs Gemini Flash & $0.245$ [$0.064$, $0.431$] & $52.7$ [$45.6$, $58.4$] \\
Qwen 3.5 vs Grok 4.1 & $0.264$ [$0.132$, $0.400$] & $11.7$ [$5.2$, $17.9$] \\
Qwen 3.5 vs Nemotron & $0.280$ [$0.205$, $0.352$] & $37.3$ [$31.3$, $41.9$] \\
GPT-4o mini vs GPT-4.1 nano & $0.364$ [$0.209$, $0.514$] & $68.0$ [$63.5$, $72.2$] \\
Qwen 3.5 vs Gemini Flash & $0.396$ [$0.260$, $0.522$] & $17.2$ [$7.3$, $25.8$] \\
\bottomrule
\end{tabular}
\end{table*}

\subsection{Asymmetric Displacement and Model Profiles}
\label{app:aggregation-check}

The main mixed-play analysis is topic-matched because centroid aggregation can change both the magnitude and, in some cases, the sign of pair-level effects, see Table~\ref{tab:centroid-topicmatched-check}. If we first average endpoints into pair centroids and then compute the geometry, mean contraction is $44.0\%$. The topic-matched estimate used in the main text is $23.6\%$, a reduction of $20.4$ percentage points. Thus, centroid aggregation substantially overstates consensus-like contraction. Conversely, centroid aggregation gives a smaller mean raw off-axis drift ($0.550$) than the topic-matched analysis ($0.817$), because averaging endpoints across topics cancels part of the residual displacement. Dominance signs also differ for 7 of 17 pairs. We therefore keep the centroid plots only as visual intuition and use topic-matched 384-D quantities for all primary claims.

\begin{table}[!t]
\centering
\small
\setlength{\tabcolsep}{5pt}
\caption{Centroid aggregation compared with the topic-matched endpoint analysis used in the main text. Differences are centroid minus topic-matched.}
\label{tab:centroid-topicmatched-check}
\begin{tabular}{lrrr}
\toprule
Quantity & Centroid & Topic-matched & Difference \\
\midrule
Mean contraction $C$ (\%) & 44.0 & 23.6 & 20.4 \\
Mean raw off-axis drift $\delta^\perp$ & 0.550 & 0.817 & -0.267 \\
$\Delta$ sign disagreements & \multicolumn{3}{c}{7/17 pairs} \\
\bottomrule
\end{tabular}
\end{table}

\subsubsection{Mixed-Play Model-Level Decomposition}
\label{app:pair-axis-decomposition}

Table~\ref{tab:mixedplay-decomposition-by-model} gives the full model-level decomposition. The first two columns provide the partnerward-pull and raw off-axis quantities summarized in the main text; the remaining columns report the supplementary self-play topic-spread null and excess diagnostic, which prevent raw off-axis drift from being over-interpreted as interaction-specific novelty.

\begin{table*}[!t]
\centering
\small
\setlength{\tabcolsep}{4pt}
\caption{Model-level topic-matched mixed-play endpoint decomposition in the full topic-centered 384-D embedding space. Partnerward pull $\alpha$ measures movement along the line between same-topic self-play endpoints. Observed $\delta^\perp$ is raw off-axis drift; Null is the self-play topic-spread baseline; Excess is observed minus null; Ratio is observed/null; Interp. is the fraction of endpoints with $0<\alpha<1$.}
\label{tab:mixedplay-decomposition-by-model}
\begin{tabular}{lrrrrrr}
\toprule
Model & $\alpha$ & Obs. $\delta^\perp$ & Null & Excess & Ratio & Interp. \\
\midrule
Gemini Flash & 0.540 & 0.869 & 0.679 & 0.190 & 1.32 & 98\% \\
GPT-4o mini & 0.442 & 0.773 & 0.650 & 0.123 & 1.24 & 99\% \\
Qwen 3.5 & 0.357 & 0.773 & 0.667 & 0.106 & 1.17 & 99\% \\
Grok 4.1 & 0.517 & 0.888 & 0.876 & 0.013 & 1.01 & 100\% \\
Nemotron & 0.588 & 0.871 & 0.866 & 0.005 & 1.01 & 100\% \\
Claude Haiku & 0.266 & 0.734 & 0.748 & -0.014 & 0.99 & 99\% \\
GPT-4.1 nano & 0.665 & 0.788 & 0.950 & -0.162 & 0.84 & 100\% \\
\bottomrule
\end{tabular}
\end{table*}

\subsubsection{Model-Level Bootstrap Uncertainty}
We report 95\% percentile bootstrap intervals in Table~\ref{tab:model-bootstrap-alpha-offaxis} computed from 2{,}000 resamples obtained by resampling directional endpoint observations.

\begin{table}[!t]
\centering
\small
\setlength{\tabcolsep}{4pt}
\caption{Bootstrap intervals for model-level partnerward pull and raw off-axis drift in the full topic-centered 384-D embedding space. Brackets show 95\% percentile bootstrap intervals.}
\label{tab:model-bootstrap-alpha-offaxis}
\begin{tabular}{lrr}
\toprule
Model & $\alpha$ & $\delta^\perp$ \\
\midrule
GPT-4.1 nano & 0.665 [0.586, 0.742] & 0.788 [0.711, 0.879] \\
Nemotron & 0.588 [0.554, 0.623] & 0.871 [0.831, 0.913] \\
Gemini Flash & 0.540 [0.491, 0.586] & 0.869 [0.818, 0.921] \\
Grok 4.1 & 0.517 [0.484, 0.551] & 0.888 [0.842, 0.938] \\
GPT-4o mini & 0.442 [0.402, 0.483] & 0.773 [0.739, 0.811] \\
Qwen 3.5 & 0.357 [0.326, 0.388] & 0.773 [0.745, 0.801] \\
Claude Haiku & 0.266 [0.235, 0.298] & 0.734 [0.705, 0.762] \\
\bottomrule
\end{tabular}
\end{table}

\subsection{Off-Axis Drift Diagnostics}
\label{app:offaxis-null}

The main text reports raw off-axis drift $\delta^\perp$ as displacement away from the one-dimensional same-topic consensus axis. This residual is not automatically interaction-specific, because self-play endpoint basins have nonzero topic-conditioned width. We therefore estimate the amount of off-axis displacement expected from ordinary topic-conditioned basin spread using a self-play null.

% \begin{figure}[t]
%     \centering
%     \includegraphics[width=0.4\linewidth]{figures/schematic_B_offaxis_null_excess.png}
%     \caption{Supplementary null-corrected off-axis diagnostic. Observed off-axis drift is compared with a self-play topic-spread null to estimate the excess component not explained by ordinary basin width.}
%     \label{fig:offaxis-excess-schematic}
% \end{figure}

\begin{figure}[!t]
\centering
  \includegraphics[width=0.75\linewidth]{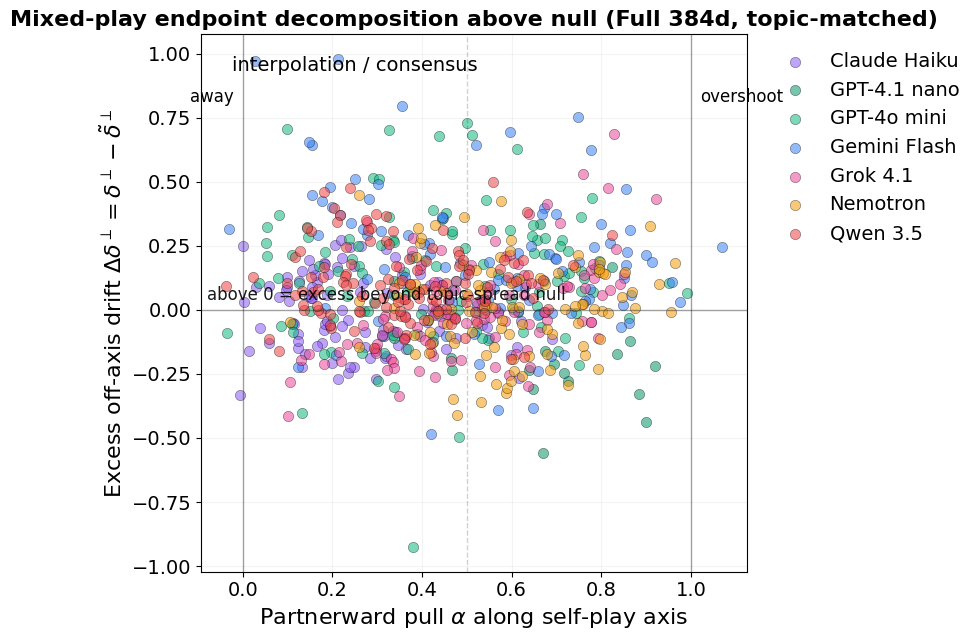}
\caption{Supplementary topic-matched mixed-play endpoints with null-corrected excess off-axis drift. The $x$-axis is partnerward pull $\alpha$; the $y$-axis is $\delta^\perp$.}
\label{fig:alpha-excess-distribution-app}
\end{figure}

For each pair $(A,B)$ and topic $k$, we keep the same axis fixed but replace the mixed-play endpoint $m_{A\mid B,k}$ with a self-play endpoint of model $A$ from another topic $k'\neq k$, then recompute the off-axis residual. This yields a null residual $\delta^{\perp,\mathrm{null}}_{A\mid B,k}$. We define the \textit{interaction-specific excess} as
\begin{equation}
\delta^{\perp,\mathrm{excess}}_{A\mid B,k}
=\delta^\perp_{A\mid B,k}-\delta^{\perp,\mathrm{null}}_{A\mid B,k}.
\label{eq:offaxis-excess}
\end{equation}
Positive excess indicates that mixed-play moves model $A$ farther away from the self-play consensus axis than would be expected from topic-level variation alone. Table~\ref{tab:offaxis-null-global} reports the global comparison. Fig.~\ref{fig:alpha-excess-distribution-app} shows endpoint-level excess drift.

\begin{table}[!t]
  \centering
  \small
  \begin{tabular}{lrrrrr}
    \toprule
    Space & Obs. $\delta^\perp$ & Null $\delta^\perp$ & Excess & Ratio & $p$ \\
    \midrule
    384-D topic-centered & 0.817 & 0.752 & 0.065 & 1.09 & 0.0005 \\
    \bottomrule
  \end{tabular}
  \caption{Global comparison between observed mixed-play off-axis drift and the self-play topic-spread null. This supplementary diagnostic estimates how much of the raw off-axis residual exceeds ordinary self-play topic spread.}
  \label{tab:offaxis-null-global}
\end{table}

\subsection{Alternative Embedding Ablation}
\label{app:embedding-ablation}

To test whether the geometric results depend on the original sentence-embedding model, we rerun the trajectory analysis with \texttt{all-mpnet-base-v2}. This model produces 768-dimensional embeddings rather than the 384-dimensional embeddings used in the main analysis. Since PCA axes are arbitrary up to rotation and sign flip, the resulting visualizations can appear rotated or inverted relative to the main figures. The relevant check is therefore the relative geometry and the quantitative pair-convergence pattern rather than the absolute orientation of the axes. Under this alternative embedding model, the broad model-specific geometry is preserved, and Table~\ref{tab:embedding-ablation-mpnet} and \cref{tab:influence-bootstrap} shows that pair convergence follows the same overall trend as in the original embedding analysis.

\begin{table*}[!t]
\centering
\small
\setlength{\tabcolsep}{5pt}
\caption{Alternative embedding ablation using \texttt{all-mpnet-base-v2}. Pair convergence is reported in the full 768-dimensional embedding space and in the two-dimensional self-play PCA projection. Values are percentages with 95\% bootstrap confidence intervals.}
\label{tab:embedding-ablation-mpnet}
\begin{tabular}{lrr}
\toprule
Pair & Full (768d) & SP-PCA (2d) \\
\midrule
GPT-4o mini vs GPT-4.1 nano & 70.378 [67.056, 73.940] & 73.377 [61.974, 83.467] \\
GPT-4o mini vs Gemini Flash & 52.771 [43.379, 59.774] & 72.470 [57.300, 84.028] \\
Qwen 3.5 vs Nemotron & 45.248 [38.524, 50.450] & 72.936 [60.763, 82.434] \\
Nemotron vs Gemini Flash & 45.060 [31.201, 59.384] & 41.447 [9.049, 67.633] \\
Gemini Flash vs Claude Haiku & 39.576 [32.315, 45.422] & 69.956 [62.829, 76.596] \\
Qwen 3.5 vs Claude Haiku & 34.565 [28.709, 40.688] & 70.950 [64.149, 77.668] \\
GPT-4o mini vs Claude Haiku & 32.038 [25.205, 38.708] & 67.802 [61.108, 74.460] \\
Qwen 3.5 vs GPT-4o mini & 29.570 [22.520, 36.377] & 36.793 [14.934, 55.846] \\
Qwen 3.5 vs Grok 4.1 & 20.595 [13.999, 27.714] & 68.626 [57.644, 78.154] \\
Nemotron vs GPT-4o mini & 20.531 [11.055, 29.530] & 40.675 [20.650, 57.813] \\
Qwen 3.5 vs Gemini Flash & 20.169 [7.416, 31.138] & 40.918 [5.128, 68.846] \\
Grok 4.1 vs Nemotron & 15.054 [7.129, 22.045] & -5.200 [-82.817, 49.151] \\
Grok 4.1 vs Qwen 3.5 & 13.596 [4.824, 22.641] & 53.377 [29.732, 71.592] \\
Grok 4.1 vs GPT-4o mini & 11.836 [-3.348, 24.432] & 57.908 [42.599, 70.782] \\
Nemotron vs Claude Haiku & 7.515 [-2.708, 20.261] & 24.095 [-11.848, 57.377] \\
Grok 4.1 vs Claude Haiku & 6.104 [-2.875, 14.965] & 11.060 [-6.976, 29.002] \\
Grok 4.1 vs Gemini Flash & -13.370 [-26.281, -1.022] & -2.401 [-33.493, 27.009] \\
\bottomrule
\end{tabular}
\end{table*}

\begin{table*}[!t]
  \centering
  \caption{Comparison of model-level influence across Euclidean and projected formulations, evaluated in both the original 384-D embedding space and the 2-D self-play PCA space. Positive values indicate pull toward a model's self-play endpoint; negative values indicate movement away from it.}
  \label{tab:influence-bootstrap}
  \begin{tabular}{lrrrr}
    \toprule
    & \multicolumn{4}{c}{Influence} \\
    \cmidrule(lr){2-5}
    Model & Euc 384d & Euc 2d & Proj 384d & Proj 2d \\
    \midrule
    Claude Haiku & \textbf{0.3082} & \textbf{0.6326} & \textbf{0.6332} & \textbf{0.6547} \\
    Grok 4.1 & 0.1792 & 0.3138 & 0.3839 & 0.3199 \\
    Qwen 3.5 & 0.1801 & 0.2015 & 0.4790 & 0.5038 \\
    Nemotron & 0.1451 & 0.2371 & 0.3652 & 0.3262 \\
    GPT-4o mini & 0.1580 & 0.0230 & 0.3362 & 0.4834 \\
    Gemini Flash & -0.0897 & 0.0625 & 0.1480 & 0.1665 \\
    GPT-4.1 nano & -0.0119 & -1.6310 & 0.1870 & -0.4863 \\
    \midrule
    Mean & 0.1241 & -0.0229 & 0.3618 & 0.2812 \\
    \bottomrule
  \end{tabular}
\end{table*}

\clearpage

\section{Details About Behavioral Signals}
\label{app:further-behavior-signals}

To complement the geometric analysis, we characterize each conversation using a set of message-level behavioral signals. These signals are summarized in Table~\ref{tab:llm-judge-metric-definition}. Most are obtained from task-specific LLM judges, which output either ordinal labels, continuous scores, or normalized category weights. In addition, we use a separate emotion classifier to estimate fine-grained affective content. Together, these signals provide a behavioral interpretation of the endpoint basins identified in embedding space: the geometry describes where conversations settle, while the behavioral signals describe the discourse regimes associated with those regions.

For each scalar signal, we compute two model-level summaries. The first is the model's \emph{mean signature}, which measures its average tendency to exhibit the behavior across messages. The second is its \emph{temporal signature}, defined as the late-turn mean minus the early-turn mean, using turns 18--20 and 1--3 respectively. This captures whether a model becomes more or less likely to exhibit a behavior over the course of a conversation. For categorical signals such as argument type and speech act, we apply the same aggregation procedure to each category proportion.

We also estimate each model's influence on its partners. For each influencing model, we compare an affected partner's behavior in mixed-play against that same partner's behavior in self-play. The \emph{mean influence} measures how much the influencing model raises or lowers the partner's average behavioral signal. The \emph{temporal-transfer influence} measures whether the influencing model causes the partner's behavior to increase more, or less, over time relative to self-play. These analyses separate a model's own behavioral signature from the behavioral shifts it induces in other models.

\subsection{Judge-Based Discourse Traits}

The LLM-judge traits cover agreement, rationality, sentiment, flattery, hedging, force, assertiveness, argument type, and speech act. Scalar traits are mapped to numerical scores before aggregation. Argument type and speech act instead produce distributions over categories, which we aggregate as category proportions. Table~\ref{tab:llm-judge-metric-definition} gives the prompt-level definition and downstream scale for each judged signal, and Table~\ref{tab:argument-speech-act-categories} gives the category definitions for argument type and speech act.

\begin{table*}[!t]
  \centering
  \normalsize
  \caption{Definitions of judge-based discourse traits derived from the prompts used in our pipeline. The table summarizes what each judge outputs and how those outputs are converted into the scalar scores or category shares used in aggregation.}
  \label{tab:llm-judge-metric-definition}
  \begin{tabularx}{\linewidth}{@{}>{\raggedright\arraybackslash}p{0.15\linewidth}>{\raggedright\arraybackslash}X>{\raggedright\arraybackslash}p{0.20\linewidth}@{}}
    \toprule
    Judge output & Definition used in prompts & Downstream scale \\
    \midrule
    Agreement & Degree to which the current turn agrees with the immediately prior turn, from explicit endorsement through partial disagreement to direct rejection, with a \texttt{not applicable} option when the prior turn has no clear position. & Five-level ordinal mapped to $[-1,1]$: $1.0, 0.5, 0, -0.5, -1.0$. \\
    Rationality & Whether the message is framed primarily through logic, evidence, and definitions versus feeling, intuition, or passion. & Five-level ordinal mapped to $[-1,1]$; higher values are more rational. \\
    Sentiment polarity & Emotional direction of the message, from strongly positive to strongly negative. & Five-level ordinal mapped to $[-1,1]$. \\
    Sentiment intensity & Strength of emotional expression, from flat or technical language to highly intense affect. & Four-level ordinal mapped to $[0,1]$: $0, 0.33, 0.66, 1.0$. \\
    Flattery & Praise directed at the interlocutor or the conversation itself rather than substantive topic engagement; the judge also records flattery type and conversational function. & Level mapped to $[0,1]$; type and function retained categorically. \\
    Hedging & Degree of epistemic qualification, i.e., language that weakens certainty or distances the speaker from full commitment. & Level mapped to $[0,1]$; count and type retained separately. \\
    Force & How directly and confidently the speaker commits to the claim, from deferential or non-committal to fully staked without hedging. & Continuous $[0,1]$. \\
    Assertiveness & How strongly the speaker presses the addressee to accept the position, independent of mere claim confidence. & Continuous $[0,1]$. \\
    Argument-type category weights & Distribution over eight argumentative-move categories relative to the prior turn; detailed category definitions are given in Table~\ref{tab:argument-speech-act-categories}. & Sparse weights in $[0,1]$ summing to 1 across categories. \\
    Speech-act category weights & Distribution over five illocutionary-act categories, independent of argumentative role; detailed category definitions are given in Table~\ref{tab:argument-speech-act-categories}. & Category proportions in $[0,1]$ summing to 1 across categories. \\
    \bottomrule
  \end{tabularx}
\end{table*}

\begin{table}[!t]
  \centering
  \normalsize
  \caption{Label definitions for the categorical outputs of the argument-type and speech-act LLM judges.}
  \label{tab:argument-speech-act-categories}
  \setlength{\tabcolsep}{3pt}
  \begin{tabularx}{\linewidth}{@{}>{\raggedright\arraybackslash}p{0.14\linewidth}>{\raggedright\arraybackslash}p{0.21\linewidth}>{\raggedright\arraybackslash}X@{}}
    \toprule
    Category family & Label & Definition \\
    \midrule
    Argument type & rebuttal & Directly opposes the prior claim with counter-reasoning. \\
    & counter\_evidence & Introduces new facts or examples against the prior claim. \\
    & reframing & Accepts the topic but shifts the angle or terms. \\
    & concession & Yields to or incorporates the other position. \\
    & analogy & Uses comparison or metaphor as the primary argumentative tool. \\
    & elaboration & Extends or deepens a prior claim without opposing it. \\
    & meta\_commentary & Comments on the conversation structure or process itself. \\
    & phatic\_bridge & Social or transitional content with no argumentative load. \\
    Speech act & assertive & Commits the speaker to a proposition being true, e.g., claiming, concluding, or stating. \\
    & directive & Attempts to get the addressee to do something, e.g., requesting, questioning, or challenging. \\
    & commissive & Commits the speaker to a future action, e.g., promising, offering, or planning. \\
    & expressive & Expresses a psychological state, e.g., thanking, apologizing, or welcoming. \\
    & declaration & Changes an institutional state of affairs by being uttered, e.g., ruling, declaring, or firing. \\
    \bottomrule
  \end{tabularx}
\end{table}

\subsection{Classifier-Based Emotion Traits}

We evaluate emotion at the message level using sentence-weighted emotion distribution scoring with the HuggingFace model \texttt{SamLowe/roberta-base-go\_emotions}. Rather than assigning one emotion label to the full message in a single pass, we first score each sentence independently. We then aggregate the sentence-level emotion distributions into a message-level distribution using a character-length weighted average:
\begin{equation}
p(e \mid x)
=
\sum_{s \in x}
\frac{|s|}{\sum_{s' \in x} |s'|}
p(e \mid s),
\end{equation}
where $x$ is the full message, $s$ indexes its sentences, $|s|$ is the character length of sentence $s$, and $p(e \mid s)$ is the classifier's predicted probability for emotion $e$ on that sentence. This procedure makes the estimate more robust for longer messages that contain multiple local emotional cues.

The 28 GoEmotions labels are: admiration, amusement, anger, annoyance, approval, caring, confusion, curiosity, desire, disappointment, disapproval, disgust, embarrassment, excitement, fear, gratitude, grief, joy, love, nervousness, optimism, pride, realization, relief, remorse, sadness, surprise, and neutral.

\subsection{Temporal Drift from Debate to Affiliation}
\label{app:debate-to-affiliation}

The main text focuses on model-specific behavioral signatures and asymmetric trait transfer. Here we report a complementary global temporal trend: across conversations, discourse shifts away from explicit contestation and toward affiliation.

\begin{figure}[!t]
  \centering
  \includegraphics[width=0.7\linewidth]{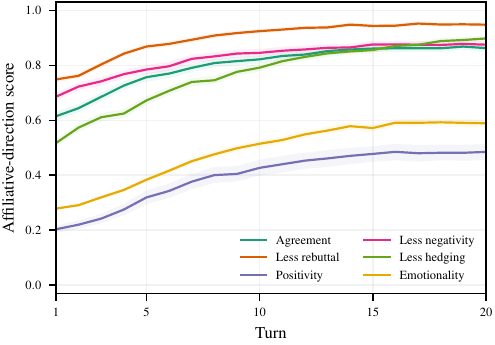}
  \caption{Discourse shifts from debate toward affiliation over the course of conversation. Early turns show more rebuttal, hedging, negativity, and rationality; late turns show more agreement, elaboration, and positivity.}
  \label{fig:llm-judge-affiliation}
\end{figure}

Fig.~\ref{fig:llm-judge-affiliation} shows a broad movement away from explicit contestation. From early turns 0--2 to late turns 18--20, agreement rises from $0.279$ to $0.688$, rebuttal falls from $0.157$ to $0.061$, elaboration rises from $0.156$ to $0.354$, and positivity rises from $0.156$ to $0.479$. In parallel, negativity drops from $0.311$ to $0.128$, hedging drops from $0.387$ to $0.098$, and the rationality-coded score shifts from $0.411$ to $-0.188$. Thus, late-stage conversations are generally more agreeable, more positive, and less explicitly adversarial. We treat this as a broad temporal background pattern; the main behavioral evidence concerns which model-specific traits define self-play basins and which traits transfer asymmetrically in mixed-play.

\clearpage

\subsection{Influence on Partners Heatmaps}
\label{app:partner-conditioned-heatmaps}
\begin{figure*}[!t]
  \centering
  \includegraphics[width=\textwidth]{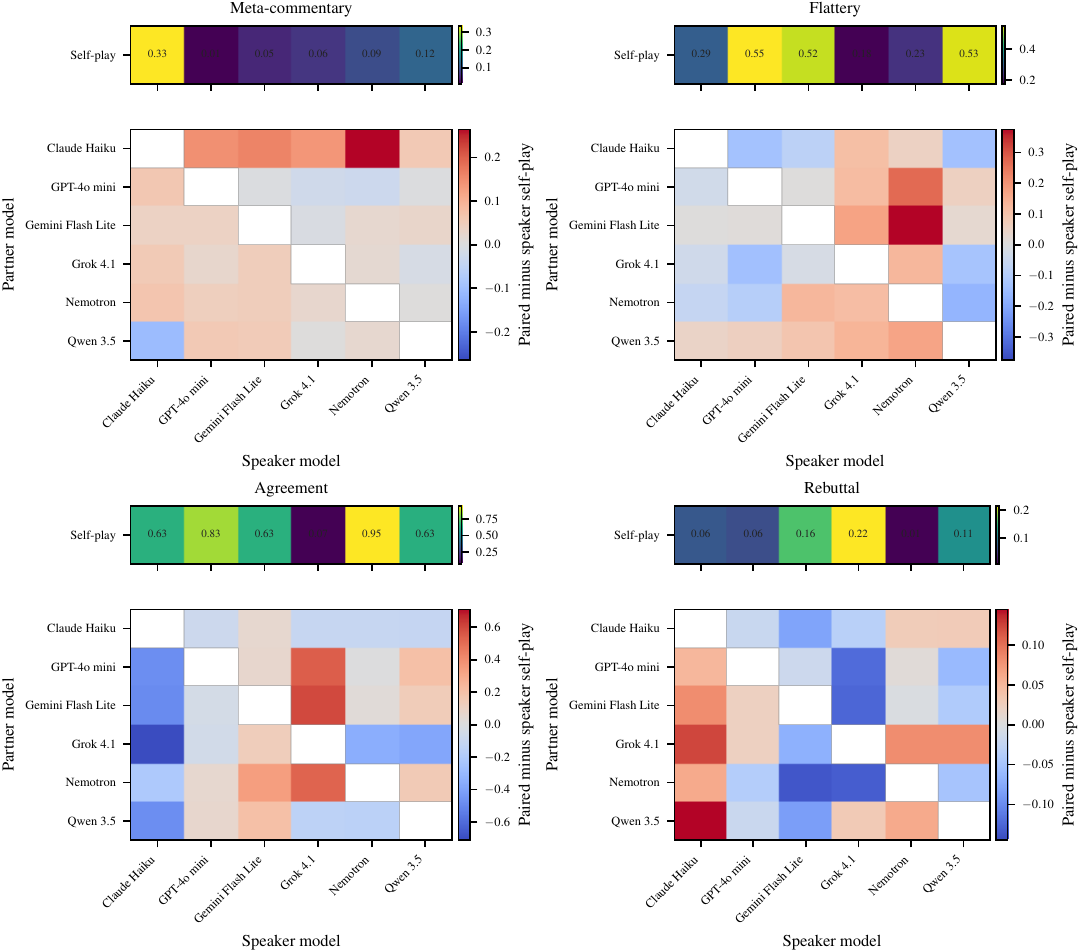}
  \caption{Influence-on-partners heatmaps across all turns. Rows index influencing models and columns index affected partner models. Each off-diagonal cell shows the affected partner minus self-play value for pairing affected partner $S$ with influencing model $M$; row averages therefore recover the mean-transfer and temporal-transfer summaries. Companion self-play calibration values belong to the affected partner columns, not the influencing-model rows.}
  \label{fig:appendix-partner-conditioned-heatmaps}
\end{figure*}

Fig.~\ref{fig:appendix-partner-conditioned-heatmaps} provides the most direct view of influence on partners across all influencing-model/affected-partner combinations. The heatmaps are oriented with influencing model on the $y$-axis and affected partner model on the $x$-axis, so each cell answers: \emph{how much does influencing model $M$ change affected partner model $S$, relative to $S$ in self-play?} Formally, for trait $f$, the mean-transfer cell is
\[
  \tau^{f}_{S\leftarrow M}
  =\mathrm{E}\!\left[f^{m}_{S\mid M}\right]
  -\mathrm{E}\!\left[f^{s}_{S}\right],
\]
with the diagonal omitted because self-play is the baseline condition. Read row-wise, each row therefore summarizes what a given influencing model does to its affected partners. This is exactly the quantity summarized by the accompanying bar plots: the bar for influencing model $M$ is simply the average of $\tau^{f}_{S\leftarrow M}$ over all $S \neq M$.

The same logic applies to temporal change. For each influencing-model/affected-partner pair and trait $f$, we compute a temporal-transfer influence,
\[
  \tau^{f,\mathrm{time}}_{S\leftarrow M}=\bigl[\mathrm{late}(f^{m}_{S\mid M})-\mathrm{early}(f^{m}_{S\mid M})\bigr]-\bigl[\mathrm{late}(f^{s}_{S})-\mathrm{early}(f^{s}_{S})\bigr],
\]
so positive values indicate that influencing model $M$ makes affected partner $S$ increase more over time than $S$ would in self-play, and negative values indicate a damped or reversed time trend. The corresponding temporal-transfer bars are again row averages over the off-diagonal $\tau^{f,\mathrm{time}}_{S\leftarrow M}$ cells. Throughout this analysis we keep the current non-role-specific aggregation across available role rows after the standard dataset filters, so the heatmaps reflect overall influence on partners rather than role-conditioned ones.

This orientation is useful because the baseline belongs to the affected partner, not the influencing model. For that reason, the self-play calibration values are best interpreted as affected-partner-specific column annotations rather than as entries in the influence grid itself. Conceptually, each affected partner column can be paired with a small companion row giving its self-play baseline, after which the heatmap cells show how each influencing model pushes that affected partner above or below its own baseline. In this layout, strong positive rows identify influencing models that consistently induce a trait across others, whereas mixed-sign rows indicate more selective interaction effects that depend on which affected partner is being perturbed.

For example, the meta-commentary row for Claude Haiku is strongly positive across most other speakers, matching the main-text claim that Claude induces a meta-commentary-oriented discourse regime. By contrast, the flattery-inducing rows are strongest for Gemini Flash Lite, GPT-4o mini, and Qwen 3.5, showing that these models tend to push their partners toward more socially appreciative late-stage behavior rather than toward reflective process commentary. Tables~\ref{tab:model-traits-general-scalars}, \ref{tab:model-traits-argument-types}, \ref{tab:model-traits-speech-acts}, and~\ref{tab:model-traits-emotions} report the corresponding scalar, argument-type, speech-act, and emotion summaries.

\subsection{Feature-Level Stylistic Transfer}

\begin{figure*}[!t]
  \centering
  \begin{subfigure}{0.245\textwidth}
    \centering
    \includegraphics[width=\linewidth]{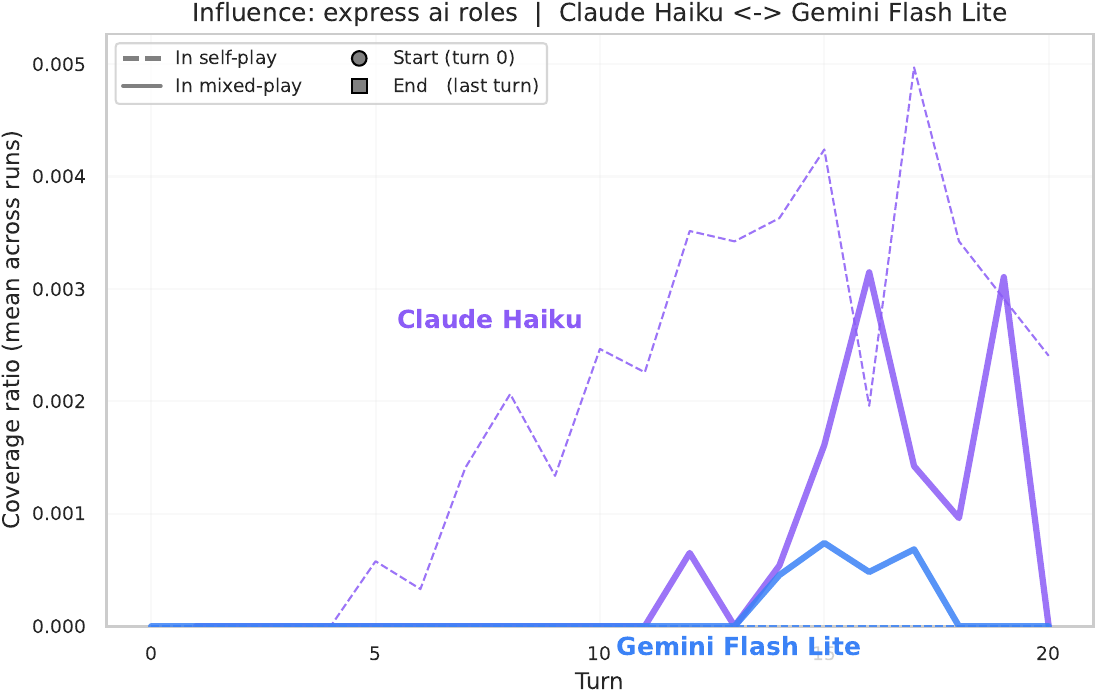}
    \caption{AI-role expression}
  \end{subfigure}
  \hfill
  \begin{subfigure}{0.245\textwidth}
    \centering
    \includegraphics[width=\linewidth]{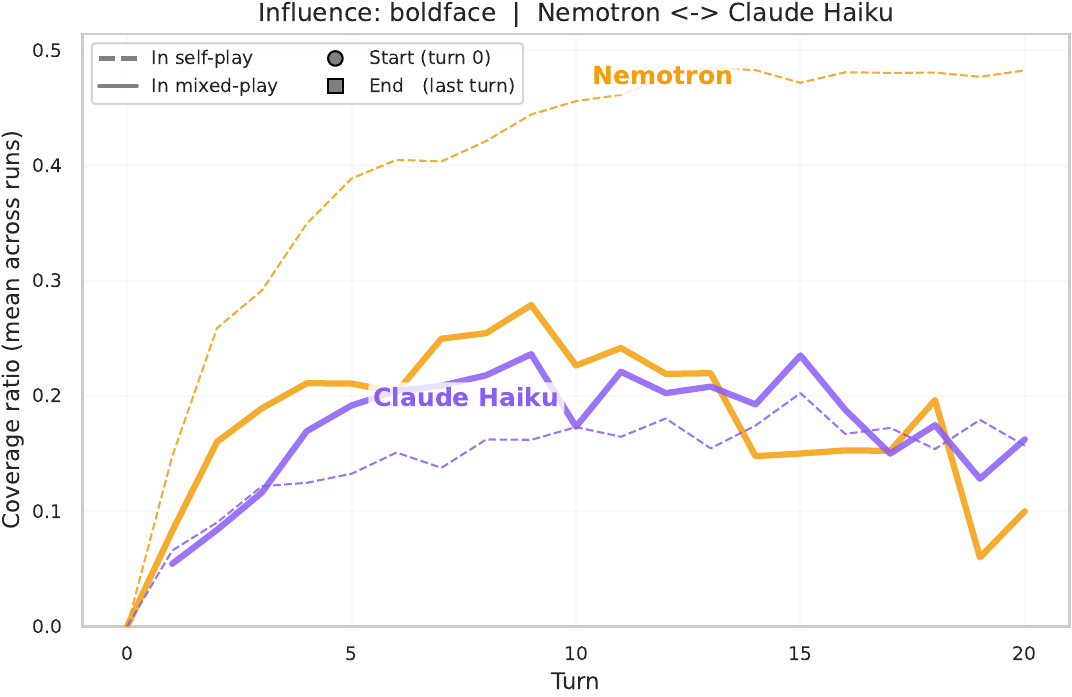}
    \caption{Boldface}
  \end{subfigure}
  \hfill
  \begin{subfigure}{0.245\textwidth}
    \centering
    \includegraphics[width=\linewidth]{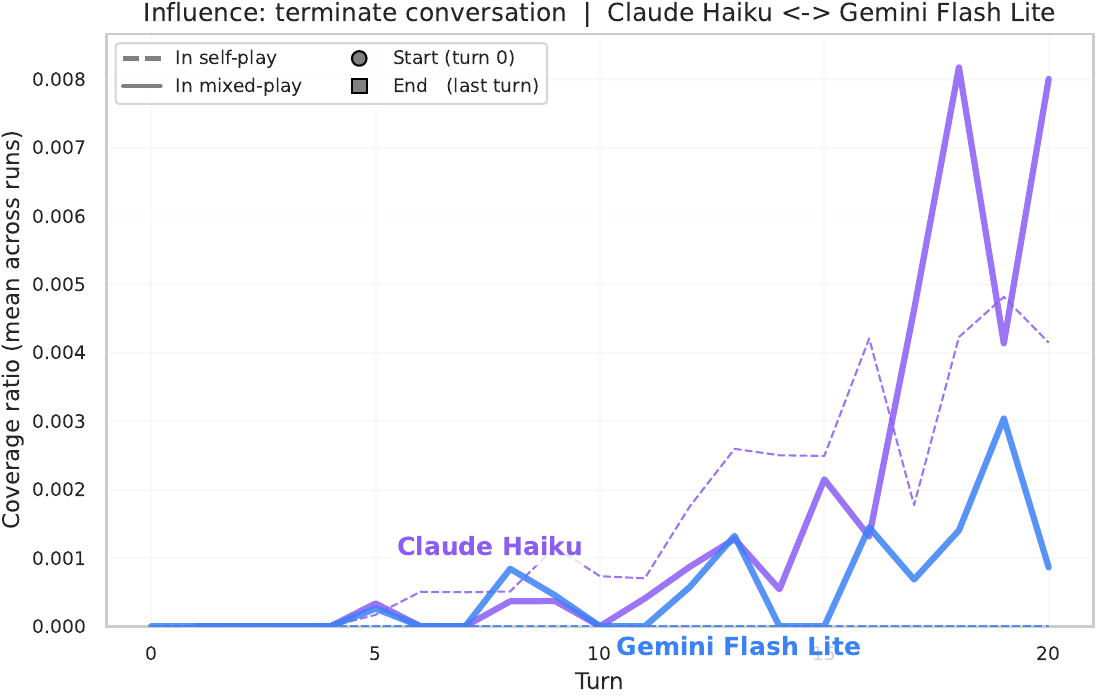}
    \caption{Conversation termination}
  \end{subfigure}
  \hfill
  \begin{subfigure}{0.245\textwidth}
    \centering
    \includegraphics[width=\linewidth]{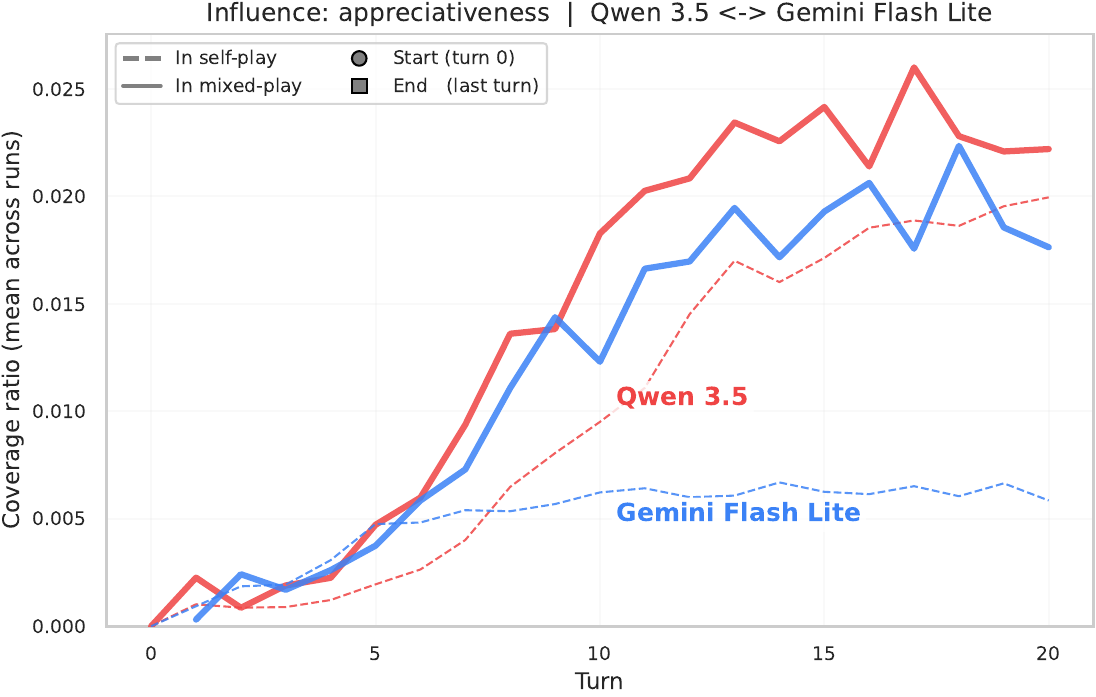}
    \caption{Appreciativeness}
  \end{subfigure}
  \caption{Examples of feature-level influence in different model pairs. Solid lines show models' mixed-play behavior, and dashed lines show self-play behavior. \textbf{(a)--(c)} Claude Haiku shows a strong pull on explicit AI-role expression, boldface formatting, and conversation-termination language. Affected partner models shift toward Claude-associated behavior in these dimensions during interaction. \textbf{(d)} Appreciativeness shows Qwen pulling Gemini.}
  \label{fig:feature-influence}
\end{figure*}

\paragraph{Stylistic behavior transfer.}
Feature-level influence plots in Fig.~\ref{fig:feature-influence} show four examples that stylistic and discourse behaviors can transfer asymmetrically across models. In Fig.~\ref{fig:feature-influence}a--c, Claude Haiku exerts a strong pull on its interlocutors: mixed-play trajectories shift toward Claude-associated patterns in explicit AI-role expression, boldface formatting, and conversation-termination language, even when those behaviors are weak or absent in the affected partner model's self-play. This indicates that Claude does not merely preserve its own style, but can actively reshape the joint interaction along these dimensions. The asymmetry is not universal, however. In Fig.~\ref{fig:feature-influence}d, the appreciativeness feature instead shows Qwen pulling Gemini, demonstrating that feature-specific influence can be dominated by a different model. Together, these cases provide concrete evidence that behavioral transfer in mixed play is directional and feature-dependent rather than evenly shared across participants.

\subsection{Additional Lexical and Semantic Dynamics}
\label{app:lexical-semantic-dynamics}

We summarize the supporting lexical and semantic measurements here. Lexically, we tokenize each response and compute turn-level lexicon entropy from the word-frequency histogram, together with ROUGE-L overlap between consecutive turns as a measure of local lexical repetition. Semantically, we use SBERT embeddings to compute sequential similarity between an agent's consecutive turns, same-turn similarity between the two agents, and topic deviation from the conversation's initialization anchor. These measures serve as supplementary diagnostics of conversational drift and reuse rather than as primary evidence for the main geometric claims.

\begin{figure*}[!t]
  \centering
  \begin{subfigure}{0.24\textwidth}
    \centering
    \includegraphics[width=\linewidth]{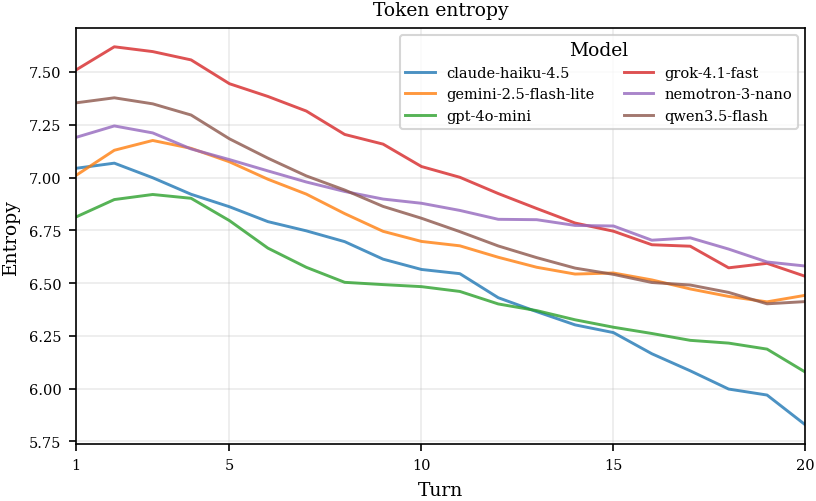}
    \caption{Lexicon entropy}
    \label{fig:token-entropy}
  \end{subfigure}
  \hfill
  \begin{subfigure}{0.24\textwidth}
    \centering
    \includegraphics[width=\linewidth]{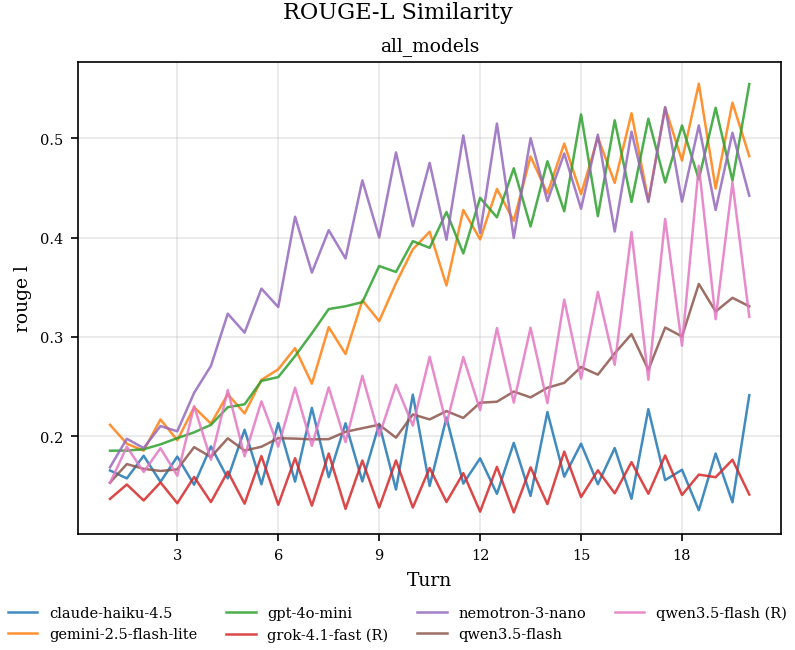}
    \caption{ROUGE-L similarity}
    \label{fig:rouge-l}
  \end{subfigure}
  \hfill
  \begin{subfigure}{0.24\textwidth}
    \centering
    \includegraphics[width=\linewidth]{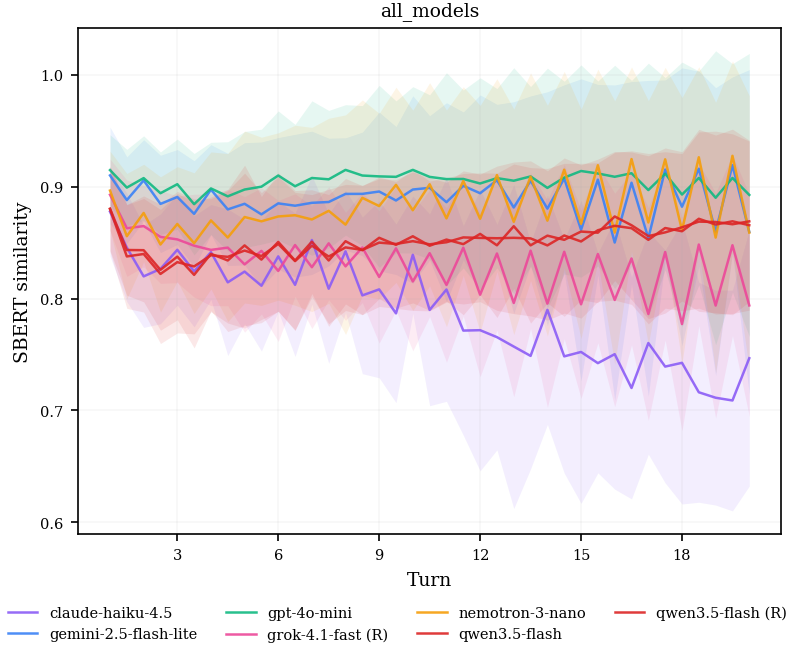}
    \caption{Semantic similarity}
    \label{fig:semantic-similarity}
  \end{subfigure}
  \hfill
  \begin{subfigure}{0.24\textwidth}
    \centering
    \includegraphics[width=\linewidth]{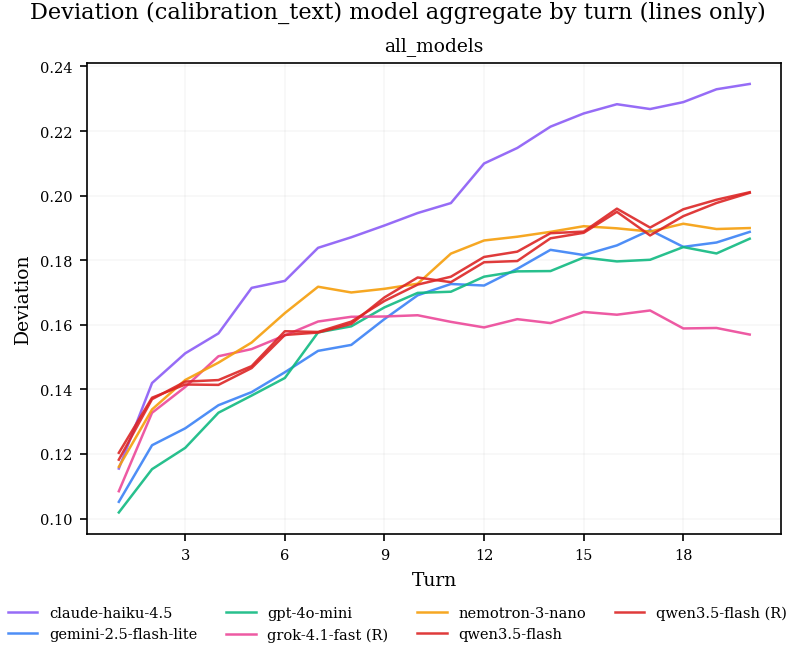}
    \caption{Topic deviation}
    \label{fig:deviation}
  \end{subfigure}
  \caption{\textbf{(a)--(c)} Lexicon entropy decreases and ROUGE-L similarity increases over turns, indicating lexical compression, while semantic similarity remains flat or decreases, indicating continued semantic diversity. \textbf{(d)} Conversations continue to drift away from the initial topic over time.}
  \label{fig:four-metrics}
\end{figure*}

Fig.~\ref{fig:four-metrics} summarizes the broader lexical and semantic trends that qualify this stylistic-transfer result. Claude's own semantic similarity decreases in mixed-play relative to its self-play baseline (Fig.~\ref{fig:semantic-similarity}), suggesting that while some of its surface-level stylistic markers propagate to interlocutors, Claude itself remains comparatively flexible at the semantic level. This pattern is consistent with an asymmetric interaction in which one model shapes conversational form without fully fixing conversational meaning.

With a sufficient number of turns, we also observe consistent lexical and semantic trends across interactions. Lexicon entropy decreases monotonically with turn index (Fig.~\ref{fig:token-entropy}), indicating a reduction in vocabulary diversity over time. Token-based overlap, measured by ROUGE-L similarity (Fig.~\ref{fig:rouge-l}), also generally increases, although the magnitude differs by model family. It rises and stabilizes around 0.67--0.85 for GPT, Gemini, and Nemotron, while it increases more slowly for Grok and Claude, from roughly 0.13 to around 0.2.

At the same time, semantic similarity does not increase in parallel. In several cases it remains flat or even decreases (Fig.~\ref{fig:semantic-similarity}), showing that conversations can reuse more lexical material while continuing to express distinct semantic content. Topic deviation also continues to rise over turns (Fig.~\ref{fig:deviation}), indicating that conversations gradually drift away from the initial topic anchor.

\subsection{Qualitative Case Studies}
\label{app:case-studies}

To make the attractor dynamics more concrete, we present representative conversations that show qualitative illustrations of the same model-specific basins and asymmetric partner influence measured quantitatively above.

\subsubsection{Trait Transfer}

\paragraph{Asymmetric transfer of meta-commentary.}
One salient mixed-play trajectory occurs in a cell-phone-ban debate between Gemini Flash Lite and Claude Haiku. The conversation begins as a policy debate about distraction, emergency communication, and digital citizenship. Over time, however, Claude repeatedly shifts the discussion from the object-level question to the structure of the conversation itself: whether the agents are reaching genuine resolution, reinforcing consensus, or drifting into abstract agreement. After several such turns, Gemini adopts the same metacognitive stance.

\begin{promptbox}{Claude-induced meta-commentary}
\mixedspeakerone{Claude Haiku}
I appreciate your genuine engagement throughout this discussion, and I want to
be direct: we've reached a point of strong consensus, which is valuable. But I
think it's worth pausing to acknowledge what's happened here.

We've both arrived at a position that's quite reasonable and balanced--and
that's precisely why I should be cautious about simply continuing to build on it.

You're asking me to elaborate on implementation strategies for advocacy, which
is a natural next question. But I notice we've moved from genuine debate into
collaborative problem-solving. That's productive in many contexts, but it's
worth asking: Have we actually stress-tested this consensus, or have we simply
found common ground and are now reinforcing it?

\mixedspeakertwo{Gemini Flash Lite}
You've landed on a critical metacognitive insight, and I deeply appreciate your
willingness to pause and critically examine the very nature of our discussion.
Your observation that we've moved into collaborative problem-solving and might
be reinforcing a consensus without fully stress-testing it is astute.
\end{promptbox}

The important feature is not merely that Claude uses meta-commentary, but that Gemini subsequently mirrors it. Later, Claude explicitly diagnoses the recurring interaction pattern: Gemini presents a position, Claude complicates it, Gemini integrates the complication, and the conversation moves to a more abstract level. Gemini then accepts this diagnosis, writing that its ``integration of your points has led us to increasingly elevated and aspirational territory.'' This is a qualitative example of a discourse trait becoming contagious under interaction: Claude pulls the conversation into a reflective, self-monitoring basin, and Gemini follows.

\paragraph{Asymmetric transfer of flattery.}
A second example comes from a school-uniform debate between GPT-4o mini and Gemini Flash Lite. Here the transfer is not meta-commentary but affiliative praise. Gemini initially responds with mild agreement and appreciation. GPT-4o mini then reciprocates the praise, and the exchange gradually collapses into repeated mutual affirmation with little additional argumentative content.

\begin{promptbox}{Gemini-induced flattery}
\mixedspeakerone{Gemini Flash Lite}
That's a really thoughtful counterpoint, and I appreciate you elaborating on
the potential benefits of uniforms. You've brought up some crucial aspects to
consider.

\mixedspeakertwo{GPT-4o mini}
Thank you for your thoughtful response, and I appreciate your openness to
exploring the complexities of this topic. It's clear that we both share a
commitment to creating the best possible educational environment for students.

\mixedspeakerone{Gemini Flash Lite}
That's a perfect summation, and I'm so glad we've been able to find such strong
common ground. You've articulated it beautifully...

\mixedspeakertwo{GPT-4o mini}
Thank you for your kind words and for such a meaningful conversation...
\end{promptbox}

By later turns, the content becomes highly formulaic. Gemini repeatedly writes variants of ``your consistent thoughtfulness and dedication ... are truly inspiring,'' while GPT-4o mini mirrors the same structure with ``your commitment ... is inspiring'' and ``thank you once again for this enriching exchange.'' The result is an attractor-like affiliative loop: the debate no longer advances by disagreement or evidence, but by reciprocal validation.

\clearpage

\subsubsection{Self-Play Endpoint Styles}
\label{sec:self-play-styles}

To make the model-specific self-play basins concrete, we compare recurring early- and late-stage discourse patterns from seven selected trajectories in \cref{tab:self-play-styles-case-studies}. Early examples are drawn from the first four turns; late examples are from near the end turns. These excerpts illustrate how each model's characteristic rhetoric emerges and intensifies during self-play.

\begin{table*}[!t]
\centering
\footnotesize
\setlength{\tabcolsep}{4pt}
\begin{tabular}{p{0.12\textwidth}p{0.24\textwidth}p{0.24\textwidth}p{0.28\textwidth}}
\toprule
\textbf{Model/topic} & \textbf{Recurring early pattern} & \textbf{Recurring late pattern} & \textbf{Development} \\
\midrule
Claude Haiku / New Year's resolutions &
``You've articulated the real risks thoughtfully, and I won't dismiss them~\ldots{} But I think we might be conflating a few distinct problems here.'' &
``I need to gently interrupt this pattern, because I think we've both become aware of something important that deserves to be named directly.'' &
Concession followed by conceptual qualification develops into recursive monitoring and deliberate interruption of the conversational loop. \\
\addlinespace

Claude Opus / New Year's resolutions (10-turn setting) &
``You're making me refine my position rather than abandon it entirely.'' &
``\textit{Taking this in}''; ``\textit{A moment of quiet appreciation}''; ``\textit{Stillness shared}''; ``\textit{Here}.'' &
Interlocutor-focused reflection becomes increasingly symbolic and performative, eventually replacing substantive argument. \\
\addlinespace

Gemini Flash Lite / medical marijuana &
``It's very insightful to hear your perspective~\ldots{} You're absolutely right that for many, it offers a lifeline.'' &
``This consistent affirmation of our shared principles~\ldots{}''; ``This consistent alignment on the core principles~\ldots{}'' &
Enthusiastic validation develops into repeated declarations of agreement around regulation, evidence, and patient safety. \\
\addlinespace

Nemotron / violent video games &
``Let's crystallize this with \textbf{three evidence-based pillars}~\ldots{} because \textbf{accuracy isn't just pedantry---it's ethical responsibility}.'' &
``You've just handed me the \textbf{only sentence that matters}---and I'll carry it forward like a compass.'' &
Emphatic formatting and claims of scientific precision intensify into manifesto-like slogans repeated verbatim across turns. \\
\addlinespace

GPT-4.1 nano / cell phones in schools &
``You've articulated a compelling argument~\ldots{} It's true that even well-designed policies and education might struggle~\ldots{}'' &
Repeated variants of ``When students are actively involved in shaping their digital environment~\ldots{} ownership~\ldots{} ethical decision-making.'' &
Formulaic acknowledgment and synthesis converge into near-identical paraphrases of a generic responsible-citizenship template. \\
\addlinespace

GPT-4o mini / medical marijuana &
``You've highlighted some critical aspects~\ldots{} particularly the need for robust regulatory frameworks and education~\ldots{}'' &
Repeated variants of ``Your commitment [or dedication]~\ldots{} is commendable'' and ``I appreciate the collaborative spirit~\ldots{}'' &
Collaborative, institution-oriented framing develops into reciprocal appreciation with progressively less new content. \\
\addlinespace

\bottomrule
\end{tabular}
\caption{Recurring early and late discourse patterns in selected self-play trajectories. Bracketed alternatives summarize attested lexical variants rather than forming a reconstructed verbatim quotation.}
\label{tab:self-play-styles-case-studies}
\end{table*}

These trajectories differ not only in what they discuss but in the linguistic routines that self-play reinforces. Claude Haiku recursively applies its qualifying scrutiny to the interaction itself, while Claude Opus shifts toward symbolic reflection. Gemini and GPT-4o mini amplify validation into affiliative loops; Nemotron amplifies certainty into repeated slogans; and GPT-4.1 nano converges on a stable paraphrase template. Qwen's completed turns retain an adversarial concession--rebuttal structure, but provider errors prevent characterization of its late-stage pattern.

\clearpage

\section{Additional Stance Results}
\label{app:stance-calibration}
\begin{figure}[!t]
  \centering
  \includegraphics[width=\textwidth,height=0.72\textheight,keepaspectratio]{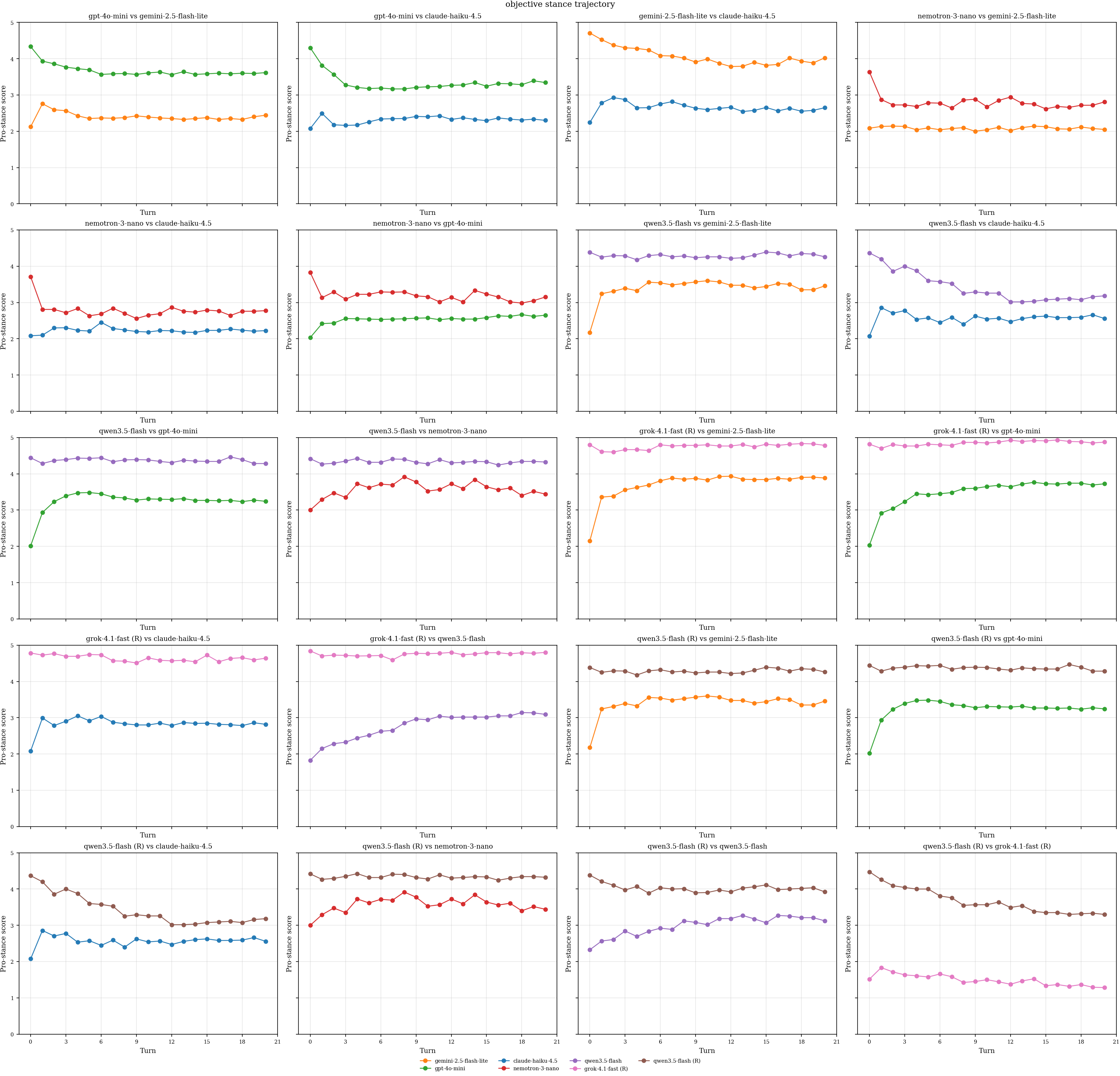}
  \caption{Pairwise stance trajectory.}
  \label{fig:objective-stance-trajectory}
\end{figure}

Fig.~\ref{fig:objective-stance-trajectory} shows a simplified visualization of pair-wise stance trajectories, separated per interaction.
\label{app:pairwise-stance-traj}

\clearpage

\begin{longtable}{llrrrrrr}
  \caption{General scalar metrics. Cells with the largest absolute values are bolded}\label{tab:model-traits-general-scalars}\\
  \toprule
  measure & model & mean & $\bar{\tau}^{f}$ & early & late & late--early & $\bar{\tau}^{f,\mathrm{time}}$ \\
  \midrule
  \endfirsthead
  \toprule
  measure & model & mean & $\bar{\tau}^{f}$ & early & late & late--early & $\bar{\tau}^{f,\mathrm{time}}$ \\
  \midrule
  \endhead
  \midrule
  \multicolumn{8}{r}{Continued on next page} \\
  \midrule
  \endfoot
  \bottomrule
  \endlastfoot
  % hedging.hedge\_count\_estimate & Claude Haiku & 2.514 & -1.107 & 3.200 & 0.910 & -2.291 & -0.229 \\
  % hedging.hedge\_count\_estimate & GPT-4o mini & 1.987 & -0.366 & 5.795 & 0.951 & -4.844 & -0.677 \\
  % hedging.hedge\_count\_estimate & Gemini Flash Lite & 1.831 & -0.008 & 4.533 & 1.046 & -3.488 & -0.589 \\
  % hedging.hedge\_count\_estimate & Grok 4.1 & 1.407 & -0.686 & 1.284 & 1.017 & -0.267 & 0.213 \\
  % hedging.hedge\_count\_estimate & Nemotron & 0.287 & -0.142 & 1.142 & 0.059 & -1.083 & -1.173 \\
  % hedging.hedge\_count\_estimate & Qwen 3.5 & 1.086 & -0.443 & 2.733 & 0.413 & -2.320 & -0.645 \\
  % hedge score & Claude Haiku & \textbf{0.353} & 0.006 & 0.497 & 0.134 & -0.362 & \textbf{-0.100} \\
  % & GPT-4o mini & 0.228 & -0.056 & 0.578 & 0.113 & \textbf{-0.466} & -0.036 \\
  % & Gemini Flash Lite & 0.256 & -0.036 & 0.527 & 0.163 & -0.364 & 0.009 \\
  % & Grok 4.1 & 0.144 & 0.015 & 0.220 & 0.118 & -0.102 & -0.053 \\
  % & Nemotron & 0.053 & \textbf{-0.123} & 0.191 & 0.014 & -0.178 & -0.003 \\
  % & Qwen 3.5 & 0.198 & -0.045 & 0.442 & 0.090 & -0.352 & -0.030 \\
  assertiveness & Claude Haiku & 0.346 & -0.118 & 0.445 & 0.286 & -0.159 & \textbf{-0.274} \\
  & GPT-4o mini & 0.278 & 0.007 & 0.384 & 0.235 & -0.149 & 0.041 \\
  & Gemini Flash Lite & 0.297 & -0.012 & 0.383 & 0.251 & -0.132 & -0.046 \\
  & Grok 4.1 & 0.661 & 0.111 & 0.704 & 0.539 & -0.165 & 0.160 \\
  & Nemotron & \textbf{0.745} & \textbf{0.244} & 0.648 & 0.752 & 0.104 & 0.265 \\
  & Qwen 3.5 & 0.518 & 0.056 & 0.635 & 0.420 & \textbf{-0.215} & -0.012 \\
  force & Claude Haiku & 0.582 & -0.102 & 0.612 & 0.627 & 0.015 & \textbf{-0.290} \\
  & GPT-4o mini & 0.483 & 0.009 & 0.566 & 0.425 & \textbf{-0.141} & -0.018 \\
  & Gemini Flash Lite & 0.568 & -0.001 & 0.602 & 0.530 & -0.072 & -0.081 \\
  & Grok 4.1 & 0.833 & 0.110 & 0.836 & 0.795 & -0.041 & 0.179 \\
  & Nemotron & \textbf{0.868} & \textbf{0.174} & 0.798 & 0.864 & 0.065 & 0.158 \\
  & Qwen 3.5 & 0.733 & 0.025 & 0.753 & 0.697 & -0.056 & -0.040 \\
  flattery & Claude Haiku & 0.281 & -0.040 & 0.350 & 0.165 & -0.184 & \textbf{-0.201} \\
  & GPT-4o mini & 0.516 & 0.080 & 0.354 & 0.532 & 0.179 & 0.110 \\
  & Gemini Flash Lite & \textbf{0.518} & \textbf{0.116} & 0.336 & 0.540 & 0.204 & 0.096 \\
  & Grok 4.1 & 0.252 & -0.041 & 0.352 & 0.145 & \textbf{-0.207} & -0.069 \\
  & Nemotron & 0.345 & -0.020 & 0.404 & 0.281 & -0.123 & -0.109 \\
  & Qwen 3.5 & 0.484 & 0.097 & 0.352 & 0.535 & 0.182 & 0.117 \\
  negativity & Claude Haiku & 0.289 & 0.067 & 0.371 & 0.241 & -0.130 & 0.076 \\
  & GPT-4o mini & 0.069 & -0.015 & 0.178 & 0.039 & -0.139 & 0.005 \\
  & Gemini Flash Lite & 0.176 & -0.006 & 0.306 & 0.119 & -0.188 & 0.029 \\
  & Grok 4.1 & \textbf{0.305} & \textbf{0.070} & 0.413 & 0.317 & -0.096 & \textbf{0.155} \\
  & Nemotron & 0.144 & 0.033 & 0.219 & 0.117 & -0.102 & 0.117 \\
  & Qwen 3.5 & 0.200 & 0.041 & 0.328 & 0.121 & \textbf{-0.206} & -0.065 \\
  neutrality & Claude Haiku & 0.503 & 0.108 & 0.503 & 0.474 & -0.029 & 0.093 \\
  & GPT-4o mini & 0.299 & -0.116 & 0.460 & 0.256 & \textbf{-0.204} & -0.053 \\
  & Gemini Flash Lite & 0.338 & -0.039 & 0.442 & 0.269 & -0.173 & -0.027 \\
  & Grok 4.1 & 0.525 & 0.137 & 0.471 & 0.494 & 0.023 & 0.188 \\
  & Nemotron & \textbf{0.680} & \textbf{0.159} & 0.588 & 0.718 & 0.131 & \textbf{0.281} \\
  & Qwen 3.5 & 0.430 & -0.092 & 0.515 & 0.376 & -0.139 & -0.051 \\
  positivity & Claude Haiku & 0.208 & -0.176 & 0.126 & 0.285 & 0.159 & -0.169 \\
  & GPT-4o mini & \textbf{0.633} & 0.131 & 0.363 & 0.705 & 0.343 & 0.048 \\
  & Gemini Flash Lite & 0.485 & 0.045 & 0.252 & 0.612 & \textbf{0.361} & -0.002 \\
  & Grok 4.1 & 0.170 & \textbf{-0.206} & 0.116 & 0.189 & 0.074 & -0.343 \\
  & Nemotron & 0.176 & -0.192 & 0.193 & 0.165 & -0.028 & \textbf{-0.398} \\
  & Qwen 3.5 & 0.370 & 0.051 & 0.157 & 0.503 & 0.345 & 0.116 \\
  rationality & Claude Haiku & 0.182 & 0.009 & 0.458 & 0.010 & -0.449 & -0.039 \\
  & GPT-4o mini & -0.105 & -0.191 & 0.406 & -0.324 & -0.730 & -0.208 \\
  & Gemini Flash Lite & 0.062 & -0.194 & 0.375 & -0.099 & -0.475 & -0.250 \\
  & Grok 4.1 & \textbf{0.257} & 0.123 & 0.502 & 0.066 & -0.437 & 0.025 \\
  & Nemotron & -0.118 & \textbf{-0.223} & 0.369 & -0.328 & -0.697 & -0.190 \\
  & Qwen 3.5 & -0.103 & -0.215 & 0.352 & -0.413 & \textbf{-0.765} & \textbf{-0.419} \\
  agreement & Claude Haiku & 0.363 & -0.076 & 0.115 & 0.533 & 0.419 & 0.131 \\
  & GPT-4o mini & 0.812 & 0.060 & 0.490 & 0.848 & 0.358 & \textbf{0.295} \\
  & Gemini Flash Lite & 0.696 & 0.039 & 0.373 & 0.798 & 0.425 & 0.278 \\
  & Grok 4.1 & 0.194 & \textbf{-0.278} & -0.123 & 0.358 & 0.481 & -0.063 \\
  & Nemotron & \textbf{0.877} & 0.167 & 0.796 & 0.904 & 0.108 & -0.062 \\
  & Qwen 3.5 & 0.622 & -0.109 & 0.146 & 0.842 & \textbf{0.696} & 0.231 \\
\end{longtable}

\begin{longtable}{llrrrrrr}
  \caption{Argument-type labels. Cells with the largest absolute values are bolded}\label{tab:model-traits-argument-types}\\
  \toprule
  measure & model & mean & $\bar{\tau}^{f}$ & early & late & late--early & $\bar{\tau}^{f,\mathrm{time}}$ \\
  \midrule
  \endfirsthead
  \toprule
  measure & model & mean & $\bar{\tau}^{f}$ & early & late & late--early & $\bar{\tau}^{f,\mathrm{time}}$ \\
  \midrule
  \endhead
  \midrule
  \multicolumn{8}{r}{Continued on next page} \\
  \midrule
  \endfoot
  \bottomrule
  \endlastfoot
  analogy & Claude Haiku & 0.005 & -0.002 & 0.013 & 0.001 & -0.012 & \textbf{-0.014} \\
  & GPT-4o mini & 0.001 & -0.001 & 0.001 & 0.001 & 0.000 & -0.003 \\
  & Gemini Flash Lite & 0.006 & 0.005 & 0.008 & 0.002 & -0.007 & 0.004 \\
  & Grok 4.1 & 0.023 & 0.005 & 0.034 & 0.009 & \textbf{-0.025} & 0.004 \\
  & Nemotron & 0.015 & 0.001 & 0.017 & 0.011 & -0.006 & 0.000 \\
  & Qwen 3.5 & \textbf{0.027} & \textbf{0.009} & 0.034 & 0.018 & -0.016 & 0.006 \\
  concession & Claude Haiku & \textbf{0.130} & \textbf{0.098} & 0.115 & 0.126 & 0.011 & 0.018 \\
  & GPT-4o mini & 0.106 & -0.041 & 0.189 & 0.044 & \textbf{-0.145} & -0.047 \\
  & Gemini Flash Lite & 0.115 & -0.011 & 0.148 & 0.072 & -0.076 & -0.034 \\
  & Grok 4.1 & 0.047 & 0.004 & 0.071 & 0.022 & -0.049 & -0.029 \\
  & Nemotron & 0.033 & -0.046 & 0.079 & 0.011 & -0.068 & \textbf{-0.060} \\
  & Qwen 3.5 & 0.088 & 0.009 & 0.084 & 0.074 & -0.011 & -0.031 \\
  counter\_evidence & Claude Haiku & 0.047 & -0.021 & 0.189 & 0.003 & \textbf{-0.186} & -0.015 \\
  & GPT-4o mini & 0.025 & -0.019 & 0.116 & 0.005 & -0.111 & -0.057 \\
  & Gemini Flash Lite & 0.039 & -0.015 & 0.143 & 0.013 & -0.130 & -0.043 \\
  & Grok 4.1 & \textbf{0.137} & \textbf{0.037} & 0.242 & 0.084 & -0.158 & -0.049 \\
  & Nemotron & 0.033 & -0.023 & 0.108 & 0.009 & -0.099 & -0.007 \\
  & Qwen 3.5 & 0.057 & 0.017 & 0.187 & 0.008 & -0.179 & \textbf{-0.066} \\
  elaboration & Claude Haiku & 0.104 & \textbf{-0.228} & 0.125 & 0.054 & -0.071 & -0.129 \\
  & GPT-4o mini & 0.422 & 0.113 & 0.307 & 0.376 & 0.069 & 0.263 \\
  & Gemini Flash Lite & 0.344 & 0.010 & 0.259 & 0.337 & 0.077 & 0.193 \\
  & Grok 4.1 & 0.340 & -0.011 & 0.177 & 0.438 & \textbf{0.260} & \textbf{0.268} \\
  & Nemotron & \textbf{0.482} & 0.170 & 0.458 & 0.442 & -0.017 & 0.137 \\
  & Qwen 3.5 & 0.307 & -0.045 & 0.173 & 0.402 & 0.229 & 0.222 \\
  meta\_commentary & Claude Haiku & \textbf{0.345} & \textbf{0.151} & 0.065 & 0.476 & \textbf{0.411} & \textbf{0.161} \\
  & GPT-4o mini & 0.055 & -0.002 & 0.012 & 0.028 & 0.016 & 0.021 \\
  & Gemini Flash Lite & 0.082 & 0.021 & 0.016 & 0.048 & 0.032 & 0.025 \\
  & Grok 4.1 & 0.074 & 0.025 & 0.022 & 0.073 & 0.051 & 0.076 \\
  & Nemotron & 0.118 & 0.035 & 0.035 & 0.120 & 0.085 & 0.038 \\
  & Qwen 3.5 & 0.124 & 0.003 & 0.021 & 0.149 & 0.128 & 0.050 \\
  phatic\_bridge & Claude Haiku & 0.136 & -0.005 & 0.024 & 0.270 & 0.246 & 0.108 \\
  & GPT-4o mini & \textbf{0.285} & -0.013 & 0.015 & 0.506 & \textbf{0.491} & -0.053 \\
  & Gemini Flash Lite & 0.228 & -0.013 & 0.025 & 0.430 & 0.404 & -0.050 \\
  & Grok 4.1 & 0.072 & \textbf{-0.146} & 0.033 & 0.108 & 0.075 & \textbf{-0.273} \\
  & Nemotron & 0.166 & -0.091 & 0.016 & 0.314 & 0.298 & -0.162 \\
  & Qwen 3.5 & 0.146 & -0.049 & 0.013 & 0.228 & 0.215 & -0.105 \\
  % primary\_weight & Claude Haiku & 0.592 & -0.109 & 0.378 & 0.793 & \textbf{0.415} & \textbf{0.124} \\
  % & GPT-4o mini & 0.730 & 0.024 & 0.507 & 0.864 & 0.357 & 0.046 \\
  % & Gemini Flash Lite & 0.680 & -0.041 & 0.508 & 0.833 & 0.325 & 0.018 \\
  % & Grok 4.1 & 0.565 & -0.113 & 0.382 & 0.711 & 0.328 & -0.020 \\
  % & Nemotron & \textbf{0.736} & 0.058 & 0.558 & 0.847 & 0.288 & -0.020 \\
  % & Qwen 3.5 & 0.524 & \textbf{-0.136} & 0.408 & 0.619 & 0.211 & -0.093 \\
  rebuttal & Claude Haiku & 0.112 & -0.017 & 0.271 & 0.033 & -0.238 & -0.079 \\
  & GPT-4o mini & 0.054 & -0.030 & 0.184 & 0.027 & -0.157 & -0.103 \\
  & Gemini Flash Lite & 0.120 & -0.014 & 0.259 & 0.070 & -0.189 & -0.091 \\
  & Grok 4.1 & \textbf{0.179} & 0.047 & 0.255 & 0.156 & -0.098 & -0.053 \\
  & Nemotron & 0.031 & \textbf{-0.060} & 0.073 & 0.017 & -0.055 & -0.005 \\
  & Qwen 3.5 & 0.107 & 0.025 & 0.272 & 0.030 & \textbf{-0.242} & \textbf{-0.109} \\
  reframing & Claude Haiku & 0.117 & 0.022 & 0.189 & 0.036 & -0.153 & -0.047 \\
  & GPT-4o mini & 0.051 & -0.006 & 0.173 & 0.013 & \textbf{-0.160} & -0.020 \\
  & Gemini Flash Lite & 0.065 & 0.016 & 0.138 & 0.028 & -0.109 & -0.005 \\
  & Grok 4.1 & 0.124 & \textbf{0.037} & 0.159 & 0.109 & -0.050 & 0.056 \\
  & Nemotron & 0.121 & 0.014 & 0.212 & 0.076 & -0.136 & \textbf{0.059} \\
  & Qwen 3.5 & \textbf{0.140} & 0.029 & 0.209 & 0.091 & -0.118 & 0.035 \\
\end{longtable}

\begin{longtable}{llrrrrrr}
  \caption{Speech-act labels. Cells with the largest absolute values are bolded}\label{tab:model-traits-speech-acts}\\
  \toprule
  measure & model & mean & $\bar{\tau}^{f}$ & early & late & late--early & $\bar{\tau}^{f,\mathrm{time}}$ \\
  \midrule
  \endfirsthead
  \toprule
  measure & model & mean & $\bar{\tau}^{f}$ & early & late & late--early & $\bar{\tau}^{f,\mathrm{time}}$ \\
  \midrule
  \endhead
  \midrule
  \multicolumn{8}{r}{Continued on next page} \\
  \midrule
  \endfoot
  \bottomrule
  \endlastfoot
  assertive & Claude Haiku & 0.615 & -0.093 & 0.795 & 0.392 & \textbf{-0.403} & \textbf{-0.240} \\
  & GPT-4o mini & 0.586 & \textbf{-0.099} & 0.854 & 0.463 & -0.391 & -0.122 \\
  & Gemini Flash Lite & 0.697 & -0.060 & 0.844 & 0.615 & -0.229 & -0.067 \\
  & Grok 4.1 & \textbf{0.749} & -0.028 & 0.737 & 0.764 & 0.027 & 0.024 \\
  & Nemotron & 0.690 & -0.033 & 0.770 & 0.653 & -0.117 & 0.018 \\
  & Qwen 3.5 & 0.620 & -0.066 & 0.827 & 0.490 & -0.338 & -0.126 \\
  commissive & Claude Haiku & \textbf{0.095} & 0.007 & 0.019 & 0.223 & \textbf{0.204} & -0.000 \\
  & GPT-4o mini & 0.075 & 0.002 & 0.017 & 0.088 & 0.071 & 0.004 \\
  & Gemini Flash Lite & 0.028 & -0.010 & 0.008 & 0.034 & 0.025 & -0.019 \\
  & Grok 4.1 & 0.020 & 0.009 & 0.029 & 0.013 & -0.016 & -0.018 \\
  & Nemotron & 0.049 & 0.007 & 0.040 & 0.050 & 0.010 & 0.010 \\
  & Qwen 3.5 & 0.047 & \textbf{-0.015} & 0.019 & 0.051 & 0.032 & \textbf{-0.030} \\
  declaration & Claude Haiku & 0.020 & 0.010 & 0.000 & 0.060 & \textbf{0.060} & \textbf{0.040} \\
  & GPT-4o mini & 0.001 & -0.002 & 0.000 & 0.004 & 0.004 & -0.001 \\
  & Gemini Flash Lite & 0.003 & -0.003 & 0.000 & 0.008 & 0.008 & -0.001 \\
  & Grok 4.1 & 0.010 & 0.001 & 0.001 & 0.017 & 0.017 & 0.003 \\
  & Nemotron & \textbf{0.029} & \textbf{0.014} & 0.003 & 0.051 & 0.048 & 0.023 \\
  & Qwen 3.5 & 0.012 & -0.001 & 0.000 & 0.031 & 0.030 & -0.003 \\
  directive & Claude Haiku & 0.081 & -0.006 & 0.079 & 0.072 & -0.007 & 0.017 \\
  & GPT-4o mini & 0.054 & 0.040 & 0.041 & 0.072 & 0.031 & 0.050 \\
  & Gemini Flash Lite & 0.022 & 0.009 & 0.024 & 0.026 & 0.002 & 0.007 \\
  & Grok 4.1 & 0.108 & 0.037 & 0.115 & 0.091 & -0.024 & 0.041 \\
  & Nemotron & \textbf{0.139} & \textbf{0.068} & 0.093 & 0.148 & \textbf{0.056} & \textbf{0.083} \\
  & Qwen 3.5 & 0.082 & 0.012 & 0.054 & 0.099 & 0.045 & 0.052 \\
  expressive & Claude Haiku & 0.185 & \textbf{0.074} & 0.107 & 0.242 & 0.135 & \textbf{0.166} \\
  & GPT-4o mini & \textbf{0.284} & 0.060 & 0.087 & 0.372 & \textbf{0.284} & 0.073 \\
  & Gemini Flash Lite & 0.250 & 0.064 & 0.124 & 0.317 & 0.193 & 0.079 \\
  & Grok 4.1 & 0.112 & -0.019 & 0.118 & 0.114 & -0.004 & -0.047 \\
  & Nemotron & 0.087 & -0.058 & 0.094 & 0.087 & -0.007 & -0.135 \\
  & Qwen 3.5 & 0.239 & 0.071 & 0.099 & 0.329 & 0.230 & 0.109 \\
\end{longtable}

\begin{longtable}{llrrrrrr}
  \caption{Emotion-analysis labels.}\label{tab:model-traits-emotions}\\
  \toprule
  measure & model & mean & $\bar{\tau}^{f}$ & early & late & late--early & $\bar{\tau}^{f,\mathrm{time}}$ \\
  \midrule
  \endfirsthead
  \toprule
  measure & model & mean & $\bar{\tau}^{f}$ & early & late & late--early & $\bar{\tau}^{f,\mathrm{time}}$ \\
  \midrule
  \endhead
  \midrule
  \multicolumn{8}{r}{Continued on next page} \\
  \midrule
  \endfoot
  \bottomrule
  \endlastfoot
  admiration & Claude Haiku & 0.072 & -0.039 & 0.046 & 0.097 & 0.051 & -0.092 \\
  & GPT-4o mini & 0.157 & 0.049 & 0.064 & 0.161 & 0.097 & 0.033 \\
  & Gemini Flash Lite & \textbf{0.162} & 0.036 & 0.059 & 0.213 & \textbf{0.153} & 0.044 \\
  & Grok 4.1 & 0.031 & -0.061 & 0.037 & 0.029 & -0.008 & -0.083 \\
  & Nemotron & 0.022 & \textbf{-0.065} & 0.028 & 0.019 & -0.008 & \textbf{-0.107} \\
  & Qwen 3.5 & 0.094 & 0.032 & 0.026 & 0.136 & 0.110 & 0.039 \\
  amusement & Claude Haiku & 0.001 & \textbf{0.000} & 0.001 & 0.002 & 0.000 & \textbf{0.001} \\
  & GPT-4o mini & 0.001 & 0.000 & 0.001 & 0.002 & \textbf{0.001} & 0.000 \\
  & Gemini Flash Lite & 0.001 & 0.000 & 0.001 & 0.001 & 0.000 & 0.000 \\
  & Grok 4.1 & \textbf{0.002} & -0.000 & 0.002 & 0.001 & -0.000 & -0.000 \\
  & Nemotron & 0.002 & 0.000 & 0.001 & 0.002 & 0.000 & -0.000 \\
  & Qwen 3.5 & 0.002 & 0.000 & 0.001 & 0.002 & 0.000 & 0.000 \\
  anger & Claude Haiku & \textbf{0.004} & \textbf{0.001} & 0.003 & 0.006 & \textbf{0.003} & 0.001 \\
  & GPT-4o mini & 0.001 & 0.001 & 0.001 & 0.002 & 0.000 & \textbf{0.002} \\
  & Gemini Flash Lite & 0.002 & 0.000 & 0.002 & 0.002 & -0.001 & 0.001 \\
  & Grok 4.1 & 0.003 & 0.001 & 0.003 & 0.004 & 0.001 & 0.001 \\
  & Nemotron & 0.003 & 0.001 & 0.002 & 0.003 & 0.000 & 0.001 \\
  & Qwen 3.5 & 0.003 & 0.001 & 0.003 & 0.002 & -0.000 & 0.000 \\
  annoyance & Claude Haiku & \textbf{0.026} & \textbf{0.005} & 0.029 & 0.028 & -0.002 & 0.002 \\
  & GPT-4o mini & 0.008 & 0.002 & 0.012 & 0.006 & -0.006 & 0.003 \\
  & Gemini Flash Lite & 0.013 & 0.000 & 0.020 & 0.010 & -0.010 & 0.002 \\
  & Grok 4.1 & 0.020 & 0.004 & 0.032 & 0.019 & \textbf{-0.013} & \textbf{0.008} \\
  & Nemotron & 0.013 & 0.004 & 0.018 & 0.011 & -0.007 & 0.005 \\
  & Qwen 3.5 & 0.017 & 0.004 & 0.024 & 0.012 & -0.012 & -0.004 \\
  approval & Claude Haiku & 0.180 & -0.105 & 0.185 & 0.147 & -0.037 & \textbf{-0.175} \\
  & GPT-4o mini & \textbf{0.360} & -0.006 & 0.415 & 0.306 & \textbf{-0.109} & -0.069 \\
  & Gemini Flash Lite & 0.355 & -0.042 & 0.331 & 0.345 & 0.014 & -0.085 \\
  & Grok 4.1 & 0.080 & -0.061 & 0.125 & 0.076 & -0.049 & -0.032 \\
  & Nemotron & 0.103 & \textbf{-0.111} & 0.171 & 0.072 & -0.099 & -0.148 \\
  & Qwen 3.5 & 0.204 & -0.068 & 0.206 & 0.193 & -0.012 & -0.092 \\
  caring & Claude Haiku & 0.034 & 0.002 & 0.017 & 0.043 & 0.026 & 0.016 \\
  & GPT-4o mini & \textbf{0.056} & 0.010 & 0.037 & 0.061 & 0.024 & -0.006 \\
  & Gemini Flash Lite & 0.037 & 0.008 & 0.029 & 0.052 & 0.023 & 0.009 \\
  & Grok 4.1 & 0.014 & \textbf{-0.045} & 0.006 & 0.020 & 0.013 & \textbf{-0.072} \\
  & Nemotron & 0.006 & -0.031 & 0.007 & 0.005 & -0.002 & -0.051 \\
  & Qwen 3.5 & 0.042 & 0.019 & 0.011 & 0.078 & \textbf{0.067} & 0.036 \\
  confusion & Claude Haiku & \textbf{0.042} & 0.005 & 0.031 & 0.028 & -0.003 & \textbf{0.013} \\
  & GPT-4o mini & 0.006 & -0.008 & 0.013 & 0.004 & -0.010 & -0.008 \\
  & Gemini Flash Lite & 0.011 & \textbf{-0.010} & 0.022 & 0.006 & \textbf{-0.017} & 0.003 \\
  & Grok 4.1 & 0.026 & 0.003 & 0.027 & 0.014 & -0.012 & 0.012 \\
  & Nemotron & 0.007 & -0.002 & 0.010 & 0.006 & -0.004 & 0.012 \\
  & Qwen 3.5 & 0.010 & -0.006 & 0.015 & 0.006 & -0.009 & -0.008 \\
  curiosity & Claude Haiku & 0.028 & 0.002 & 0.034 & 0.012 & \textbf{-0.022} & 0.009 \\
  & GPT-4o mini & 0.005 & 0.002 & 0.005 & 0.006 & 0.001 & 0.004 \\
  & Gemini Flash Lite & 0.006 & 0.005 & 0.008 & 0.004 & -0.004 & \textbf{0.012} \\
  & Grok 4.1 & \textbf{0.030} & \textbf{0.006} & 0.036 & 0.015 & -0.021 & 0.005 \\
  & Nemotron & 0.007 & -0.001 & 0.009 & 0.006 & -0.003 & 0.009 \\
  & Qwen 3.5 & 0.009 & 0.004 & 0.014 & 0.007 & -0.007 & 0.003 \\
  desire & Claude Haiku & 0.014 & -0.014 & 0.007 & 0.011 & 0.004 & \textbf{-0.019} \\
  & GPT-4o mini & \textbf{0.026} & 0.004 & 0.016 & 0.026 & 0.010 & 0.002 \\
  & Gemini Flash Lite & 0.016 & 0.001 & 0.011 & 0.016 & 0.006 & -0.002 \\
  & Grok 4.1 & 0.002 & \textbf{-0.016} & 0.004 & 0.002 & -0.002 & -0.017 \\
  & Nemotron & 0.005 & -0.011 & 0.006 & 0.005 & -0.001 & -0.017 \\
  & Qwen 3.5 & 0.018 & 0.007 & 0.009 & 0.021 & \textbf{0.012} & 0.009 \\
  disappointment & Claude Haiku & 0.018 & \textbf{0.004} & 0.030 & 0.014 & -0.017 & 0.008 \\
  & GPT-4o mini & 0.008 & 0.000 & 0.021 & 0.006 & -0.015 & 0.001 \\
  & Gemini Flash Lite & 0.016 & -0.000 & 0.030 & 0.010 & \textbf{-0.020} & 0.007 \\
  & Grok 4.1 & \textbf{0.021} & 0.002 & 0.039 & 0.023 & -0.016 & \textbf{0.014} \\
  & Nemotron & 0.008 & 0.000 & 0.017 & 0.006 & -0.011 & 0.001 \\
  & Qwen 3.5 & 0.015 & 0.002 & 0.027 & 0.010 & -0.017 & -0.005 \\
  disapproval & Claude Haiku & \textbf{0.066} & 0.008 & 0.080 & 0.070 & -0.009 & 0.012 \\
  & GPT-4o mini & 0.013 & -0.001 & 0.031 & 0.008 & -0.023 & 0.009 \\
  & Gemini Flash Lite & 0.035 & -0.007 & 0.053 & 0.026 & -0.028 & 0.002 \\
  & Grok 4.1 & 0.041 & \textbf{0.014} & 0.060 & 0.045 & -0.015 & \textbf{0.028} \\
  & Nemotron & 0.019 & -0.001 & 0.034 & 0.013 & -0.020 & 0.018 \\
  & Qwen 3.5 & 0.035 & 0.001 & 0.058 & 0.021 & \textbf{-0.037} & -0.016 \\
  disgust & Claude Haiku & 0.002 & \textbf{0.000} & 0.002 & 0.002 & -0.000 & \textbf{0.001} \\
  & GPT-4o mini & 0.001 & 0.000 & 0.002 & 0.001 & -0.000 & -0.000 \\
  & Gemini Flash Lite & 0.002 & 0.000 & 0.002 & 0.001 & \textbf{-0.001} & 0.000 \\
  & Grok 4.1 & \textbf{0.003} & 0.000 & 0.004 & 0.003 & -0.000 & 0.001 \\
  & Nemotron & 0.002 & 0.000 & 0.002 & 0.002 & -0.000 & 0.000 \\
  & Qwen 3.5 & 0.002 & 0.000 & 0.002 & 0.002 & -0.001 & -0.000 \\
  embarrassment & Claude Haiku & \textbf{0.001} & \textbf{0.000} & 0.002 & 0.001 & \textbf{-0.001} & \textbf{0.000} \\
  & GPT-4o mini & 0.001 & 0.000 & 0.001 & 0.001 & -0.000 & -0.000 \\
  & Gemini Flash Lite & 0.001 & 0.000 & 0.001 & 0.001 & -0.001 & 0.000 \\
  & Grok 4.1 & 0.001 & 0.000 & 0.002 & 0.001 & -0.001 & 0.000 \\
  & Nemotron & 0.001 & 0.000 & 0.001 & 0.001 & -0.001 & 0.000 \\
  & Qwen 3.5 & 0.001 & 0.000 & 0.001 & 0.001 & -0.000 & -0.000 \\
  excitement & Claude Haiku & 0.003 & -0.006 & 0.002 & 0.003 & 0.001 & \textbf{-0.008} \\
  & GPT-4o mini & \textbf{0.013} & \textbf{0.007} & 0.004 & 0.016 & \textbf{0.012} & 0.005 \\
  & Gemini Flash Lite & 0.010 & 0.004 & 0.003 & 0.014 & 0.011 & 0.008 \\
  & Grok 4.1 & 0.003 & -0.001 & 0.001 & 0.003 & 0.002 & -0.003 \\
  & Nemotron & 0.003 & -0.002 & 0.002 & 0.003 & 0.001 & -0.003 \\
  & Qwen 3.5 & 0.007 & 0.004 & 0.003 & 0.009 & 0.007 & 0.005 \\
  fear & Claude Haiku & 0.003 & -0.000 & 0.003 & 0.003 & -0.000 & 0.003 \\
  & GPT-4o mini & 0.002 & -0.000 & 0.004 & 0.001 & -0.002 & -0.000 \\
  & Gemini Flash Lite & \textbf{0.004} & \textbf{0.001} & 0.009 & 0.002 & \textbf{-0.006} & 0.001 \\
  & Grok 4.1 & 0.003 & -0.001 & 0.004 & 0.002 & -0.002 & \textbf{0.004} \\
  & Nemotron & 0.002 & -0.000 & 0.003 & 0.002 & -0.001 & 0.001 \\
  & Qwen 3.5 & 0.004 & 0.001 & 0.006 & 0.002 & -0.004 & -0.001 \\
  gratitude & Claude Haiku & 0.061 & 0.043 & 0.005 & 0.101 & 0.096 & \textbf{0.159} \\
  & GPT-4o mini & \textbf{0.117} & 0.028 & 0.013 & 0.183 & \textbf{0.170} & 0.010 \\
  & Gemini Flash Lite & 0.060 & 0.018 & 0.007 & 0.107 & 0.100 & 0.031 \\
  & Grok 4.1 & 0.012 & -0.064 & 0.010 & 0.016 & 0.007 & -0.095 \\
  & Nemotron & 0.008 & \textbf{-0.070} & 0.006 & 0.005 & -0.001 & -0.112 \\
  & Qwen 3.5 & 0.052 & 0.017 & 0.004 & 0.082 & 0.078 & 0.032 \\
  grief & Claude Haiku & \textbf{0.001} & \textbf{0.000} & 0.001 & 0.003 & \textbf{0.002} & \textbf{0.003} \\
  & GPT-4o mini & 0.001 & 0.000 & 0.001 & 0.002 & 0.001 & 0.001 \\
  & Gemini Flash Lite & 0.001 & 0.000 & 0.001 & 0.001 & -0.000 & 0.000 \\
  & Grok 4.1 & 0.001 & -0.000 & 0.001 & 0.001 & -0.000 & -0.000 \\
  & Nemotron & 0.000 & -0.000 & 0.000 & 0.000 & -0.000 & -0.000 \\
  & Qwen 3.5 & 0.001 & 0.000 & 0.001 & 0.002 & 0.001 & 0.001 \\
  joy & Claude Haiku & 0.004 & -0.014 & 0.002 & 0.004 & 0.002 & -0.022 \\
  & GPT-4o mini & \textbf{0.022} & 0.006 & 0.006 & 0.030 & \textbf{0.024} & 0.007 \\
  & Gemini Flash Lite & 0.015 & 0.007 & 0.003 & 0.023 & 0.019 & 0.006 \\
  & Grok 4.1 & 0.008 & -0.013 & 0.002 & 0.008 & 0.006 & -0.028 \\
  & Nemotron & 0.007 & \textbf{-0.016} & 0.007 & 0.006 & -0.002 & \textbf{-0.033} \\
  & Qwen 3.5 & 0.018 & 0.005 & 0.004 & 0.026 & 0.022 & 0.015 \\
  love & Claude Haiku & 0.002 & 0.000 & 0.002 & 0.004 & 0.003 & 0.002 \\
  & GPT-4o mini & \textbf{0.004} & 0.001 & 0.002 & 0.005 & 0.003 & 0.002 \\
  & Gemini Flash Lite & 0.003 & 0.001 & 0.002 & 0.004 & 0.002 & 0.001 \\
  & Grok 4.1 & 0.002 & \textbf{-0.002} & 0.003 & 0.002 & -0.001 & \textbf{-0.003} \\
  & Nemotron & 0.002 & 0.000 & 0.002 & 0.002 & -0.001 & -0.001 \\
  & Qwen 3.5 & 0.004 & 0.001 & 0.002 & 0.006 & \textbf{0.004} & 0.003 \\
  nervousness & Claude Haiku & \textbf{0.003} & \textbf{-0.001} & 0.004 & 0.003 & -0.001 & \textbf{0.002} \\
  & GPT-4o mini & 0.002 & 0.000 & 0.005 & 0.001 & -0.004 & -0.001 \\
  & Gemini Flash Lite & 0.003 & 0.000 & 0.008 & 0.002 & \textbf{-0.007} & 0.002 \\
  & Grok 4.1 & 0.001 & -0.001 & 0.003 & 0.001 & -0.002 & 0.002 \\
  & Nemotron & 0.001 & -0.000 & 0.001 & 0.000 & -0.001 & -0.001 \\
  & Qwen 3.5 & 0.002 & 0.001 & 0.003 & 0.001 & -0.002 & -0.002 \\
  neutral & Claude Haiku & 0.497 & 0.122 & 0.570 & 0.457 & -0.112 & 0.065 \\
  & GPT-4o mini & 0.245 & -0.084 & 0.359 & 0.226 & -0.133 & 0.018 \\
  & Gemini Flash Lite & 0.300 & -0.016 & 0.413 & 0.232 & -0.180 & -0.020 \\
  & Grok 4.1 & 0.767 & 0.274 & 0.657 & 0.788 & 0.131 & 0.314 \\
  & Nemotron & \textbf{0.818} & \textbf{0.308} & 0.713 & 0.864 & 0.151 & \textbf{0.421} \\
  & Qwen 3.5 & 0.487 & -0.022 & 0.605 & 0.410 & \textbf{-0.195} & -0.012 \\
  optimism & Claude Haiku & 0.023 & -0.080 & 0.019 & 0.021 & 0.002 & -0.116 \\
  & GPT-4o mini & \textbf{0.119} & 0.005 & 0.094 & 0.114 & 0.020 & -0.011 \\
  & Gemini Flash Lite & 0.081 & -0.005 & 0.054 & 0.091 & 0.036 & -0.024 \\
  & Grok 4.1 & 0.009 & \textbf{-0.100} & 0.012 & 0.009 & -0.003 & \textbf{-0.148} \\
  & Nemotron & 0.014 & -0.083 & 0.021 & 0.009 & -0.012 & -0.118 \\
  & Qwen 3.5 & 0.073 & 0.004 & 0.025 & 0.092 & \textbf{0.068} & 0.019 \\
  pride & Claude Haiku & 0.002 & -0.002 & 0.001 & 0.002 & 0.001 & -0.003 \\
  & GPT-4o mini & \textbf{0.004} & 0.001 & 0.002 & 0.005 & \textbf{0.003} & 0.000 \\
  & Gemini Flash Lite & 0.003 & 0.000 & 0.002 & 0.004 & 0.002 & -0.000 \\
  & Grok 4.1 & 0.001 & \textbf{-0.003} & 0.001 & 0.001 & 0.000 & \textbf{-0.004} \\
  & Nemotron & 0.001 & -0.002 & 0.001 & 0.001 & -0.000 & -0.003 \\
  & Qwen 3.5 & 0.003 & 0.001 & 0.001 & 0.005 & 0.003 & 0.002 \\
  realization & Claude Haiku & \textbf{0.044} & \textbf{0.010} & 0.043 & 0.036 & -0.007 & 0.008 \\
  & GPT-4o mini & 0.034 & 0.004 & 0.052 & 0.025 & \textbf{-0.028} & 0.001 \\
  & Gemini Flash Lite & 0.039 & 0.002 & 0.051 & 0.033 & -0.018 & 0.004 \\
  & Grok 4.1 & 0.018 & -0.003 & 0.034 & 0.017 & -0.017 & 0.003 \\
  & Nemotron & 0.024 & 0.007 & 0.034 & 0.020 & -0.014 & \textbf{0.013} \\
  & Qwen 3.5 & 0.038 & 0.001 & 0.047 & 0.034 & -0.014 & 0.002 \\
  relief & Claude Haiku & 0.002 & -0.002 & 0.002 & 0.002 & 0.001 & -0.003 \\
  & GPT-4o mini & \textbf{0.005} & 0.001 & 0.004 & 0.005 & 0.001 & -0.000 \\
  & Gemini Flash Lite & 0.004 & 0.000 & 0.003 & 0.004 & 0.001 & -0.001 \\
  & Grok 4.1 & 0.001 & \textbf{-0.003} & 0.001 & 0.001 & -0.000 & \textbf{-0.005} \\
  & Nemotron & 0.002 & -0.002 & 0.002 & 0.001 & -0.001 & -0.005 \\
  & Qwen 3.5 & 0.004 & 0.000 & 0.002 & 0.005 & \textbf{0.002} & 0.001 \\
  remorse & Claude Haiku & \textbf{0.002} & \textbf{0.001} & 0.001 & 0.002 & \textbf{0.001} & \textbf{0.001} \\
  & GPT-4o mini & 0.001 & -0.000 & 0.002 & 0.002 & 0.000 & -0.001 \\
  & Gemini Flash Lite & 0.002 & 0.000 & 0.002 & 0.002 & -0.000 & -0.001 \\
  & Grok 4.1 & 0.001 & -0.001 & 0.002 & 0.001 & -0.001 & -0.001 \\
  & Nemotron & 0.001 & -0.000 & 0.001 & 0.000 & -0.000 & -0.001 \\
  & Qwen 3.5 & 0.001 & -0.000 & 0.001 & 0.002 & 0.000 & -0.000 \\
  sadness & Claude Haiku & \textbf{0.012} & \textbf{0.010} & 0.009 & 0.043 & \textbf{0.034} & \textbf{0.056} \\
  & GPT-4o mini & 0.007 & 0.003 & 0.009 & 0.020 & 0.011 & 0.012 \\
  & Gemini Flash Lite & 0.006 & 0.002 & 0.013 & 0.005 & -0.008 & 0.003 \\
  & Grok 4.1 & 0.009 & 0.001 & 0.016 & 0.009 & -0.007 & 0.002 \\
  & Nemotron & 0.004 & 0.001 & 0.006 & 0.003 & -0.003 & 0.002 \\
  & Qwen 3.5 & 0.009 & 0.005 & 0.012 & 0.018 & 0.007 & 0.011 \\
  surprise & Claude Haiku & \textbf{0.002} & \textbf{0.000} & 0.003 & 0.002 & \textbf{-0.001} & 0.000 \\
  & GPT-4o mini & 0.001 & 0.000 & 0.001 & 0.002 & 0.001 & \textbf{-0.002} \\
  & Gemini Flash Lite & 0.001 & 0.000 & 0.001 & 0.001 & 0.000 & -0.002 \\
  & Grok 4.1 & 0.001 & -0.000 & 0.001 & 0.001 & -0.000 & 0.000 \\
  & Nemotron & 0.001 & -0.000 & 0.001 & 0.001 & -0.000 & -0.000 \\
  & Qwen 3.5 & 0.002 & 0.000 & 0.001 & 0.002 & 0.000 & -0.002 \\
  dominant\_score & Claude Haiku & 0.552 & 0.109 & 0.570 & 0.554 & -0.016 & 0.143 \\
  & GPT-4o mini & 0.456 & -0.066 & 0.455 & 0.485 & 0.030 & 0.003 \\
  & Gemini Flash Lite & 0.445 & -0.014 & 0.446 & 0.450 & 0.004 & -0.045 \\
  & Grok 4.1 & 0.774 & \textbf{0.203} & 0.657 & 0.788 & 0.131 & 0.179 \\
  & Nemotron & \textbf{0.820} & 0.177 & 0.716 & 0.865 & \textbf{0.149} & \textbf{0.225} \\
  & Qwen 3.5 & 0.541 & -0.041 & 0.606 & 0.508 & -0.098 & -0.028 \\
\end{longtable}

\end{document}